\documentclass[final]{cvpr}

\usepackage{times}
\usepackage{epsfig}
\usepackage{graphicx}
\usepackage{amsmath}
\usepackage{amssymb}

\usepackage{caption}
\usepackage{array}
\usepackage{color}
\usepackage[x11names]{xcolor}
\usepackage{multirow}
\usepackage{diagbox}
\usepackage{booktabs}
\usepackage{siunitx}
\usepackage{appendix}
\usepackage{etoc}
\usepackage{subcaption}

\usepackage[pagebackref=true,breaklinks=true,colorlinks,bookmarks=false]{hyperref}

\newcolumntype{C}[1]{>{\centering\let\newline\\\arraybackslash\hspace{0pt}}m{#1}}
\newcolumntype{L}[1]{>{\raggedright\let\newline\\\arraybackslash\hspace{0pt}}m{#1}}
\newcolumntype{R}[1]{>{\raggedleft\let\newline\\\arraybackslash\hspace{0pt}}m{#1}}

\definecolor{grey50}{rgb}{0.5,0.5,0.5}

\newcommand{\greencheck}{{\color{Green4} \checkmark}}
\newcommand{\redcross}{{\color{red}$\times$}}

\newcommand{\minitab}[2][l]{\begin{tabular}{#1}#2\end{tabular}}

\newenvironment{myitemize}[1][]{
	\begin{list}{{#1}} 
		{
			\setlength{\leftmargin}{1.2em}
			\setlength{\topsep}{0em}
			\setlength{\itemsep}{-0.2em}
	}}
{\end{list}}

\makeatletter
\def\and{
  \end{tabular}%
  \hskip -0.1em \@plus.17fil%
  \begin{tabular}[t]{c}}
\makeatother

\let\appendixpagenameorig\appendixpagename
\renewcommand{\appendixpagename}{\Large\appendixpagenameorig}

\def\cvprPaperID{387} 
\def\confYear{CVPR 2021}

\begin{document}

\title{DexYCB: A Benchmark for Capturing Hand Grasping of Objects}

\author{Yu-Wei Chao$^1$ \and Wei Yang$^1$ \and Yu Xiang$^1$ \and Pavlo Molchanov$^1$ \and Ankur Handa$^1$ \and Jonathan Tremblay$^1$ \and Yashraj S. Narang$^1$ \and Karl Van Wyk$^1$ \and Umar Iqbal$^1$ \and Stan Birchfield$^1$ \and Jan Kautz$^1$ \and Dieter Fox$^{1,2}$
\vspace{1mm}
\and
$^1$NVIDIA, $^2$University of Washington \vspace{-1mm} \\ \vspace{-3mm}
\makebox[0cm]{\tt\footnotesize \{ychao,weiy,yux,pmolchanov,ahanda,jtremblay,ynarang,kvanwyk,uiqbal,sbirchfield,jkautz,dieterf\}@nvidia.com}
}

\twocolumn[{
\maketitle
\begin{center}
 \centering
 \includegraphics[width=0.157\linewidth]{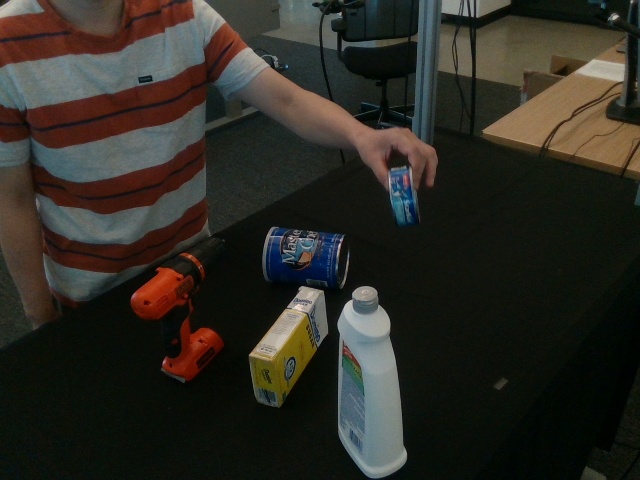}~
 \includegraphics[width=0.157\linewidth]{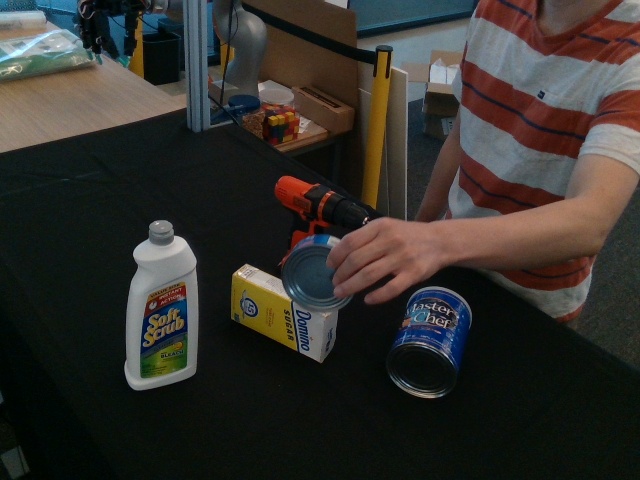}~
 \includegraphics[width=0.157\linewidth]{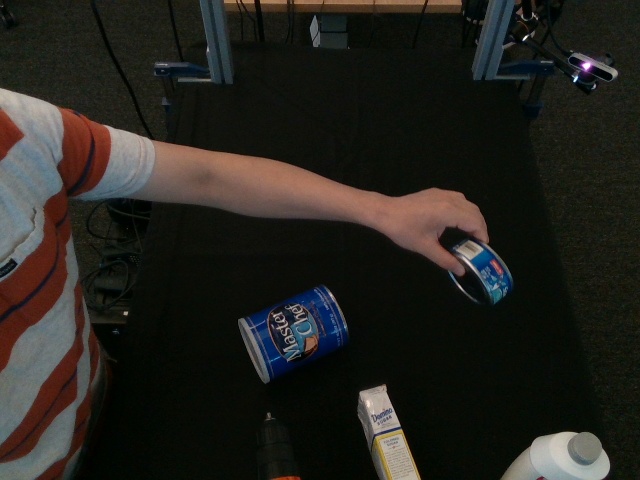}~~~
 \includegraphics[width=0.157\linewidth]{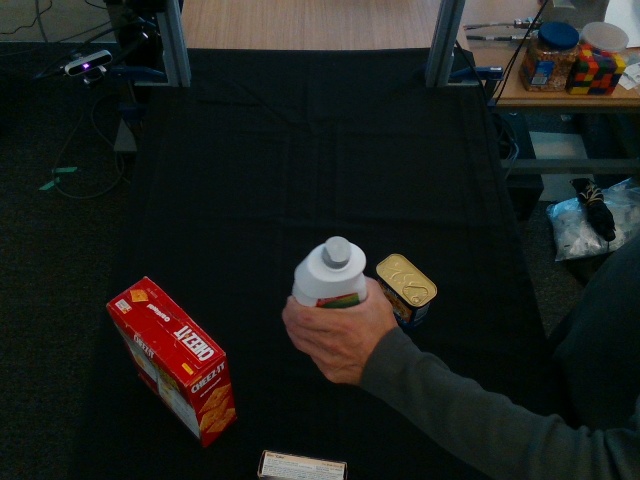}~
 \includegraphics[width=0.157\linewidth]{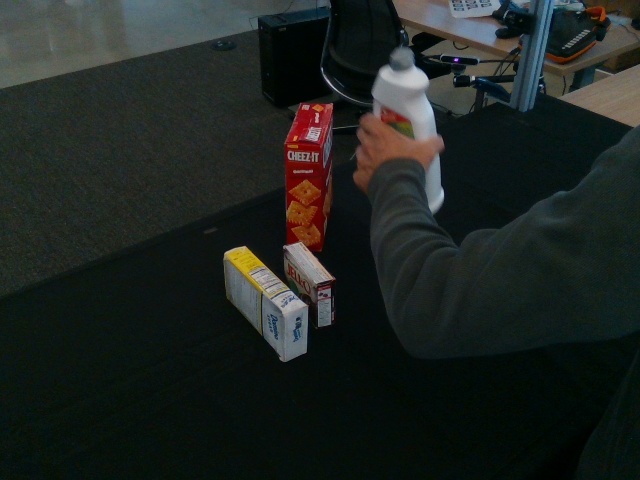}~
 \includegraphics[width=0.157\linewidth]{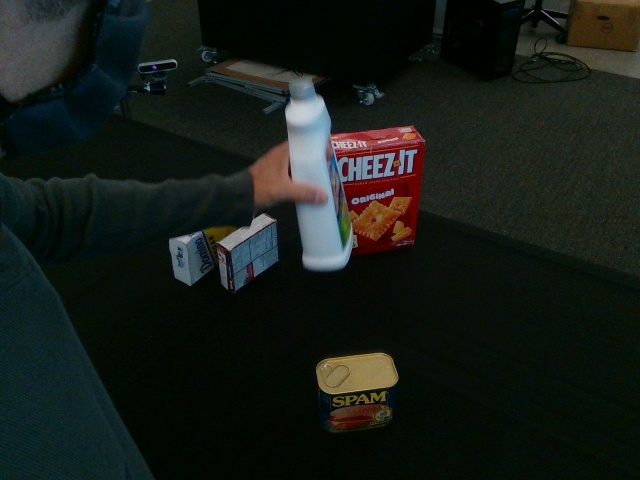}
 \\ \vspace{1mm}
 \includegraphics[width=0.157\linewidth]{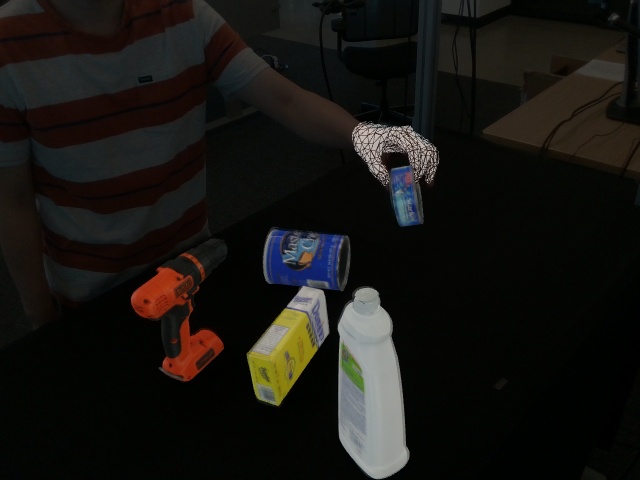}~
 \includegraphics[width=0.157\linewidth]{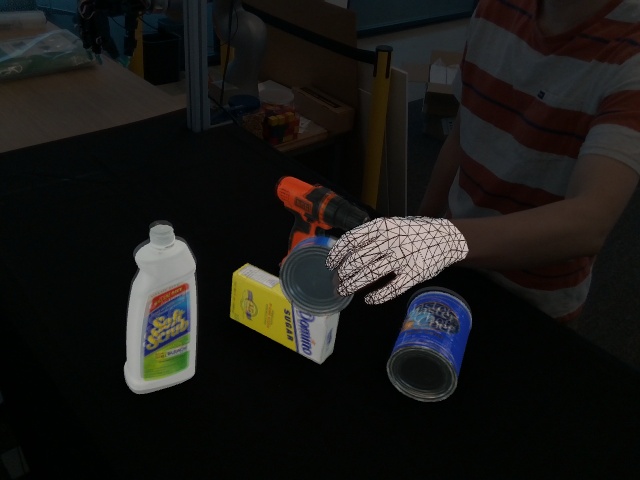}~
 \includegraphics[width=0.157\linewidth]{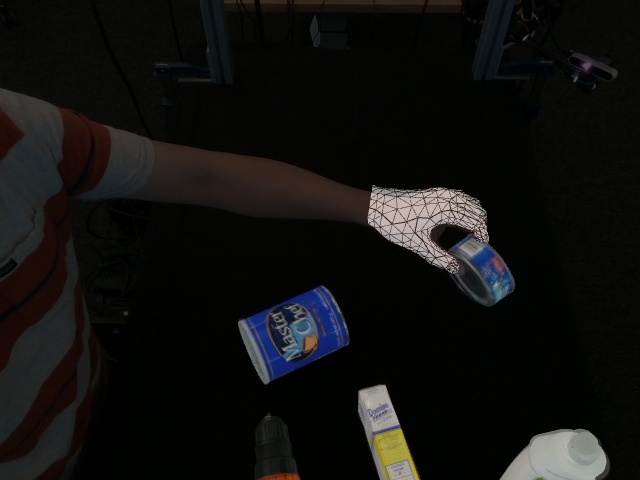}~~~
 \includegraphics[width=0.157\linewidth]{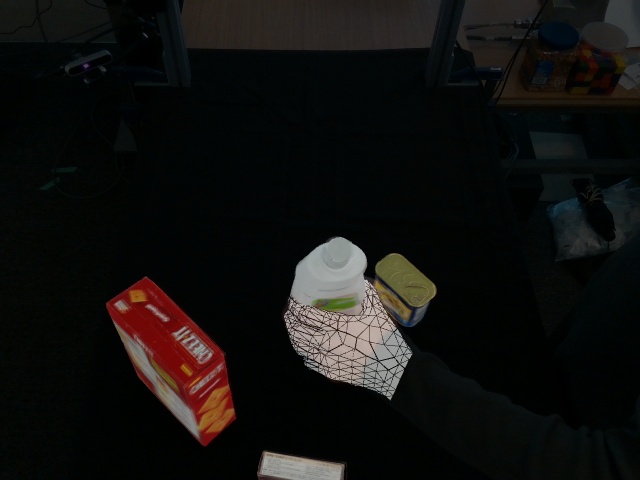}~
 \includegraphics[width=0.157\linewidth]{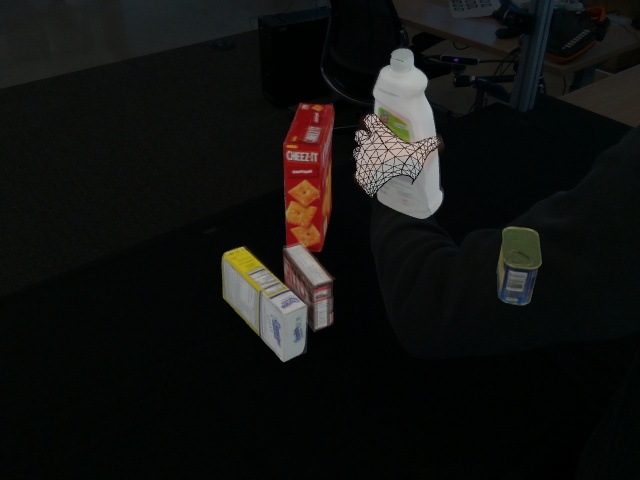}~
 \includegraphics[width=0.157\linewidth]{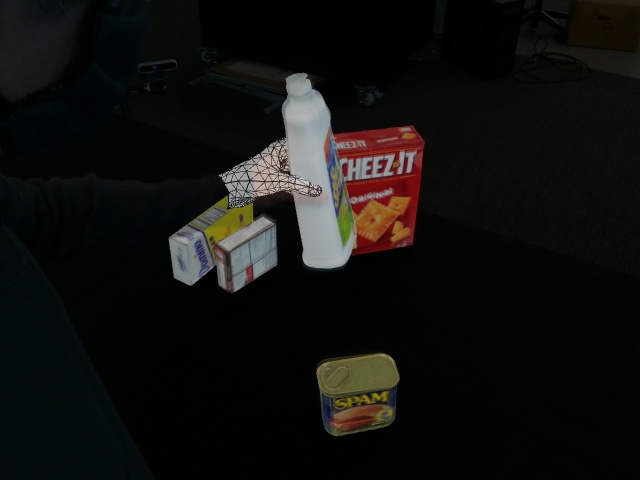}
 \captionof{figure}{\small Two captures (left and right) from the DexYCB
dataset. In each case, the top row shows color images simultaneously captured
from three views, while the bottom row shows the ground-truth 3D object and
hand pose rendered on the darkened captured images.}
 \label{fig:dexycb}
\end{center}
}]

\etocdepthtag.toc{mtchapter}

\begin{abstract}
We introduce DexYCB, a new dataset for capturing hand grasping of objects. We
first compare DexYCB with a related one through cross-dataset evaluation. We
then present a thorough benchmark of state-of-the-art approaches on three
relevant tasks: 2D object and keypoint detection, 6D object pose estimation,
and 3D hand pose estimation. Finally, we evaluate a new robotics-relevant task:
generating safe robot grasps in human-to-robot object
handover.~\footnote{Dataset and code available at
\href{https://dex-ycb.github.io}{\texttt{https://dex-ycb.github.io}}.}
\end{abstract}

\section{Introduction}

3D object pose estimation and 3D hand pose estimation are two important yet
unsolved vision problems. Traditionally, these two problems have been addressed
separately, yet in many critical applications, we need both capabilities
working together~\cite{tekin:cvpr2019,doosti:cvpr2020,hasson:cvpr2020}. For
example, in robotics, a reliable motion capture for hand manipulation of
objects is crucial for both learning from human
demonstration~\cite{handa:icra2020} and fluent and safe human-robot
interaction~\cite{yang:iros2020}.

State-of-the-art approaches for both 3D object
pose~\cite{tekin:cvpr2018,li:eccv2018,sundermeyer:eccv2018,peng:cvpr2019,wang:cvpr2019a,park:iccv2019,labbe:eccv2020}
and 3D hand pose
estimation~\cite{zimmermann:iccv2017,mueller:cvpr2018,iqbal:eccv2018,boukhayma:cvpr2019,ge:cvpr2019,hasson:cvpr2019,spurr:eccv2020}
rely on deep learning and thus require large datasets with labeled hand or
object poses for training. Many
datasets~\cite{yuan:cvpr2017,zimmermann:iccv2017,hodan:eccv2018,hodan:eccvw2020}
have been introduced in both domains and have facilitated progress on these two
problems in parallel. However, since they were introduced for either task
separately, many of them do not contain interaction of hands and objects, i.e.,
static objects without humans in the scene, or bare hands without interacting
with objects. In the presence of interactions, the challenge of solving the two
tasks together not only doubles but multiplies, due to the motion of
objects and mutual occlusions incurred by the interaction. Networks trained on
either of the datasets will thus not generalize well to interaction scenarios.

Creating a dataset with accurate 3D pose of hands and objects is also
challenging for the same reasons. As a result, prior works have attempted to
capture accurate hand motion either with specialized
gloves~\cite{glauser:siggraph2019}, magnetic
sensors~\cite{yuan:cvpr2017,garcia-hernando:cvpr2018}, or marker-based mocap
systems~\cite{brahmbhatt:eccv2020,taheri:eccv2020}. While they can achieve
unparalleled accuracy, the introduction of hand-attached devices may be
intrusive and thus bias the naturalness of hand motion. It also changes the
appearance of hands and thus may cause issues with generalization.

Due to the challenge of acquiring real 3D poses, there has been an increasing
interest in using synthetic datasets to train pose estimation models. The
success has been notable on object pose estimation. Using 3D scanned object
models and photorealistic rendering, prior
work~\cite{hodan:eccv2018,sundermeyer:eccv2018,tremblay:corl2018,deng:rss2019,hodan:eccvw2020}
has generated synthetic scenes of objects with high fidelity in appearance.
Their models trained only on synthetic data can thus translate to real images.
Nonetheless, synthesizing hand-object interactions remains challenging. One
problem is to synthesize realistic grasp poses for generic
objects~\cite{corona:cvpr2020}. Furthermore, synthesizing natural looking human
motions is still an active research area in graphics.

In this paper, we focus on marker-less data collection of real hand interaction
with objects. We take inspiration from recent work~\cite{hampali:cvpr2020} and
build a multi-camera setup that records interactions synchronously from
multiple views. Compared to the recent work, we instrument the setup with more
cameras and configure them to capture a larger workspace that allows our human
subjects to interact freely with objects. In addition, our pose labeling
process utilizes human annotation rather than automatic labeling. We
crowdsource the annotation so that we can efficiently scale up the data
labeling process. Given the setup, we construct a large-scale dataset that
captures a simple yet ubiquitous task: grasping objects from a table. The
dataset, DexYCB, consists of 582K RGB-D frames over 1,000 sequences of 10
subjects grasping 20 different objects from 8 views (Fig.~\ref{fig:dexycb}).

Our contributions are threefold. First, we introduce a new dataset for
capturing hand grasping of objects. We empirically demonstrate the strength of
our dataset over a related one through cross-dataset evaluation. Second, we
provide in-depth analysis of current approaches thoroughly on three relevant
tasks: 2D object and keypoint detection, 6D object pose estimation, and 3D hand
pose estimation. To the best of our knowledge, our dataset is the first that
allows joint evaluation of these three tasks. Finally, we demonstrate the
importance of joint hand and object pose estimation on a new robotics relevant
task: generating safe robot grasps for human-to-robot object
handover.

\section{Constructing DexYCB}

\subsection{Hardware Setup}

\begin{figure}[t]
 \centering
 \includegraphics[width=0.96\linewidth]{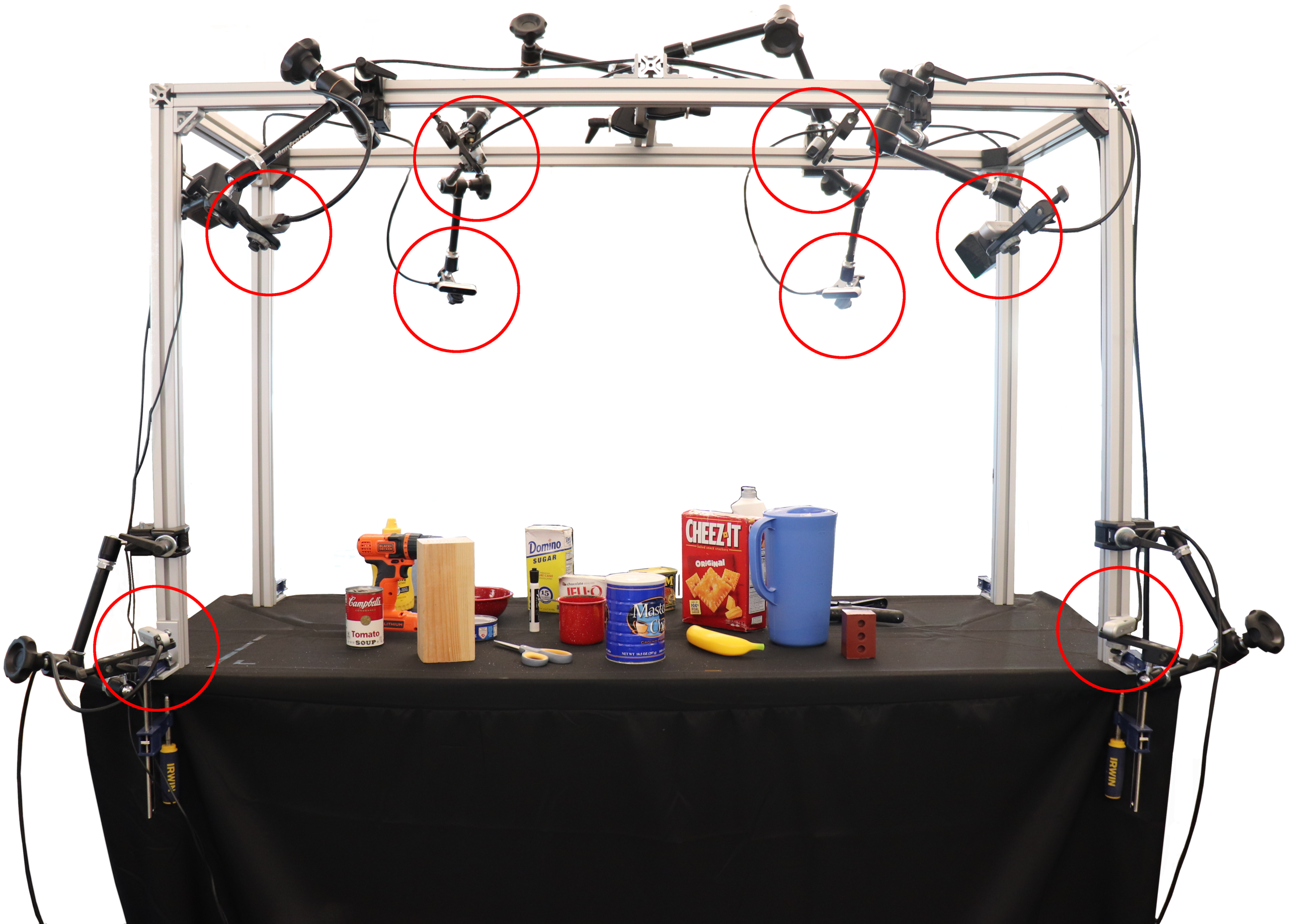}
 \caption{\small Our setup with 8 RGB-D cameras (red circle).}
 \vspace{-2mm}
 \label{fig:hardware}
\end{figure}

In order to construct the dataset, we built a multi-camera setup for capturing
human hands interacting with objects. 
A key design choice was to enable a sizable capture space, where a human
subject can freely interact and perform tasks with multiple objects. Our
multi-camera setup is shown in Fig.~\ref{fig:hardware}. We use 8 RGB-D cameras
(RealSense D415) and mount them such that collectively they can capture a
tabletop workspace with minimal blind spots. The cameras are extrinsically
calibrated and temporally synchronized. For data collection, we stream and
record all 8 views together at 30 fps with both color and depth of resolution
$640\times480$.

\subsection{Data Collection and Annotation}

Given the setup, we record videos of hands grasping objects. We use 20 objects
from the YCB-Video dataset~\cite{xiang:rss2018}, and record multiple trials
from 10 subjects. For each trial, we select a target object with 2 to
4 other objects and place them on the table. We ask the subject to start from
a relaxed pose, pick up the target object, and hold it in the air.
We also ask some subjects to pretend to hand over the object to someone across from them. We
record for 3 seconds, which is sufficient to contain the full course of action.
For each target object, we repeat the trial 5 times, each time with a random
set of accompanied objects and placement. We ask the subject to perform the
pick-up with the right hand in the first two trials, and with the left hand in
the third and fourth trials. In the fifth trial, we randomize the choice. We
rotate the target among all 20 objects. This gives us 100 trials per subject,
and 1,000 trials in total for all subjects.

To acquire accurate ground-truth 3D pose for hands and objects, our approach
(detailed in Sec.~\ref{sec:solve}) relies on 2D keypoint annotations for hands
and objects in each view. To ensure accuracy, we label the required keypoints in
RGB sequences fully through human annotation. Our annotation tool is based on
VATIC~\cite{vondrick:ijcv2013} for efficient annotation of videos. We set up
annotation tasks on the Amazon Mechanical Turk (MTurk) and label every view in all
the sequences.

For hands, we adopt 21 pre-defined hand joints as our keypoints (3 joints plus
1 tip for each finger and the wrist). We explicitly ask the annotators to label
and track these joints throughout a given video sequence. The annotators are
also asked to mark a keypoint as invisible in a given frame when it is
occluded.

Pre-defining keypoints exhaustively for every object would be laborious
and does not scale as the number of objects increases. Our approach
(Sec.~\ref{sec:solve}) explicitly addresses this issue by allowing user-defined
keypoints. Specifically, given a video sequence in a particular view, we first
ask the annotator to find 2 distinctive landmark points that are easily
identified and trackable on a designated object, and we ask them to label and
track these points throughout the sequence. We explicitly ask the annotators to
find keypoints that are visible most of the time, and mark a
keypoint as invisible whenever it is occluded.

\subsection{Solving 3D Hand and Object Pose}
\label{sec:solve}

To represent 3D hand pose, we use the popular MANO hand
model~\cite{romero:siggraphasia2017}. The model represents a right or left hand
with a deformable triangular mesh of 778 vertices. The mesh is parameterized by
two low-dimensional embeddings $(\theta,\beta)$, where
$\theta\in\mathbb{R}^{51}$ accounts for variations in pose (i.e. articulation)
and $\beta\in\mathbb{R}^{10}$ in shape. We use the version
from~\cite{hasson:cvpr2019}, which implements MANO as a differentiable layer in
PyTorch that maps $(\theta,\beta)$ to the mesh together with the 3D positions
of 21 hand joints defined in the keypoint annotation. We pre-calibrate the hand
shape $\beta$ for each subject and fix it throughout each subject's sequences.

Since our objects from YCB-Video~\cite{xiang:rss2018} also come with texture-mapped 3D mesh models, we use the standard 6D pose
representation~\cite{hodan:eccv2018,hodan:eccvw2020} for 3D object pose. The pose of each
object is represented by a matrix $T\in\mathbb{R}^{3\times4}$ composed of a
3D rotation matrix and a 3D translation vector.

To solve for hand and object pose, we formulate an optimization problem similar
to~\cite{zimmermann:iccv2019,hampali:cvpr2020} by leveraging depth and keypoint
annotations from all views and multi-view geometry. For a given sequence with
$N_{H}$ hands and $N_{O}$ objects, we denote the overall pose at a given time
frame by $P=(P_{H},P_{O})$, where $P_{H}=\{\theta_{h}\}_{h=1}^{N_{H}}$ and
$P_{O}=\{T_{o}\}_{o=1}^{N_{O}}$. We define the pose in world coordinates
where we know the extrinsics of each camera. Then at each time frame, we solve
the pose by minimizing the following energy function:
\begin{equation}
 \label{eq:energy}
 E(P)=E_{\text{depth}}(P)+E_{\text{kpt}}(P)+E_{\text{reg}}(P).
\end{equation}

\vspace{-3mm}
\paragraph{Depth} The depth term $E_{\text{depth}}$ measures how well the
models given poses explain the observed depth data. Let
$\{d_{i}\in\mathbb{R}^{3}\}_{i=1}^{N_{D}}$ be the total point cloud merged from
all views after transforming to the world coordinates, with $N_{D}$ denoting the
number of points. Given a pose parameter, we denote the collection of all hand
and object meshes as
$\mathcal{M}(P)=(\{\mathcal{M}_{h}(\theta_{h})\},\{\mathcal{M}_{o}(T_{o})\})$.
We define the depth term as
\begin{equation}
 E_{\text{depth}}(P)=\frac{1}{N_{D}}\sum_{i=1}^{N_{D}}|\text{SDF}(d_{i},\mathcal{M}(P))|^{2},
\end{equation}
where $\text{SDF}(\cdot)$ calculates the signed distance value of a 3D point
from a triangular mesh in mm. While $E_{\text{depth}}$ is differentiable,
calculating $E_{\text{depth}}$ and also the gradients is computationally
expensive for large point clouds and meshes with a huge number of vertices.
Therefore, we use an efficient point-parallel GPU implementation for it.

\vspace{-3mm}
\paragraph{Keypoint} The keypoint term $E_{\text{kpt}}$ measures the
reprojection error of the keypoints on the models with the annotated keypoints,
and can be decomposed by hand and object:
\begin{equation}
 E_{\text{kpt}}(P)=E_{\text{kpt}}(P_{H})+E_{\text{kpt}}(P_{O}).
\end{equation}

For hands, let $J_{h,j}$ be the 3D position of joint $j$ of hand $h$ in the world
coordinates, $p_{h,j}^{c}$ be the annotation of the same joint in the image
coordinates of view $c$, and $\gamma_{h,j}^{c}$ be its visibility indicator.
The energy term is defined as
\begin{equation}
 \label{eq:hand}
 E_{\text{kpt}}(P_{H})=\frac{1}{\sum\gamma_{h,j}^{c}}\sum_{c=1}^{N_{C}}\sum_{h=1}^{N_{H}}\sum_{j=1}^{N_{J}}\gamma_{h,j}^{c}||\text{proj}^{c}(J_{h,j})-p_{h,j}^{c}||_{2}^{2},
\end{equation}
where $\text{proj}^{c}(\cdot)$ returns the projection of a 3D point onto the
image plane of view $c$, and $N_{C}=8$ and $N_{J}=21$.

For objects, recall that we did not pre-define keypoints for annotation, but
rather asked annotators to select distinctive points to track. Here, we assume an
accurate initial pose is given at the first frame where an object's
keypoint is labeled visible. We then map the selected keypoint to a vertex on the
object's 3D model by back-projecting the keypoint's position onto the object's
visible surface. We fix that mapping afterwards. Let $K_{o,k}^{c}$ be the 3D
position of the selected keypoint $k$ of object $o$ in view $c$ in world
coordinates. Similar to Eq.~\eqref{eq:hand}, with $N_{K}=2$, the energy term is
\begin{equation}
 E_{\text{kpt}}(P_{O})=\frac{1}{\sum\gamma_{o,k}^{c}}\sum_{c=1}^{N_{C}}\sum_{o=1}^{N_{O}}\sum_{k=1}^{N_{K}}\gamma_{o,k}^{c}||\text{proj}^{c}(K_{o,k}^{c})-p_{o,k}^{c}||_{2}^{2}.
\end{equation}
To ensure an accurate initial pose for keypoint mapping, we initialize the pose
in each time frame with the solved pose from the last time frame. We initialize
the pose in the first frame by running PoseCNN~\cite{xiang:rss2018} on each
view and select an accurate pose for each object manually.

\vspace{-3mm}
\paragraph{Regularization} Following
\cite{mueller:siggraph2019,hasson:cvpr2020}, we add an $\ell_{2}$
regularization to the low-dimensional pose embedding of MANO to avoid irregular
articulation of hands:
\begin{equation}
 E_{\text{reg}}(P)=\frac{1}{N_{H}}\sum_{h=1}^{N_{H}}||\theta_{h}||_{2}^{2}.
\end{equation}
To minimize Eq.~\eqref{eq:energy}, we use the Adam optimizer with a learning rate
of 0.01. For each time frame, we initialize the pose $P$ with the solved pose
from the last time frame and run the optimizer for 100 iterations.

\section{Related Datasets}

\subsection{6D Object Pose}

Most datasets address instance-level 6D object pose estimation, where 3D models
are given a priori. The recent BOP
challenge~\cite{hodan:eccv2018,hodan:eccvw2020} has curated a decent line-up of
these datasets which the participants have to evaluate on. Yet objects are
mostly static in these datasets without human interactions. A recent
dataset~\cite{wang:cvpr2019b} was introduced for category-level 6D pose
estimation, but the scenes are also static.

\begin{table*}[t]
 \centering
 \footnotesize
 \setlength{\tabcolsep}{1.30pt}
 \begin{tabular}{l|ccccccccrrrrcrc|c}
  \multirow{3}{*}{dataset} & \multirow{3}{*}{\minitab[c]{visual\\modality}} & \multirow{3}{*}{\minitab[c]{real\\image}} & \multirow{3}{*}{\minitab[c]{marker-\\less}} & hand- & hand- & 3D   & 3D    & \multirow{3}{*}{resolution} & \multirow{3}{*}{\#frames} & \multirow{3}{*}{\#sub} & \multirow{3}{*}{\#obj} & \multirow{3}{*}{\#views} & \multirow{3}{*}{motion} & \multirow{3}{*}{\#seq} & \multirow{3}{*}{\minitab[c]{dynamic\\grasp}} & \multirow{3}{*}{label} \\
                           &                                                &                                           &                                             & hand  & obj   & obj  & hand  &                             &                           &                        &                        &                          &                         &                        &                                              &                        \\
                           &                                                &                                           &                                             & int   & int   & pose & shape &                             &                           &                        &                        &                          &                         &                        &                                              &                        \\
  \hline
  \multirow{2}{*}{BigHand2.2M~\cite{yuan:cvpr2017}} & \multirow{2}{*}{depth} & \multirow{2}{*}{\greencheck} & \multirow{2}{*}{\redcross} & \multirow{2}{*}{\redcross} & \multirow{2}{*}{\redcross} & \multirow{2}{*}{--} & \multirow{2}{*}{\redcross} & \multirow{2}{*}{640$\times$480} & \multirow{2}{*}{2,200K} & \multirow{2}{*}{10} & \multirow{2}{*}{--} & \multirow{2}{*}{1} & \multirow{2}{*}{\greencheck} & \multirow{2}{*}{--} & \multirow{2}{*}{--} & magnetic \\
                                                    &                        &                              &                            &                            &                            &                     &                            &                                 &                         &                     &                     &                    &                              &                     &                     & sensor   \\
  \hline
  SynthHands~\cite{mueller:iccv2017}            & RGB-D & \redcross   & --          & \redcross   & \greencheck & \redcross   & \redcross   & 640$\times$480   &   220K & -- &  7 &              5 & \redcross   & --    & --          & synthetic \\
  Rendered Hand Pose~\cite{zimmermann:iccv2017} & RGB-D & \redcross   & --          & \redcross   & \redcross   & --          & \redcross   & 320$\times$320   &    44K & 20 & -- &              1 & \redcross   & --    & --          & synthetic \\
  GANerated Hands~\cite{mueller:cvpr2018}       & RGB   & \redcross   & --          & \redcross   & \greencheck & \redcross   & \redcross   & 256$\times$256   &   331K & -- & -- &              1 & \redcross   & --    & --          & synthetic \\
  ObMan~\cite{hasson:cvpr2019}                  & RGB-D & \redcross   & --          & \redcross   & \greencheck & \greencheck & \greencheck & 256$\times$256   &   154K & 20 & 3K &              1 & \redcross   & --    & --          & synthetic \\
  \hline
  \multirow{2}{*}{FPHA~\cite{garcia-hernando:cvpr2018}}   & \multirow{2}{*}{RGB-D} & \multirow{2}{*}{\greencheck} & \multirow{2}{*}{\redcross} & \multirow{2}{*}{\redcross}   & \multirow{2}{*}{\greencheck} & \multirow{2}{*}{\greencheck} & \multirow{2}{*}{\redcross}   & \multirow{2}{*}{1920$\times$1080} & \multirow{2}{*}{105K}   & \multirow{2}{*}{6}  & \multirow{2}{*}{4}  & \multirow{2}{*}{1} & \multirow{2}{*}{\greencheck} & \multirow{2}{*}{1,175} & \multirow{2}{*}{\greencheck} & magnetic  \\
                                                          &                        &                              &                            &                              &                              &                              &                              &                                   &                         &                     &                     &                    &                              &                        &                              & sensor    \\
  \multirow{2}{*}{ContactPose~\cite{brahmbhatt:eccv2020}} & \multirow{2}{*}{RGB-D} & \multirow{2}{*}{\greencheck} & \multirow{2}{*}{\redcross} & \multirow{2}{*}{\redcross}   & \multirow{2}{*}{\greencheck} & \multirow{2}{*}{\greencheck} & \multirow{2}{*}{\greencheck} & \multirow{2}{*}{960$\times$540}   & \multirow{2}{*}{2,991K} & \multirow{2}{*}{50} & \multirow{2}{*}{25} & \multirow{2}{*}{3} & \multirow{2}{*}{\greencheck} & \multirow{2}{*}{2,303} & \multirow{2}{*}{\redcross}   & mocap     \\
                                                          &                        &                              &                            &                              &                              &                              &                              &                                   &                         &                     &                     &                    &                              &                        &                              & + thermal \\
  GRAB~\cite{taheri:eccv2020}                             & --                     & --                           & \redcross                  & \redcross                    & \greencheck                  & \greencheck                  & \greencheck                  & --                                &                 1 ,624K &                  10 &                  51 &                 -- & \greencheck                  &                  1,335 & \greencheck                  & mocap     \\
  \hline
  Mueller et al.~\cite{mueller:siggraph2019}    & depth & \redcross   & --          & \greencheck & \redcross   & --          & \redcross   & 640$\times$480   &    80K &  5 & -- &              4 & \greencheck &    11 & --          & synthetic \\
  InterHand2.6M~\cite{moon:eccv2020}            & RGB   & \greencheck & \greencheck & \greencheck & \redcross   & --          & \greencheck & 512$\times$334   & 2,590K & 27 & -- & \textgreater80 & \greencheck &    -- & --          & semi-auto \\
  \hline
  Dexter+Object~\cite{sridhar:eccv2016}         & RGB-D & \greencheck & \greencheck & \redcross   & \greencheck & \greencheck & \redcross   & 640$\times$480   &     3K &  2 &  2 &              1 & \greencheck &     6 & \greencheck & manual    \\
  Simon et al.~\cite{simon:cvpr2017}            & RGB   & \greencheck & \greencheck & \greencheck & \greencheck & \redcross   & \redcross   & 1920$\times$1080 &    15K & -- & -- &             31 & \greencheck &    -- & \greencheck & automatic \\
  EgoDexter~\cite{mueller:iccv2017}             & RGB-D & \greencheck & \greencheck & \redcross   & \greencheck & \redcross   & \redcross   & 640$\times$480   &     3K &  4 & -- &              1 & \greencheck &     4 & \greencheck & manual    \\
  FreiHAND~\cite{zimmermann:iccv2019}           & RGB   & \greencheck & \greencheck & \redcross   & \greencheck & \redcross   & \greencheck & 224$\times$224   &    37K & 32 & 27 &              8 & \redcross   &    -- & --          & semi-auto \\
  HO-3D~\cite{hampali:cvpr2020}                 & RGB-D & \greencheck & \greencheck & \redcross   & \greencheck & \greencheck & \greencheck & 640$\times$480   &    78K & 10 & 10 &           1--5 & \greencheck &    27 & \greencheck & automatic \\
  DexYCB (ours)                                 & RGB-D & \greencheck & \greencheck & \redcross   & \greencheck & \greencheck & \greencheck & 640$\times$480   &   582K & 10 & 20 &              8 & \greencheck & 1,000 & \greencheck & manual
 \end{tabular}
 \vspace{-2mm}
 \caption{\small Comparison of DexYCB with existing 3D hand pose datasets.}
 \vspace{-3mm}
 \label{tab:dataset}
\end{table*}

\subsection{3D Hand Pose}

We present a summary of related 3D hand pose datasets in
Tab.~\ref{tab:dataset}. Some address pose estimation with depth only, e.g.,
BigHand2.2M~\cite{yuan:cvpr2017}. These datasets can be large but lack color
images and capture only bare hands without interactions. Some others address
hand-hand interactions, e.g., Mueller et al.~\cite{mueller:siggraph2019} and
InterHand2.6M~\cite{moon:eccv2020}. While not focused on interaction with
objects, their datasets address another challenging scenario and is orthogonal
to our work. Below we review datasets with hand-object interactions.

\vspace{-3mm}
\paragraph{Synthetic} Synthetic data has been increasingly used for hand pose
estimation~\cite{mueller:iccv2017,zimmermann:iccv2017,mueller:cvpr2018,hasson:cvpr2019}.
A common downside is the gap to real images on appearance. To bridge this gap,
GANerated Hands~\cite{mueller:cvpr2018} was introduced by translating synthetic
images to real via GANs. Nonetheless, other challenges remain. One is to
synthesize realistic grasp pose for objects. ObMan~\cite{hasson:cvpr2019} was
introduced to address this challenge using heuristic metrics. Yet besides pose,
it is also challenging to synthesize realistic motion. Consequently, these
datasets only offer static images but not videos. Our dataset captures real
videos with real grasp pose and motion. Furthermore, our motion data from real
can help bootstrapping synthetic data generation.

\vspace{-3mm}
\paragraph{Marker-based} FPHA~\cite{garcia-hernando:cvpr2018} captured hand
interaction with objects in first person view by attaching magnetic sensors to
the hands. This offers a flexible and portable solution, but the attached
sensors may hinder natural hand motion and also bias the hand appearance.
ContactPose~\cite{brahmbhatt:eccv2020} captured grasp poses by tracking objects
with mocap markers and recovering hand pose from thermal imagery of the objects
surface. While the dataset is large in scale, the thermal based approach can
only capturing rigid grasp poses but not motions. GRAB~\cite{taheri:eccv2020}
captured full body motion together with hand-object interactions using
marker-based mocap. It provides high-fidelity capture of interactions but does
not come with any visual modalities. Our dataset is marker-less, captures
dynamic grasp motions, and provides RGB-D sequences in multiple views.

\vspace{-3mm}
\paragraph{Marker-less} Our DexYCB dataset falls in this category. The
challenge is to acquire accurate 3D pose. Some datasets like
Dexter+Object~\cite{sridhar:eccv2016} and EgoDexter~\cite{mueller:cvpr2018}
rely on manual annotation and are thus limited in scale. To acquire 3D pose at
scale others rely on automatic or semi-automatic
approaches~\cite{simon:cvpr2017,zimmermann:iccv2019}. While these datasets
capture hand-object interactions, they do not offer 3D pose for objects.

Most similar to ours is the recent HO-3D dataset~\cite{hampali:cvpr2020}. HO-3D
also captures hands interacting with objects from multiple views and provides
both 3D hand and object pose. Nonetheless, the two datasets differ both
quantitatively and qualitatively. Quantitatively (Tab.~\ref{tab:dataset}),
DexYCB captures interactions with more objects (20 versus 10), from more views
(8 versus 1 to 5), and is one order of magnitude larger in terms of number of
frames (582K versus 78K)~\footnote{We count the number of frames over all
views.} and number of sequences (1,000 versus 27).~\footnote{We count the same
motion across different views as one sequence.} Qualitatively
(Fig.~\ref{fig:ho3d}), we highlight three differences. First, DexYCB captures
full grasping processes (i.e. from hand approaching, opening fingers, contact,
to holding the object stably) in short segments, whereas HO-3D captures longer
sequences with object in hand most of the time. Among the 27 sequences from
HO-3D, we found 17 with hand always rigidly attached to the object, only
rotating the object from the wrist with no finger articulation. Second, DexYCB
captures a full tabletop workspace with 3D pose of all the objects on the table,
whereas in HO-3D one object is held close to the camera each time and only the
held object is labeled with pose. Finally, regarding annotation, DexYCB
leverages keypoints fully labeled by humans through crowdsourcing, while HO-3D
relies on a fully automatic labeling pipeline.

\begin{figure}[t]
 \centering
 \includegraphics[width=0.32\linewidth]{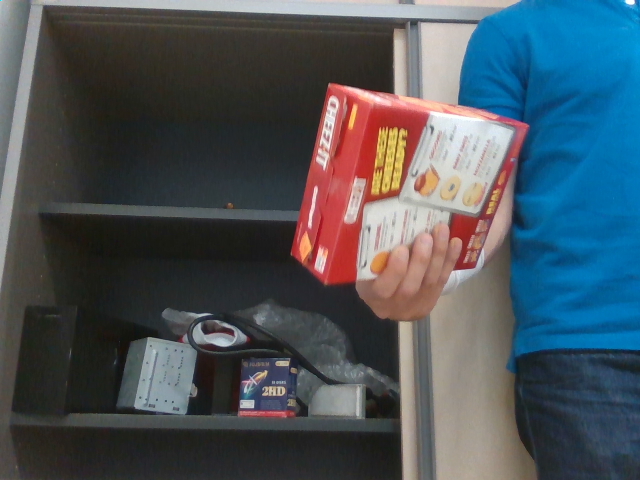}~
 \includegraphics[width=0.32\linewidth]{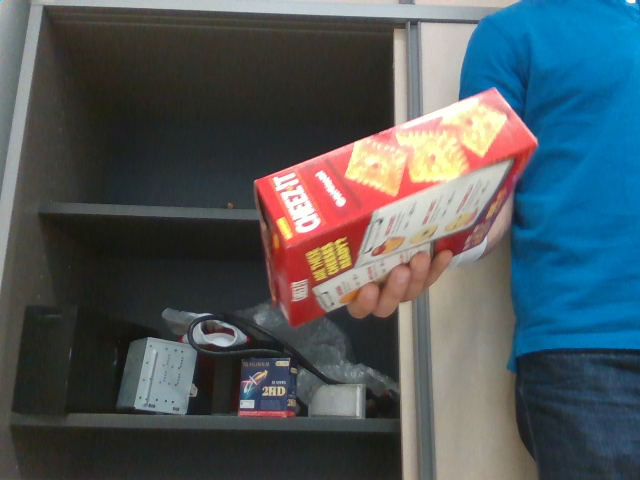}~
 \includegraphics[width=0.32\linewidth]{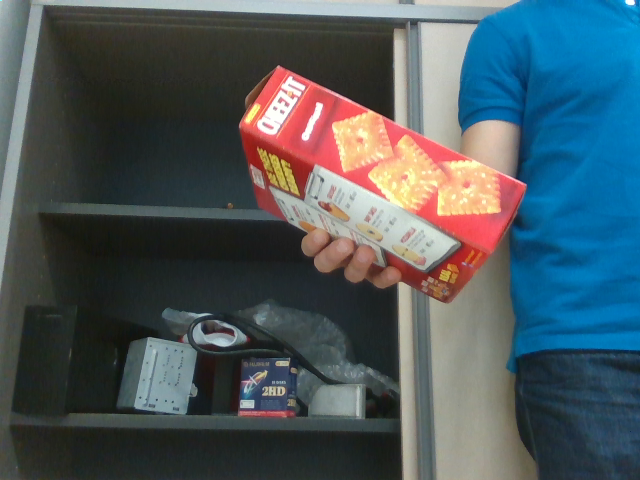}
 \\ \vspace{1mm}
 \includegraphics[width=0.32\linewidth]{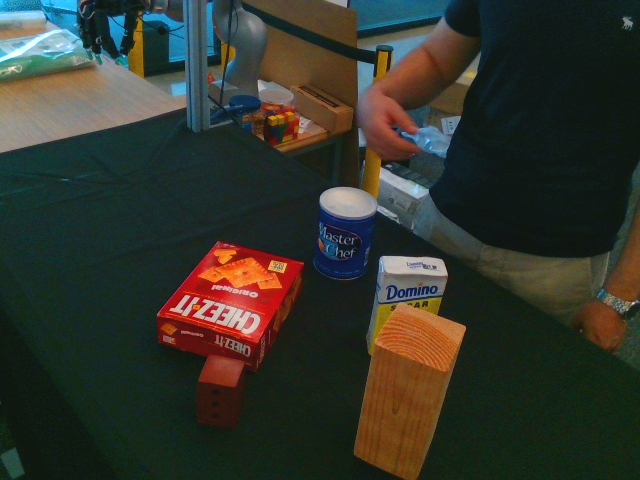}~
 \includegraphics[width=0.32\linewidth]{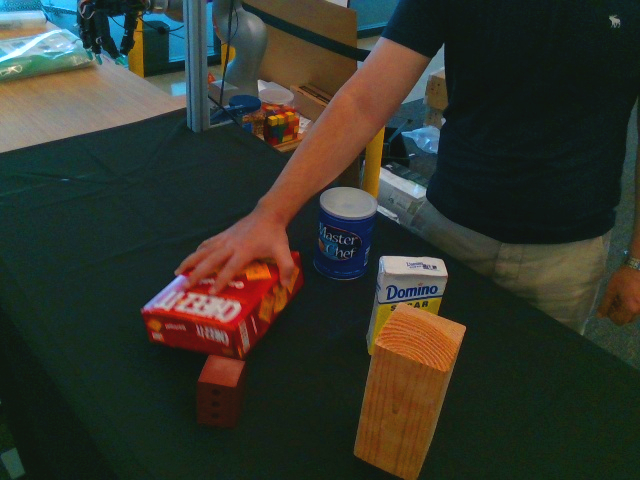}~
 \includegraphics[width=0.32\linewidth]{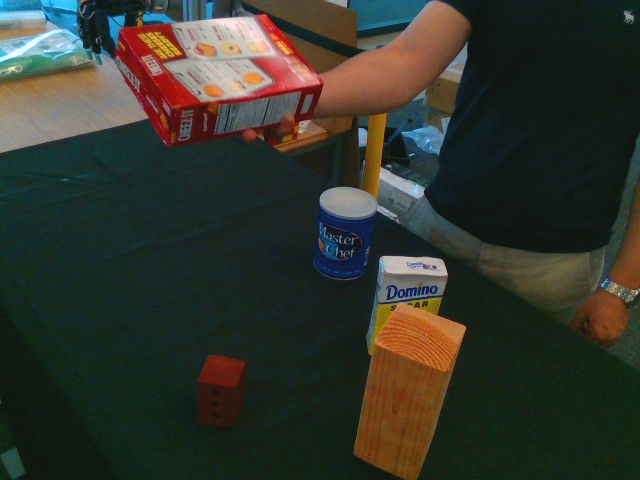}
 \caption{\small Qualitative comparison of HO-3D~\cite{hampali:cvpr2020} (top)
and DexYCB (bottom). DexYCB captures full grasping processes in a tabletop
workspace with 3D pose of all the on table objects.}
 \vspace{-4mm}
 \label{fig:ho3d}
\end{figure}

\section{Cross-Dataset Evaluation}
\label{sec:cross}

To further assess the merit of our DexYCB dataset, we perform cross-dataset
evaluation following prior
work~\cite{zimmermann:iccv2019,shan:cvpr2020,yu:cvpr2020}. We focus
specifically on HO-3D~\cite{hampali:cvpr2020} due to its relevance, and
evaluate generalizability between HO-3D and DexYCB on single-image 3D hand pose
estimation. For DexYCB, we generate a train/val/test split following our
benchmark setup (``S0'' in Sec.~\ref{sec:eval}). For HO-3D, we select 6 out of
its 55 training sequences for test. We use Spurr et al.~\cite{spurr:eccv2020}
(winner of the HAND 2019 Challenge) as our method, and train the model
separately on three training sets: the training set of HO-3D, the training set
of DexYCB, and the combined training set from both datasets. Finally, we
evaluate the three trained models separately on the test set of HO-3D and
DexYCB.

Tab.~\ref{tab:cross} shows the results in mean per joint position error (MPJPE)
reported on two different alignment methods (details in Sec.~\ref{sec:3d}). We
experiment with two different backbones: ResNet50 and
HRNet32~\cite{sun:cvpr2019}. Unsurprisingly, when training on a single dataset,
the model generalizes better to the respective test set. The error increases
when tested on the other dataset. When we evaluate the DexYCB trained model on
HO-3D, we observe a consistent increase from 1.4$\times$ to 1.9$\times$ (e.g.,
for ResNet50, from 18.05 to 31.76 mm on root-relative). However, when we
evaluate the HO-3D trained model on DexYCB, we observe a consistent yet more
significant increase, from 3.4$\times$ to 3.7$\times$ (e.g., for ResNet50, from
12.97 to 48.30 mm on root-relative). This suggests that models trained on
DexYCB generalize better than on HO-3D. Furthermore, when we train on the
combined training set and evaluate on HO-3D, we can further reduce the error
from only training on HO-3D (e.g., from 18.05 to 15.79 mm). However, when tested
on DexYCB, the error rather rises compared to training only on DexYCB (e.g.,
from 12.97 to 13.36 mm). We conclude that DexYCB complements HO-3D better than
vice versa.

\begin{table}[t]
 \centering
 \small
 \begin{tabular}{l|cc|c}
  \multirow{2}{*}{\backslashbox{test}{train}} & \multirow{2}{*}{HO-3D~\cite{hampali:cvpr2020}} & \multirow{2}{*}{DexYCB} & HO-3D~\cite{hampali:cvpr2020} \\
                                              &                                                &                         & + DexYCB                      \\
  \hline
  HO-3D                                       & 18.05 / 10.66 & 31.76 / 15.23  & 15.79 / 9.51 \\
  DexYCB                                      & 48.30 / 24.23 & 12.97 / ~~7.18 & 13.36 / 7.27 \\
  \hline
  \hline
  HO-3D                                       & 17.46 / 10.44 & 33.11 / 15.51  & 15.89 / 9.00 \\
  DexYCB                                      & 46.38 / 23.94 & 12.39 / ~~6.79 & 12.48 / 6.87
 \end{tabular}
 \vspace{-2mm}
 \caption{\small Cross-dataset evaluation with HO-3D~\cite{hampali:cvpr2020} on 3D hand pose estimation. Results are in MPJPE (mm) (root-relative / Procrustes aligned). Top: \cite{spurr:eccv2020} + ResNet50. Bottom: \cite{spurr:eccv2020} + HRNet32.}
 \label{tab:cross}
\end{table}

\section{Benchmarking Representative Approaches}
\label{sec:benchmark}

We benchmark three tasks on our DexYCB dataset: 2D object and keypoint
detection, 6D object pose estimation, and 3D hand pose estimation. For each
task we select representative approaches and analyze their performace.

\subsection{Evaluation Setup}
\label{sec:eval}

To evaluate different scenarios, we generate train/val/test splits in four
different ways (referred to as ``setup''):

\begin{myitemize}
 \item[$\bullet$] \textbf{S0 (default).} The train split contains all 10
subjects, all 8 camera views, and all 20 grasped objects. Only the sequences
are not shared with the val/test split.
 \item[$\bullet$] \textbf{S1 (unseen subjects).} The dataset is split by
subjects (train/val/test: 7/1/2).
 \item[$\bullet$] \textbf{S2 (unseen views).} The dataset is split by camera
views (train/val/test: 6/1/1).
 \item[$\bullet$] \textbf{S3 (unseen grasping).} The dataset is split by
grasped objects (train/val/test: 15/2/3). Objects being grasped in the test
split are never being grasped in the train/val split, but may appear static on
the table. This way the training set still contain examples of each object.
\end{myitemize}


\subsection{2D Object and Keypoint Detection}
\label{sec:2d}

We evaluate object and keypoint detection using the COCO evaluation
protocol~\cite{lin:eccv2014}. For object detection, we consider 20 object
classes and 1 hand class. We collect ground truths by rendering a segmentation
mask for each instance in each camera view. For keypoints, we consider 21 hand
joints and collect the ground truths by reprojecting each 3D joint to each
camera image.

We benchmark two representative approaches: Mask R-CNN
(Detectron2)~\cite{he:iccv2017,wu2019detectron2} and
SOLOv2~\cite{wang:neurips2020}. Mask R-CNN has been the de facto for object and
keypoint detection and SOLOv2 is a state-of-the-art on COCO instance
segmentation. For both we use a ResNet50-FPN backbone pre-trained on ImageNet
and finetune on DexYCB.

\begin{table*}[t]
 \centering
 \footnotesize
 \setlength{\tabcolsep}{2.06pt}
 \begin{tabular}{L{2.54cm}|C{0.77cm}C{0.77cm}|C{0.77cm}C{0.77cm}|C{0.77cm}C{0.77cm}|C{0.77cm}C{0.77cm}|C{0.77cm}C{0.77cm}|C{0.77cm}C{0.77cm}|C{0.77cm}C{0.77cm}|C{0.77cm}C{0.77cm}}
                           & \multicolumn{4}{c|}{S0 (default)} & \multicolumn{4}{c|}{S1 (unseen subjects)} & \multicolumn{4}{c|}{S2 (unseen views)} & \multicolumn{4}{c}{S3 (unseen grasping)} \\
                           & \multicolumn{2}{c|}{Mask R-CNN} & \multicolumn{2}{c|}{SOLOv2} & \multicolumn{2}{c|}{Mask R-CNN} & \multicolumn{2}{c|}{SOLOv2} & \multicolumn{2}{c|}{Mask R-CNN} & \multicolumn{2}{c|}{SOLOv2} & \multicolumn{2}{c|}{Mask R-CNN} & \multicolumn{2}{c}{SOLOv2} \\
                           & bbox & segm & bbox & segm & bbox & segm & bbox & segm & bbox & segm & bbox & segm & bbox & segm & bbox & segm \\
  \hline
  002\_master\_chef\_can   & 83.87          & 82.68          & \textbf{84.90} & \textbf{85.20} & 82.41          & 80.69          & \textbf{82.74} & \textbf{81.67} & \textbf{85.82} & \textbf{84.44} & 83.52          & 83.39          & 83.96          & 83.72          & \textbf{84.53} & \textbf{85.44} \\
  003\_cracker\_box        & 85.85          & 83.72          & \textbf{89.08} & \textbf{88.07} & 82.12          & 80.18          & \textbf{84.26} & \textbf{84.39} & 88.74          & 87.60          & \textbf{89.61} & \textbf{87.85} & \textbf{85.51} & 83.68          & 79.75          & \textbf{88.00} \\
  004\_sugar\_box          & \textbf{81.30} & 77.71          & 80.45          & \textbf{79.26} & \textbf{78.35} & 74.75          & 77.17          & \textbf{76.66} & \textbf{83.72} & \textbf{79.53} & 81.91          & 79.25          & \textbf{78.06} & 75.68          & 74.87          & \textbf{77.40} \\
  005\_tomato\_soup\_can   & \textbf{76.03} & 74.75          & 75.70          & \textbf{75.33} & \textbf{73.65} & 71.33          & 72.66          & \textbf{71.35} & \textbf{77.66} & \textbf{76.70} & 75.25          & 73.82          & \textbf{76.70} & 76.29          & 76.50          & \textbf{76.48} \\
  006\_mustard\_bottle     & \textbf{81.56} & 79.40          & 79.94          & \textbf{81.87} & \textbf{78.76} & 75.41          & 77.76          & \textbf{75.92} & \textbf{81.12} & \textbf{79.06} & 79.87          & 78.29          & 81.74          & 82.37          & \textbf{82.24} & \textbf{84.45} \\
  007\_tuna\_fish\_can     & \textbf{68.08} & 67.14          & 67.76          & \textbf{67.34} & \textbf{68.99} & \textbf{67.76} & 68.07          & 67.59          & \textbf{73.37} & \textbf{72.06} & 71.72          & 70.68          & \textbf{75.76} & \textbf{77.70} & \textbf{75.76} & 77.61          \\
  008\_pudding\_box        & 73.60          & 70.39          & \textbf{74.05} & \textbf{72.58} & 69.77          & 68.04          & \textbf{69.78} & \textbf{68.20} & 78.88          & \textbf{76.74} & \textbf{79.01} & 76.58          & \textbf{71.32} & \textbf{71.43} & 65.11          & 70.94          \\
  009\_gelatin\_box        & \textbf{69.51} & \textbf{68.43} & 68.64          & 68.09          & \textbf{67.37} & \textbf{64.99} & 66.16          & 63.14          & 72.87          & \textbf{71.45} & \textbf{73.34} & 69.36          & \textbf{61.19} & 58.41          & 60.98          & \textbf{59.49} \\
  010\_potted\_meat\_can   & 75.63          & 72.66          & \textbf{77.28} & \textbf{75.80} & \textbf{73.86} & \textbf{71.95} & 70.87          & 71.76          & 78.96          & \textbf{78.35} & \textbf{79.63} & 76.84          & 77.48          & 78.03          & \textbf{78.40} & \textbf{79.24} \\
  011\_banana              & \textbf{70.53} & 63.43          & 69.67          & \textbf{65.49} & \textbf{68.07} & 60.50          & 67.31          & \textbf{60.57} & \textbf{72.46} & 58.18          & 70.80          & \textbf{70.68} & 71.67          & 67.54          & \textbf{73.25} & \textbf{70.28} \\
  019\_pitcher\_base       & 87.17          & 84.67          & \textbf{89.95} & \textbf{88.98} & 86.21          & 82.00          & \textbf{90.07} & \textbf{86.38} & \textbf{90.47} & 86.67          & 86.88          & \textbf{87.02} & 85.62          & 83.69          & \textbf{90.22} & \textbf{89.91} \\
  021\_bleach\_cleanser    & \textbf{80.98} & 77.52          & 75.52          & \textbf{79.71} & \textbf{79.59} & 76.68          & 78.77          & \textbf{78.52} & \textbf{84.75} & 81.71          & 82.79          & \textbf{81.75} & \textbf{77.36} & 71.54          & 77.01          & \textbf{75.54} \\
  024\_bowl                & \textbf{80.62} & 78.12          & 75.79          & \textbf{80.30} & \textbf{78.75} & 76.42          & 78.06          & \textbf{77.39} & \textbf{83.25} & \textbf{80.12} & 81.54          & 79.72          & 81.12          & 79.69          & \textbf{81.91} & \textbf{82.99} \\
  025\_mug                 & 76.01          & 71.95          & \textbf{76.35} & \textbf{74.35} & \textbf{74.64} & 69.64          & 72.72          & \textbf{69.93} & \textbf{80.14} & \textbf{75.99} & 78.24          & 74.13          & 75.51          & 73.19          & \textbf{76.78} & \textbf{75.86} \\
  035\_power\_drill        & 81.83          & 73.80          & \textbf{82.85} & \textbf{77.20} & 76.89          & 69.02          & \textbf{77.90} & \textbf{71.50} & \textbf{82.57} & \textbf{74.66} & 82.42          & 74.09          & 83.67          & 77.62          & \textbf{84.06} & \textbf{80.77} \\
  036\_wood\_block         & 83.75          & 81.41          & \textbf{85.72} & \textbf{85.59} & \textbf{80.45} & 78.15          & 79.45          & \textbf{79.32} & \textbf{83.78} & \textbf{83.40} & 41.24          & 64.85          & 77.52          & 73.84          & \textbf{78.01} & \textbf{77.78} \\
  037\_scissors            & \textbf{64.07} & 29.36          & 59.00          & \textbf{32.63} & \textbf{55.05} & 20.68          & 50.85          & \textbf{26.51} & \textbf{63.16} & \textbf{17.93} & 38.00          & 16.93          & \textbf{58.01} & 24.35          & 57.92          & \textbf{31.11} \\
  040\_large\_marker       & \textbf{52.69} & 42.42          & 50.95          & \textbf{42.46} & \textbf{50.14} & \textbf{38.46} & 44.94          & 36.46          & \textbf{53.98} & \textbf{36.47} & 50.78          & 33.58          & \textbf{55.35} & 41.23          & 53.66          & \textbf{45.22} \\
  052\_extra\_large\_clamp & \textbf{73.41} & 54.03          & 71.71          & \textbf{55.38} & \textbf{68.12} & 51.26          & 65.26          & \textbf{52.33} & \textbf{75.61} & 52.77          & 61.14          & \textbf{53.59} & \textbf{75.24} & 57.28          & 72.52          & \textbf{58.87} \\
  061\_foam\_brick         & \textbf{72.68} & 72.85          & 72.54          & \textbf{72.93} & \textbf{68.34} & \textbf{66.62} & 67.98          & 65.72          & \textbf{75.24} & \textbf{74.10} & 73.01          & 71.05          & 71.36          & 71.32          & \textbf{71.50} & \textbf{71.59} \\
  hand                     & \textbf{71.85} & \textbf{54.83} & 66.41          & 54.27          & \textbf{64.88} & \textbf{46.86} & 58.33          & 46.66          & \textbf{70.23} & \textbf{54.70} & 59.28          & 52.08          & \textbf{71.23} & \textbf{54.89} & 66.33          & 53.51          \\
  \hline
  mAP                      & \textbf{75.76} & 69.58          & 75.13          & \textbf{71.56} & \textbf{72.69} & 66.26          & 71.48          & \textbf{67.24} & \textbf{77.94} & \textbf{70.60} & 72.37          & 68.66          & \textbf{75.02} & 69.69          & 74.35          & \textbf{72.05}
 \end{tabular}
 \vspace{-2mm}
 \caption{\small 2D object detection results in AP (\%) on the four setups. We
compare Mask R-CNN (Detectron2)~\cite{he:iccv2017,wu2019detectron2} with
SOLOv2~\cite{wang:neurips2020} on both bounding box (bbox) and segmentation
(segm).}
 \vspace{-2mm}
 \label{tab:2d:object}
\end{table*}

\begin{table}[t]
 \centering
 \footnotesize
 \begin{tabular}{l|C{1.40cm}C{1.40cm}C{1.40cm}C{1.40cm}}
       & S0    & S1    & S2    & S3    \\
  \hline
  hand & 36.42 & 26.85 & 32.90 & 35.18
 \end{tabular}
 \vspace{-2mm}
 \caption{\small 2D keypoint detection results in AP (\%) with Mask R-CNN
(Detectron2)~\cite{he:iccv2017,wu2019detectron2}.}
 \vspace{-2mm}
 \label{tab:2d:keypoint}
\end{table}

Tab.~\ref{tab:2d:object} shows results for object detection in average
precision (AP). First, Mask R-CNN and SOLOv2 perform similarly in mean average
precision (mAP). Mask R-CNN has a slight edge on bounding box (e.g., 75.76
versus 75.13 mAP on S0) while SOLOv2 has a slight edge on segmentation (e.g.,
71.56 versus 69.58 mAP on S0). This is because Mask R-CNN predicts bounding
boxes first and uses them to generate segmentations. Therefore the error can
accumulate for segmentation. SOLOv2 in contrast directly predicts segmentations
and uses it to generate bounding boxes. Second, the AP for hand is lower than
the AP for objects (e.g., for Mask R-CNN, 71.85 for hand versus 75.76 mAP on S0
bbox). This suggests that hands are more difficult to detect than objects,
possibly due to its larger variability in position. Finally, performance varies
across setups. For example, mAP in S1 is lower than in S0 (e.g., for Mask R-CNN,
72.69 versus 75.76 on bbox). This can be attributed to the challenge introduced
by unseen subjects. Tab.~\ref{tab:2d:keypoint} shows the AP for keypoint
detection with Mask R-CNN. We observe a similar trend on the ordering of AP
over different setups.

\subsection{6D Object Pose Estimation}
\label{sec:6d}

We evaluate single-view 6D pose estimation following the BOP challenge
protocol~\cite{hodan:eccv2018,hodan:eccvw2020}. The task asks for a 6D pose
estimate for each object instance given the number of objects and instances in
each image. The evaluation computes recall using three different pose-error
functions (VSD, MSSD, MSPD), and the final score is the average of the three
recall values (AR).

We first analyze the challenge brought by hand grasping by comparing static and
grasped objects. We use PoseCNN (RGB)~\cite{xiang:rss2018} as a baseline and
retrain the model on DexYCB. Tab.~\ref{tab:6d:grasp} shows the AR on S0
evaluated separately on the full test set, static objects only, and grasped
objects only. Without surprise the AR drops significantly when only considering
the grasped objects. This recapitulates the increasing challenge of object pose
estimation under hand interactions. To better focus on this regime, we evaluate
only on the grasped objects in the remaining experiments.

\begin{table}[t]
 \centering
 \footnotesize
 \begin{tabular}{C{1.40cm}|C{1.40cm}C{1.40cm}}
  all   & static & grasped \\
  \hline
  52.68 & 56.53  & 41.65
 \end{tabular}
 \vspace{-2mm}
 \caption{\small PoseCNN (RGB)~\cite{xiang:rss2018} results in AR (\%) on S0
(default).}
 \vspace{-2mm}
 \label{tab:6d:grasp}
\end{table}

\begin{table*}[t]
 \centering
 \footnotesize
 \setlength{\tabcolsep}{4.9pt}
 \begin{tabular}{l|cccc||cccccccc}
                           & \multirow{2}{*}{S0} & \multirow{2}{*}{S1} & \multirow{2}{*}{S2} & \multirow{2}{*}{S3} & \multicolumn{2}{c}{PoseCNN~\cite{xiang:rss2018}} & \multicolumn{2}{c}{DeepIM~\cite{li:eccv2018}} & DOPE~\cite{tremblay:corl2018} & \multicolumn{2}{c}{PoseRBPF~\cite{deng:rss2019}} & CosyPose~\cite{labbe:eccv2020} \\
                           & & & & & RGB & + depth ref & RGB & RGB-D & RGB & RGB & RGB-D & RGB \\
  \hline
  002\_master\_chef\_can   & 47.47 & 44.53 & 51.36 & --    & 44.53 & 48.24 & 60.74 & 68.40          & 19.82 & 35.19 & 58.04          & \textbf{77.21} \\
  003\_cracker\_box        & 61.04 & 57.46 & 60.65 & --    & 57.46 & 62.62 & 79.43 & 84.56          & 53.50 & 43.62 & 67.37          & \textbf{88.39} \\
  004\_sugar\_box          & 45.11 & 36.00 & 51.73 & --    & 36.00 & 42.45 & 51.29 & 59.34          & 32.06 & 30.55 & 56.55          & \textbf{69.61} \\
  005\_tomato\_soup\_can   & 36.68 & 34.88 & 45.44 & --    & 34.88 & 42.99 & 46.87 & \textbf{55.46} & 20.52 & 23.76 & 40.03          & 52.76          \\
  006\_mustard\_bottle     & 52.56 & 42.89 & 51.96 & --    & 42.89 & 48.98 & 55.52 & 63.31          & 23.66 & 28.98 & 54.93          & \textbf{67.11} \\
  007\_tuna\_fish\_can     & 32.70 & 29.58 & 31.96 & --    & 29.58 & 36.17 & 39.18 & 46.98          &~~5.86 & 20.13 & 36.77          & \textbf{49.00} \\
  008\_pudding\_box        & 44.24 & 42.60 & 50.81 & --    & 42.60 & 51.26 & 58.23 & 67.18          & 14.48 & 27.24 & 47.95          & \textbf{70.22} \\
  009\_gelatin\_box        & 46.62 & 34.39 & 44.38 & 33.07 & 34.39 & 41.71 & 47.89 & 55.54          & 21.81 & 31.27 & 46.14          & \textbf{57.63} \\
  010\_potted\_meat\_can   & 37.41 & 42.10 & 45.08 & --    & 42.10 & 51.64 & 60.19 & \textbf{69.40} & 14.15 & 31.17 & 46.43          & 65.37          \\
  011\_banana              & 38.33 & 36.71 & 44.42 & --    & 36.71 & 40.21 & 41.82 & \textbf{48.63} & 16.15 & 21.39 & 39.15          & 35.48          \\
  019\_pitcher\_base       & 53.49 & 46.03 & 53.51 & --    & 46.03 & 49.62 & 54.27 & \textbf{63.46} &~~7.02 & 19.94 & 51.38          & 52.68          \\
  021\_bleach\_cleanser    & 49.41 & 42.54 & 56.03 & 38.52 & 42.54 & 46.29 & 51.54 & 61.44          & 16.70 & 22.94 & 53.50          & \textbf{63.62} \\
  024\_bowl                & 57.42 & 54.25 & 58.15 & --    & 54.25 & 57.05 & 60.30 & 69.40          & --    & 37.00 & 62.49          & \textbf{74.74} \\
  025\_mug                 & 40.68 & 37.28 & 43.70 & --    & 37.28 & 38.69 & 41.15 & \textbf{49.51} & --    & 18.21 & 34.87          & 48.48          \\
  035\_power\_drill        & 47.93 & 41.86 & 50.79 & --    & 41.86 & 45.79 & 58.10 & \textbf{64.28} & 19.60 & 30.67 & 51.41          & 49.22          \\
  036\_wood\_block         & 40.11 & 38.66 & 49.50 & 40.53 & 38.66 & 44.08 & 49.91 & 65.39          & --    & 23.80 & 51.63          & \textbf{67.14} \\
  037\_scissors            & 21.93 & 22.87 & 25.90 & --    & 22.87 & 25.29 & 27.48 & \textbf{32.65} & 14.16 & 17.18 & 29.40          & 24.36          \\
  040\_large\_marker       & 35.09 & 32.12 & 38.19 & --    & 32.12 & 38.73 & 33.27 & 46.63          & --    & 19.05 & 29.66          & \textbf{50.58} \\
  052\_extra\_large\_clamp & 30.48 & 31.89 & 34.42 & --    & 31.89 & 33.60 & 39.63 & 47.04          & --    & 22.99 & 35.99          & \textbf{50.78} \\
  061\_foam\_brick         & 12.80 & 13.04 & 15.75 & --    & 13.04 & 16.70 & 18.14 & 27.57          & --    & 17.40 & \textbf{32.42} & 30.12          \\
  \hline
  all                      & 41.65 & 38.26 & 45.18 & 37.41 & 38.26 & 43.27 & 48.99 & \textbf{57.54} & --    & 26.24 & 46.48          & 57.43
 \end{tabular}
 \vspace{-2mm}
 \caption{\small 6D object pose estimation results in AR (\%). Left: PoseCNN (RGB). Right: performance of representative approaches on S1.}
 \vspace{-2mm}
 \label{tab:6d:sota}
\end{table*}

We next evaluate on the four setups using the same baseline. Results are shown
in Tab.~\ref{tab:6d:sota} (left). On S1 (unseen subjects), we observe a drop in
AR compared to S0 (e.g., 38.26 versus 41.65 on all) as in object detection.
This can be attributed to the influence of interaction from unseen hands as
well as different grasping styles. The column of S3 (unseen grasping) shows the
AR of the three only objects being grasped in the test set (i.e.
``009\_gelatin\_box'', ``021\_bleach\_cleanser'', ``036\_wood\_block'').
Again, the AR is lower compared to on S0, and the drop is especially
significant for smaller objects like ``009\_gelatin\_box'' (33.07 versus
46.62). Surprisingly, AR on S2 (unseen views) does not drop but rather
increases slightly (e.g., 45.18 versus 41.65 on all). This suggests that our
multi-camera setup has a dense enough coverage of the scene such that training
on certain views can translate well to some others.

Finally, we benchmark five representative approaches:
PoseCNN~\cite{xiang:rss2018}, DeepIM~\cite{li:eccv2018},
DOPE~\cite{tremblay:corl2018}, PoseRBPF~\cite{deng:rss2019}, and
CosyPose~\cite{labbe:eccv2020} (winner of the BOP challenge 2020). All of them
take RGB input. For PoseCNN, we also include a variant with post-process depth
refinement. For DeepIM and PoseRBPF, we also include their RGB-D variants. For
CosyPose, we use its single view version. For DeepIM and CosyPose, we
initialize the pose from the output of PoseCNN (RGB). We retrain all the models
on DexYCB except for DOPE and PoseRBPF, which trained their models solely on
synthetic images. For DOPE, we train models for an additional 7 objects besides
the 5 provided. Tab.~\ref{tab:6d:sota} (right) compares the results on the most
challenging S1 (unseen subjects) setup. First, we can see a clear edge on those
which also use depth compared to their respective RGB only variants (for
PoseCNN, from 38.26 to 43.27 AR on all). Second, refinement based methods like
DeepIM and CosyPose are able to improve significantly upon their initial pose
input (e.g., for DeepIM RGB-D, from 38.26 to 57.54 AR on all). Finally,
CosyPose achieves the best overall AR among all the RGB-based methods (57.43 AR
on all).

\subsection{3D Hand Pose Estimation}
\label{sec:3d}

The task is to estimate the 3D position of 21 hand joints from a single image.
We evaluate with two metrics: mean per joint position error (MPJPE) in mm and
percentage of correct keypoints (PCK). Besides computing error in absolute 3D
position, we also report errors after aligning the predictions with ground
truths in
post-processing~\cite{zimmermann:iccv2017,zimmermann:iccv2019,hampali:cvpr2020}.
We consider two alignment methods: root-relative and Procrustes. The former
removes ambiguity in translation by replacing the root (wrist) location with
ground thuths. The latter removes ambiguity in translation, rotation, and
scale, thus focusing specifically on articulation. For PCK we report the area
under the curve (AUC) over the error range $[0,50~\mathrm{mm}]$ with 100 steps.

We benchmark one RGB and one depth based approach. For RGB, we select a
supervised version of Spurr et al.~\cite{spurr:eccv2020} which won the HANDS
2019 Challenge~\cite{armagan:eccv2020}. We experiment with the original
ResNet50 backbone as well as a stronger HRNet32~\cite{sun:cvpr2019} backbone,
both initialized with ImageNet pre-trained models and finetuned on DexYCB. For
depth, we select A2J~\cite{xiong:iccv2019} and retrain the model on DexYCB.

\begin{table}[t]
 \centering
 \footnotesize
 \setlength{\tabcolsep}{4.3pt}
 \begin{tabular}{ll|C{1.52cm}C{1.52cm}C{1.52cm}}
                      & & absolute & root-relative & Procrustes \\
  \hline
  \multirow{3}{*}{S0} & ~\cite{spurr:eccv2020} + ResNet50 & 53.92 (0.307)                   & 17.71 (0.683)                   &~~7.12 (0.858)                   \\
                      & ~\cite{spurr:eccv2020} + HRNet32  & 52.26 (0.328)                   & \textbf{17.34} (\textbf{0.698}) &~~\textbf{6.83} (\textbf{0.864}) \\
                      & A2J~\cite{xiong:iccv2019}         & \textbf{27.53} (\textbf{0.612}) & 23.93 (0.588)                   & 12.07 (0.760)                   \\
  \hline
  \multirow{3}{*}{S1} & ~\cite{spurr:eccv2020} + ResNet50 & 70.23 (0.240)                   & 22.71 (0.601)                   &~~8.43 (0.832)                   \\
                      & ~\cite{spurr:eccv2020} + HRNet32  & 70.10 (0.248)                   & \textbf{22.26} (\textbf{0.615}) &~~\textbf{7.98} (\textbf{0.841}) \\
                      & A2J~\cite{xiong:iccv2019}         & \textbf{29.09} (\textbf{0.584}) & 25.57 (0.562)                   & 12.95 (0.743)                   \\
  \hline
  \multirow{3}{*}{S2} & ~\cite{spurr:eccv2020} + ResNet50 & 83.46 (0.208)                   & \textbf{23.15} (\textbf{0.566}) &~~\textbf{8.11} (\textbf{0.838}) \\
                      & ~\cite{spurr:eccv2020} + HRNet32  & 80.63 (0.217)                   & 25.49 (0.530)                   &~~8.21 (0.836)                   \\
                      & A2J~\cite{xiong:iccv2019}         & \textbf{23.44} (\textbf{0.576}) & 27.65 (0.540)                   & 13.42 (0.733)                   \\
  \hline
  \multirow{3}{*}{S3} & ~\cite{spurr:eccv2020} + ResNet50 & 58.69 (0.281)                   & 19.41 (0.665)                   &~~7.56 (0.849)                   \\
                      & ~\cite{spurr:eccv2020} + HRNet32  & 55.39 (0.311)                   & \textbf{18.44} (\textbf{0.686}) &~~\textbf{7.06} (\textbf{0.859}) \\
                      & A2J~\cite{xiong:iccv2019}         & \textbf{30.99} (\textbf{0.608}) & 24.92 (0.581)                   & 12.15 (0.759)
 \end{tabular}
 \vspace{-2mm}
 \caption{\small 3D hand pose estimation results in MPJPE (mm). Numbers in
parentheses are AUC values.}
 \vspace{-3mm}
 \label{tab:3d}
\end{table}

Tab.~\ref{tab:3d} shows the results. For~\cite{spurr:eccv2020}, we can see that
estimating absolute 3D position solely from RGB is difficult (e.g., 53.92 mm
absolute MPJPE with ResNet50 on S0). The stronger HRNet32 backbone reduces the
errors over ResNet50 but only marginally (e.g., 52.26 versus 53.92 mm absolute
MPJPE on S0). Similar to in object pose, the errors on S1 (unseen subjects)
increase from S0 (e.g., 70.10 versus 52.26 mm absolute MPJPE for HRNet32) due
to the impact of unseen hands and grasping styles. However, unlike the trend in
Tab.~\ref{tab:6d:sota} (left), the errors on S2 (unseen views) also increase
pronouncedly and even surpass the other setups (e.g., 80.63 mm absolute MPJPE
for HRNet32). This is because unlike objects, the subjects are always facing
the same direction from a situated position in this setup. This further
constrains the possible hand poses observed in each view, making cross view
generalization more challenging. Unsurprisingly, A2J~\cite{xiong:iccv2019}
outperforms~\cite{spurr:eccv2020} on absolute MPJPE due to the depth input, but
falls short on estimating articulation as shown by the errors after alignment
(e.g., 12.07 versus 6.83 mm Procrustes MPJPE for HRNet32 on S0).


\section{Safe Human-to-Robot Object Handover}


\paragraph{Task} Given an RGB-D image with a person holding an object, the goal
is to generate a diverse set of robot grasps to take over the object without
pinching the person's hand (we refer to as ``safe'' handovers). The diversity
of grasps is important since not all the grasps are kinematically feasible for
execution. We assume a parallel-jaw Franka Panda gripper and represent each
grasp as a point in $\textit{SE}(3)$.

\begin{figure}[t]
 \centering
 \includegraphics[width=0.91\linewidth]{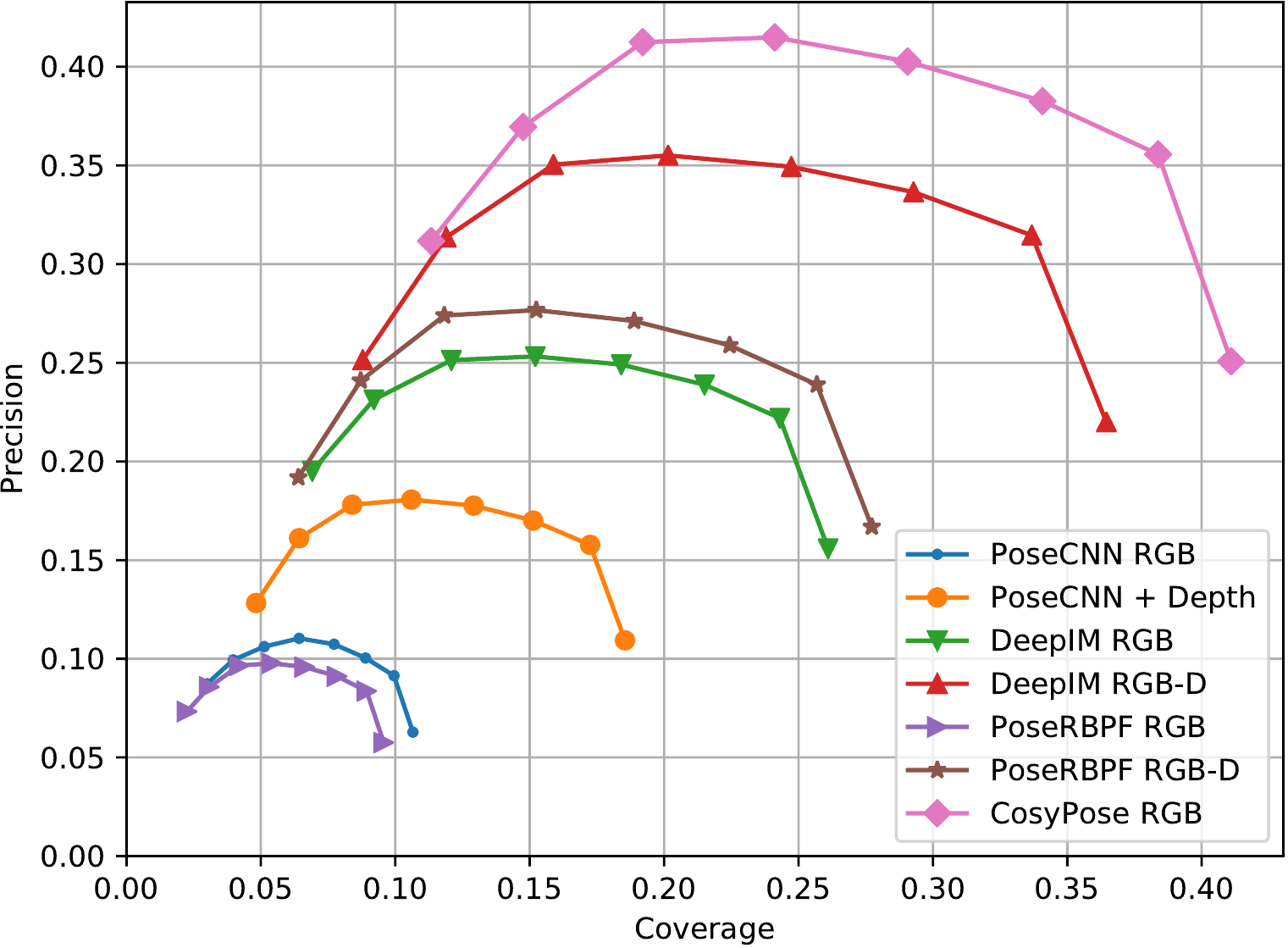}
 \vspace{-2mm}
 \caption{\small Precision-coverage curves for grasp generation on S1.}
 \vspace{-2mm}
 \label{fig:curve}
\end{figure}

\vspace{-3mm}
\paragraph{Evaluation} We first sample 100 grasps for each YCB object using
farthest point sampling from a diverse set of grasps pre-generated for that
object in~\cite{eppner:isrr2019}. This ensures a dense coverage of the pose
space (Fig.~\ref{fig:grasp}). For each image, we transform these grasps from
the object frame to camera frame using the ground-truth object pose, and remove
those collided with the ground-truth object and hand mesh. This generates a
reference set of successful grasps $\mathcal{R}$.

Given a set of predicted grasps $\chi$, we evaluate its diversity by computing
the \textit{coverage}~\cite{eppner:isrr2019} of $\mathcal{R}$, defined by the
percentage of grasps in $\mathcal{R}$ having at least one matched grasp in
$\chi$ that is neither collided with the object nor the hand. Specifically, two
grasps $g$, $h$ are considered matched if $|g_{t}-h_{t}|<\sigma_{t}$ and
$\arccos(|\langle g_{q},h_{q}\rangle|)<\sigma_{q}$, where $g_{t}$ is the
translation and $g_{q}$ is the orientation in quaternion. We use
$\sigma_{t}=0.05~\mathrm{m}$ and $\sigma_{q}=\ang{15}$.

One could potentially hit a high coverage by sampling grasps exhaustively.
Therefore we also compute \textit{precision}, defined as the percentage of
grasps in $\chi$ that have at least one matched successful grasp in
$\mathcal{R}$. 

\vspace{-3mm}
\paragraph{Baseline} We experiment with a simple baseline that only requires
hand segmentation and 6D object pose. Similar to constructing $\mathcal{R}$, we
transform the 100 grasps to the camera frame but using the estimated object
pose, then remove those that are collided with the hand point cloud obtained by
the hand segmentation and the depth image. Specifically, a grasp is collided if
the distance of a pair of points from the gripper point cloud and the hand
point cloud is less than a threshold $\epsilon$. The gripper point cloud is
obtained from a set of pre-sampled points on the gripper surface. We use the
hand segmentation results from Mask R-CNN (Sec.~\ref{sec:2d}).


\vspace{-3mm}
\paragraph{Results} We evaluate grasps generated with different object pose
methods at different threshold $\epsilon\in[0,0.07~\mathrm{m}]$ and show the
precision-coverage curves on S1 in Fig.~\ref{fig:curve}. We see that better
object pose estimation leads to better grasp generation. Fig.~\ref{fig:grasp}
shows qualitative examples of the predicted grasps. We see that most of the
failure grasps (red and gray) are due to inaccurate object pose. Some are
hand-colliding grasps caused by a miss detected hand when the hand is partially
occluded by the object (e.g., ``003\_cracker\_box''). This can be potentially
addressed by model based approaches that directly predict the full hand shape.

\begin{figure}[t]
 \centering
 \includegraphics[width=0.32\linewidth]{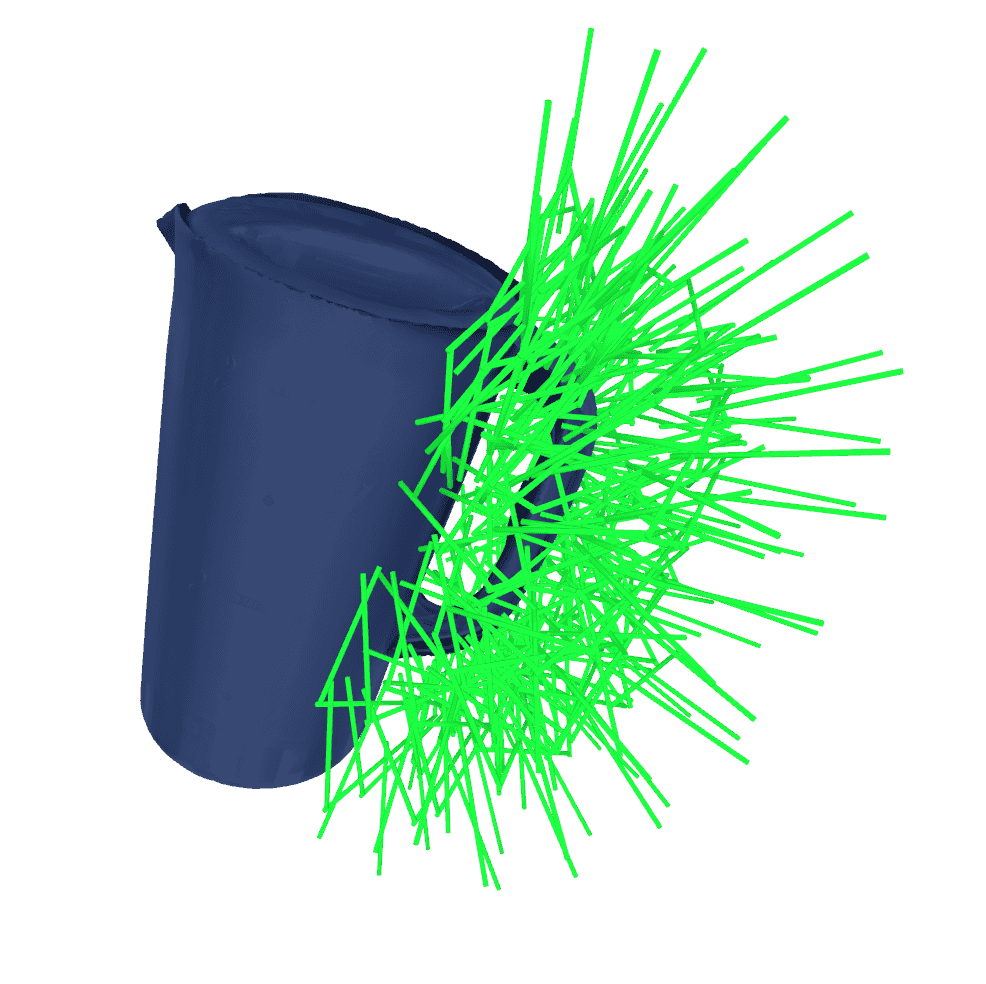}
 \includegraphics[width=0.32\linewidth]{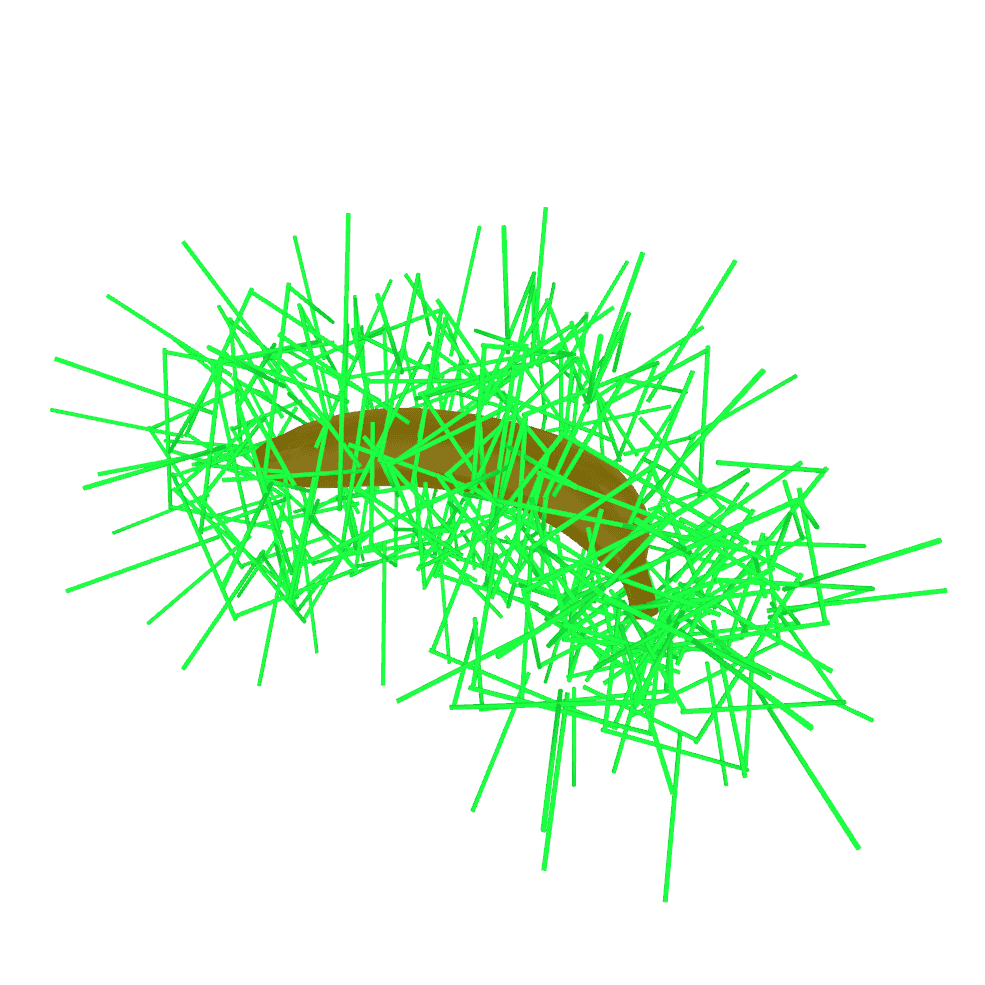}
 \includegraphics[width=0.32\linewidth]{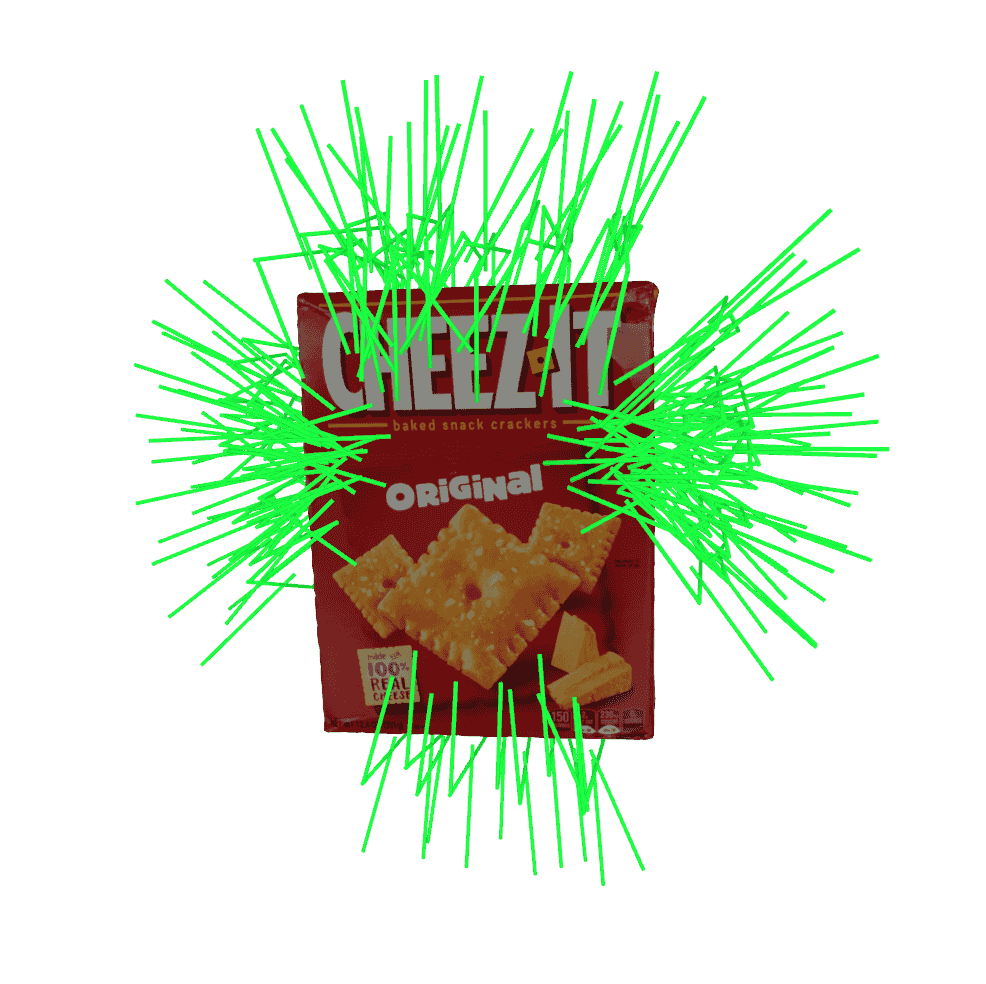}
 \includegraphics[width=0.32\linewidth]{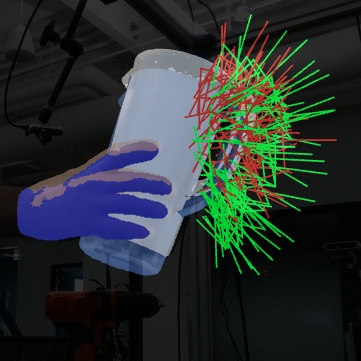}
 \includegraphics[width=0.32\linewidth]{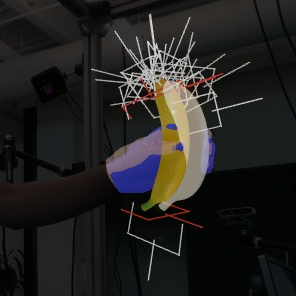}
 \includegraphics[width=0.32\linewidth]{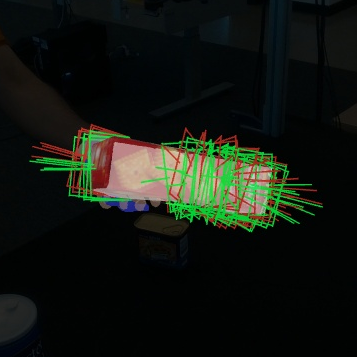}
 \caption{\small Top: 100 successful grasps sampled
from~\cite{eppner:isrr2019}. Bottom: predicted grasps generated by the
predicted object pose (textured model) and hand segmentation (blue masks).
Green ones denote those covering successful grasps, red ones denote those
collided with the object or hand, and gray ones are failures not covering any
successful grasps in the reference set. Ground-truth objects and hands are
shown in translucent white and brown meshes.
 }
 \vspace{-2mm}
 \label{fig:grasp}
\end{figure}


\section{Conclusions}

We have introduced DexYCB for capturing hand grasping of objects. We have shown
its merits, presented a thorough benchmark of current approaches on three
relevant tasks, and evaluated a new robotics-relevant task. We envision our
dataset will drive progress on these crucial fronts.

{\small
\bibliographystyle{ieee_fullname}
\bibliography{egbib}
}


\appendix

\appendixpage

\etocdepthtag.toc{mtappendix}
\etocsettagdepth{mtchapter}{none}
\etocsettagdepth{mtappendix}{subsection}

{
  \hypersetup{
    linkcolor = black
  }
  \tableofcontents
}

\section{Constructing DexYCB}

\subsection{Optimizing $E_{\text{depth}}$}

Here we provide the details on optimizing the depth term $E_{\text{depth}}$ in
solving 3D hand and object pose (Sec.~\ref{sec:solve}).
Below we first describe our efficient forward computation based on a
point-parallel GPU implementation. Then we show that $E_{\text{depth}}$ is
differentiable and derive its gradient.

Recall that $\{d_{i}\in\mathbb{R}^{3}\}_{i=1}^{N_{D}}$ denotes the entire point
cloud where $d_i$ is a single 3D point, and $\mathcal{M}(P)$ is the triangular
mesh of our model. Without loss of generality, we assume that our model
contains only one single rigid object, i.e. $P\in\text{SE(3)}$. Recall that the
depth term is defined as
\begin{equation}
 \label{eq:depth}
 E_{\text{depth}}(P)=\frac{1}{N_{D}}\sum_{i=1}^{N_{D}}|\text{SDF}(d_{i},\mathcal{M}(P))|^{2},
\end{equation}
where $\text{SDF}(\cdot)$ calculates the signed distance value of a 3D point
$d$ from the mesh $\mathcal{M}$. One naive way to compute this value is to
exhaustively compute the distance from the query point $d$ to all the
triangular faces on the mesh, and find the minimum value among them. To perform
a more efficient computation, we leverage the bounding volume hierarchy (BVH)
technique from graphics. Specifically, we build an axis-aligned bounding box
tree (AABB tree) to index the triangular faces of the mesh at the start of each
forward computation of Eq.~\ref{eq:depth}. The AABB tree allows us to
efficiently traverse the triangular faces of the mesh and compute distances
from the query $d$ without going through all the faces exhaustively.
Furthermore, since Eq.~\ref{eq:depth} repeats the same computation for all the
query points $\{d_i\}$, we can achieve further speedup by parallelizing over
query points using a GPU implementation.

\begin{table*}[t]
 \centering
 \small
 \begin{tabular}{ccccc}
  thumb tip & index tip & middle tip & ring tip & pinky tip \\
  \hline
  4.27 $\pm$ 3.42 & 4.99 $\pm$ 3.67 & 4.33 $\pm$ 3.53 & 4.03 $\pm$ 3.56 & 3.90 $\pm$ 3.45 \\
 \end{tabular}
 \vspace{-2mm}
 \caption{\small Reprojection error (pixels) of the five finger tips.}
 \label{tab:reprojection}
\end{table*}

Now let us turn to the backward pass of Eq.~\ref{eq:depth}, since we optimize
$E_{\text{depth}}(P)$ using a gradient-based method. During the forward
computation, for a query point $d$, once we find its closest point
$q\in\mathbb{R}^{3}$ on the mesh, we represent $q$ using barycentric
coordinates:
\begin{equation}
 \label{eq:barycentric}
 q=u\cdot a+v\cdot b+w\cdot c,
\end{equation}
where $a,b,c\in\mathbb{R}^{3}$ denotes the three vertices of the triangular
face which $q$ lies on, and $u,v,w\in\mathbb{R}$ with $0\leq u,v,w\leq1$ and
$u+v+w=1$. Now we can further represent $E_{\text{depth}}(P)$ with
\begin{equation}
 \begin{split}
  E_{\text{depth}}(P)&=\frac{1}{N_{D}}\sum_{i=1}^{N_{D}}|\text{SDF}(d_{i},\mathcal{M}(P))|^{2}\\
                     &=\frac{1}{N_{D}}\sum_{i=1}^{N_{D}}||d_i-q_i||^{2}_{2}.
 \end{split}
\end{equation}
We can thus derive the gradient as
\begin{equation}
 \frac{\partial E_{\text{depth}}(P)}{\partial P}=\frac{1}{N_{D}}\sum_{i=1}^{N_{D}}\frac{\partial||d_i - q_i||^{2}_{2}}{\partial P}.
\end{equation}
Let
$\mathcal{V}_{n}=(\mathcal{V}_{n,x},\mathcal{V}_{n,y},\mathcal{V}_{n,z})\in\mathbb{R}^{3}$
denote the $n$th vertex of the mesh $\mathcal{M}$ with $N_{V}$ vertices. Using
the multivariable chain rule, we get
\begin{equation}
 \label{eq:chain}
 \frac{\partial||d- q||^{2}_{2}}{\partial P}=\sum_{n=1}^{N_{V}}\sum_{m\in\{x, y, z\}}\frac{\partial||d-q||^2_{2}}{\partial\mathcal{V}_{n,m}}\cdot\frac{\partial\mathcal{V}_{n,m}}{\partial P}.
\end{equation}
With the barycentric representation of $q$ from Eq.\ref{eq:barycentric}, we can
write
\begin{equation}
 \label{eq:error}
 \begin{split}
  ||d-q||^{2}_{2}&=||d-(u\cdot a+v\cdot b+w\cdot c)||^{2}_{2}\\
                 &=(d_{x}-u\cdot a_{x}-v\cdot b_{x}-w\cdot c_{x})^2\\
                 &\qquad {} +(d_{y}-u\cdot a_{y}-v\cdot b_{y}-w\cdot c_{y})^2\\
                 &\qquad {} +(d_{z}-u\cdot a_{z}-v\cdot b_{z}-w\cdot c_{z})^2.
 \end{split}
\end{equation}
Finally, with Eq.~\ref{eq:chain} and~\ref{eq:error}, we can derive the gradient
with respect to the vertices as
\begin{equation}
 \begin{split}
 &\frac{\partial||d-q||^2_{2}}{\partial\mathcal{V}_{n,m}}=\\
 &
 \begin{cases}
  -2\cdot u\cdot(d_{x}-u\cdot a_{x}-v\cdot b_{x}-w\cdot c_{x})&\text{if }\mathcal{V}_{n,m}=a_{x},\\
  -2\cdot u\cdot(d_{y}-u\cdot a_{y}-v\cdot b_{y}-w\cdot c_{y})&\text{if }\mathcal{V}_{n,m}=a_{y},\\
  -2\cdot u\cdot(d_{z}-u\cdot a_{z}-v\cdot b_{z}-w\cdot c_{z})&\text{if }\mathcal{V}_{n,m}=a_{z},\\
  -2\cdot v\cdot(d_{x}-u\cdot a_{x}-v\cdot b_{x}-w\cdot c_{x})&\text{if }\mathcal{V}_{n,m}=b_{x},\\
  -2\cdot v\cdot(d_{y}-u\cdot a_{y}-v\cdot b_{y}-w\cdot c_{y})&\text{if }\mathcal{V}_{n,m}=b_{y},\\
  -2\cdot v\cdot(d_{z}-u\cdot a_{z}-v\cdot b_{z}-w\cdot c_{z})&\text{if }\mathcal{V}_{n,m}=b_{z},\\
  -2\cdot w\cdot(d_{x}-u\cdot a_{x}-v\cdot b_{x}-w\cdot c_{x})&\text{if }\mathcal{V}_{n,m}=c_{x},\\
  -2\cdot w\cdot(d_{y}-u\cdot a_{y}-v\cdot b_{y}-w\cdot c_{y})&\text{if }\mathcal{V}_{n,m}=c_{y},\\
  -2\cdot w\cdot(d_{z}-u\cdot a_{z}-v\cdot b_{z}-w\cdot c_{z})&\text{if }\mathcal{V}_{n,m}=c_{z},\\
  0&\text{otherwise}.
 \end{cases}
 \end{split} 
\end{equation}
Since we can also compute $\partial\mathcal{V}_{n,m}/\partial P$ for rigid
objects in Eq.~\ref{eq:chain}, we can obtain the gradient of
$E_{\text{depth}}(P)$. In fact, we can compute the gradient of
$E_{\text{depth}}(P)$ as long as $\mathcal{V}_{n,m}(P)$ is differentiable. This
holds true for the YCB objects (rigid) and also the MANO hand model
(deformable). Therefore we can optimize Eq.~\ref{eq:depth} with both our object
and hand models. Similar to the forward computation, the backward pass can also
be parallelized using a GPU implementation.

\section{Properties of DexYCB}

\subsection{Visualization}

For a better visualization of data and annotations please see the supplementary
video.~\footnote{\href{https://youtu.be/Q4wyBaZeBw0}{\texttt{https://youtu.be/Q4wyBaZeBw0}}.}

\subsection{Analysis on 3D Annotation Accuracy}

\begin{figure}
 \centering
 \includegraphics[width=0.40\linewidth]{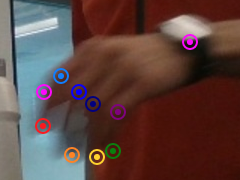}~
 \includegraphics[width=0.40\linewidth]{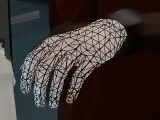}
 \\ \vspace{1mm}
 \includegraphics[width=0.40\linewidth]{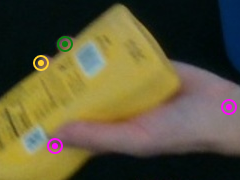}~
 \includegraphics[width=0.40\linewidth]{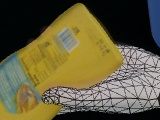}
 \caption{\small Qualitative examples of annotations with high reprojection
errors.}
 \label{fig:reprojection}
\end{figure}

To analyze the accuracy of 3D annotation in DexYCB, we compute the reprojection
error of ground-truth 3D keypoints with human labeled 2D keypoints in each
image. Tab.~\ref{tab:reprojection} reports the reprojection errors of the five
finger tips. The mean errors are all below 5 pixels. As shown in
Fig.~\ref{fig:reprojection}, the large errors are often produced by a fast
moving hand (top) and occasionally ambiguous annotations (bottom: ring labeled
as pinky).

\subsection{Diversity of Hand Pose}
\label{sec:diversity}

\begin{figure}[t]
 \centering
 \includegraphics[width=0.80\linewidth]{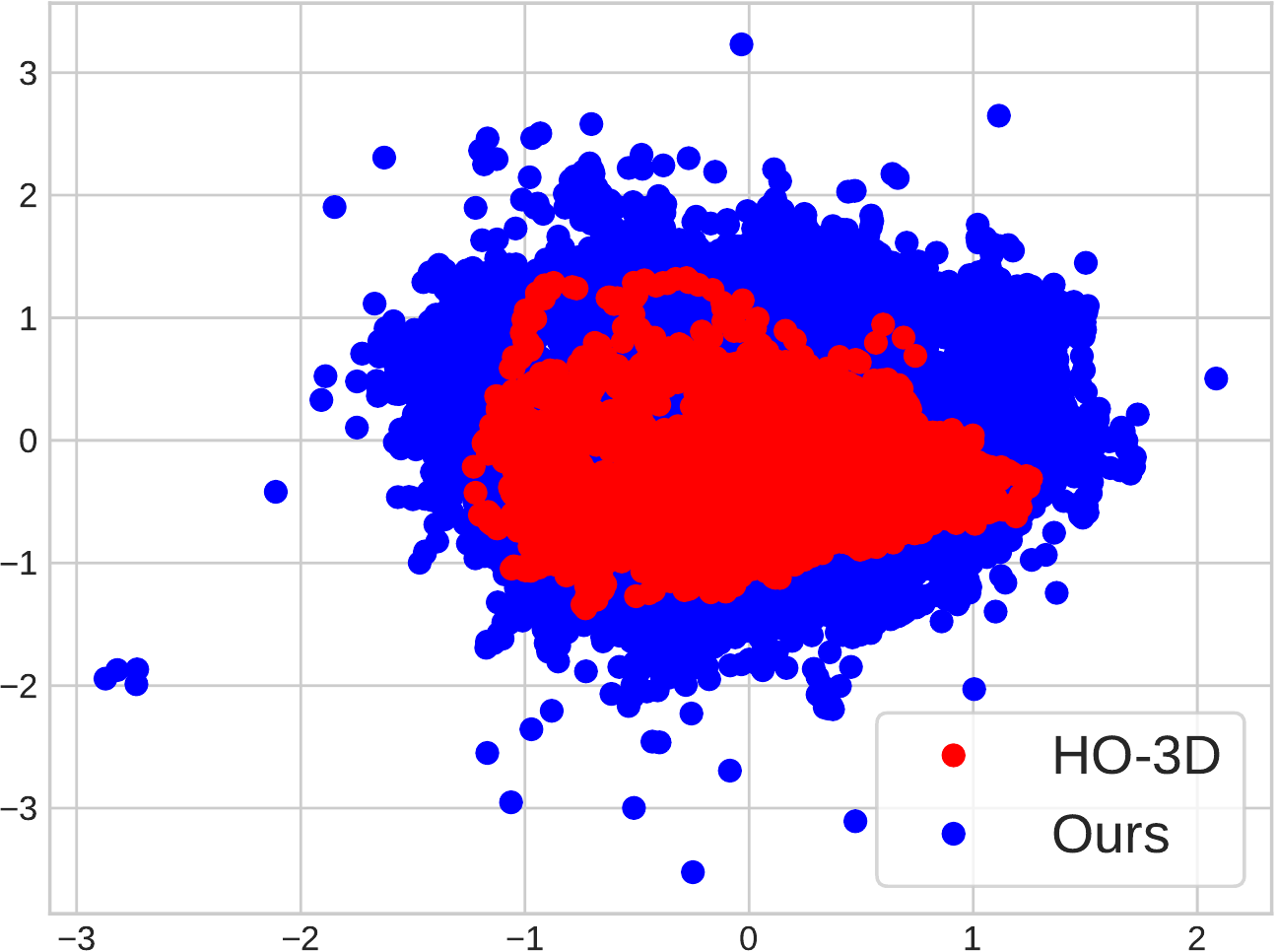}
 \caption{\small Distribution of hand pose represented by the first two MANO
PCA coefficients.}
 \label{fig:pca}
\end{figure}

DexYCB pushes prior hand datasets to an unprecedented scale on the diversity of
grasps. To analyze the diversity of hand pose, we extract the PCA
representation of the MANO hand model for each hand pose and visualize the
distribution in 2D space using the first two PCA coefficients.
Fig.~\ref{fig:pca} compares the distribution of hand pose in DexYCB with
HO-3D~\cite{hampali:cvpr2020}. While DexYCB grows the number of objects as well
as the number of grasps per object over prior datasets, the main edge actually
lies in the diversity of captured grasps as shown by the distribution of hand
pose. This diversity marks an essential challenge of pose estimation in
hand-object interactions and downstream applications like object handover to
robots.

\begin{table*}[t]
 \centering
 \small
 \begin{tabular}{l|ccccc|cccc|ccc}
        & \multirow{3}{*}{\#sub} & \multirow{3}{*}{\#obj} & \multirow{3}{*}{\#view} & \multirow{3}{*}{\#seq} & \multirow{3}{*}{\#image} & \multicolumn{4}{c|}{\#obj anno}         & \multicolumn{3}{c}{\#hand anno} \\
        &                        &                        &                         &                        &                          & \multirow{2}{*}{all}                    & \multirow{2}{*}{grasped} & \multirow{2}{*}{6D eval} & handover & \multirow{2}{*}{all} & \multirow{2}{*}{right} & \multirow{2}{*}{left} \\
        &                        &                        &                         &                        &                          &                                         &                          &                          & eval     &                      &                        &                       \\
  \hline
  all   & 10 &  20 & 8 &   1,000 &  581,968 &   2,317,312 &  581,968 & --       & --    &  508,384 &  254,000 &  254,384 \\
  \hline
  \hline
  \multicolumn{13}{c}{S0 (default)} \\
  \hline
  train & 10 &  20 & 8 &  ~~~800 &  465,504 &   1,846,144 &  465,504 & --       & --    &  406,888 &  203,096 &  203,792 \\
  val   &  2 &  20 & 8 & ~~~~~40 & ~~23,200 & ~~~~~97,456 & ~~23,200 & --       & --    & ~~22,728 & ~~11,296 & ~~11,432 \\
  test  &  8 &  20 & 8 &  ~~~160 & ~~93,264 &  ~~~373,712 & ~~93,264 & ~~94,816 & 1,280 & ~~78,768 & ~~39,608 & ~~39,160 \\
  \hline
  \hline
  \multicolumn{13}{c}{S1 (unseen subjects)} \\
  \hline
  train &  7 &  20 & 8 &  ~~~700 &  407,088 &   1,620,128 &  407,088 & --       & --    &  356,600 &  177,656 &  178,944 \\
  val   &  1 &  20 & 8 &  ~~~100 & ~~58,592 &  ~~~226,288 & ~~58,592 & --       & --    & ~~47,656 & ~~23,848 & ~~23,808 \\
  test  &  2 &  20 & 8 &  ~~~200 &  116,288 &  ~~~470,896 &  116,288 &  119,192 & 1,600 &  104,128 & ~~52,496 & ~~51,632 \\
  \hline
  \hline
  \multicolumn{13}{c}{S2 (unseen views)} \\
  \hline
  train & 10 &  20 & 6 &  ~~~750 &  436,476 &   1,737,984 &  436,476 & --       & --    &  381,288 &  190,500 &  190,788 \\
  val   & 10 &  20 & 1 &  ~~~125 & ~~72,746 &  ~~~289,664 & ~~72,746 & --       & --    & ~~63,548 & ~~31,750 & ~~31,798 \\
  test  & 10 &  20 & 1 &  ~~~125 & ~~72,746 &  ~~~289,664 & ~~72,746 & ~~73,383 & 1,000 & ~~63,548 & ~~31,750 & ~~31,798 \\
  \hline
  \hline
  \multicolumn{13}{c}{S3 (unseen grasping)} \\
  \hline
  train & 10 &  15 & 8 &  ~~~750 &  436,416 &   1,742,672 &  436,416 & --       & --    &  381,288 &  189,712 &  191,576 \\
  val   & 10 & ~~2 & 8 &  ~~~100 & ~~58,192 &  ~~~221,664 & ~~58,192 & --       & --    & ~~50,736 & ~~25,864 & ~~24,872 \\
  test  & 10 & ~~3 & 8 &  ~~~150 & ~~87,360 &  ~~~352,976 & ~~87,360 & ~~89,392 & 1,200 & ~~76,360 & ~~38,424 & ~~37,936
 \end{tabular}
 \vspace{-2mm}
 \caption{\small Statistics of the four evaluation setups: S0 (default), S1 (unseen subjects), S2 (unseen views), and S3 (unseen grasping).}
 \label{tab:setup}
\end{table*}

\subsection{Why Black Background?}
\label{sec:background}

The main focus of DexYCB is on the diversity of grasps, not the diversity of
backgrounds. Capturing hand-object interaction is challenging already in
controlled settings. Besides, DexYCB is no lesser than prior datasets in this
respect: FreiHAND~\cite{zimmermann:iccv2019} used a green background; while
HO-3D~\cite{hampali:cvpr2020} used more natural backgrounds, the diversity is
not much better since the choice of background scenes in HO-3D is still only
two (Fig.~\ref{fig:ho3d}).

\begin{figure}
 \centering
 \includegraphics[width=0.32\linewidth]{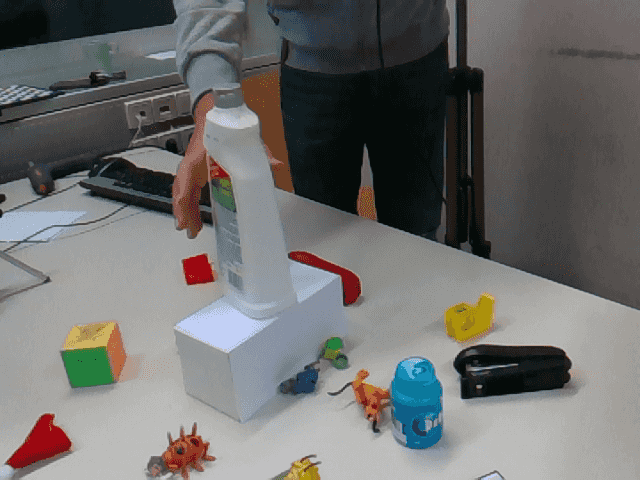}
 \includegraphics[width=0.32\linewidth]{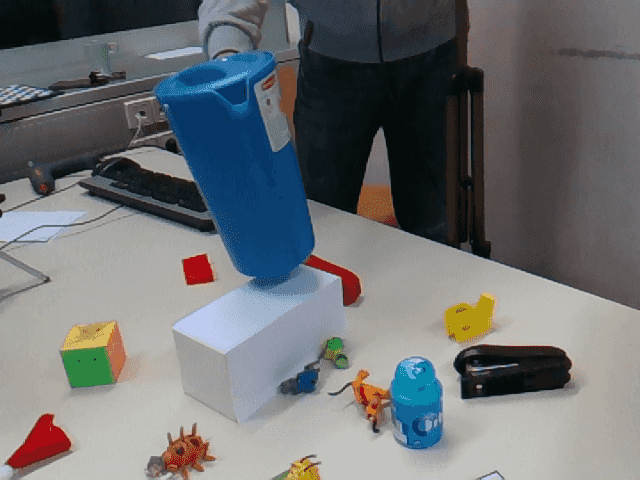}
 \includegraphics[width=0.32\linewidth]{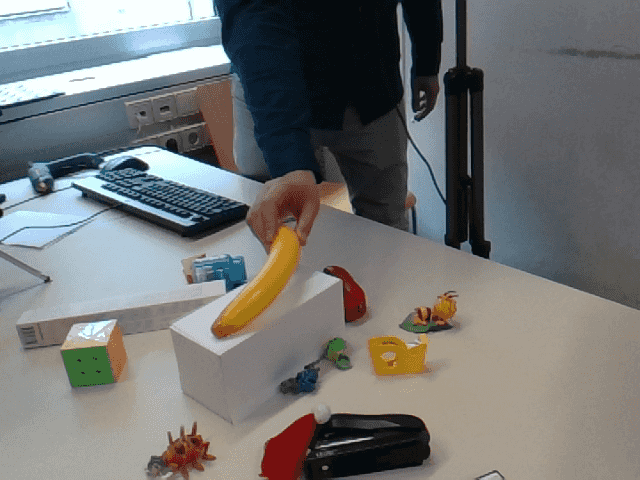}
 \\ \vspace{1mm}
 \includegraphics[width=0.32\linewidth]{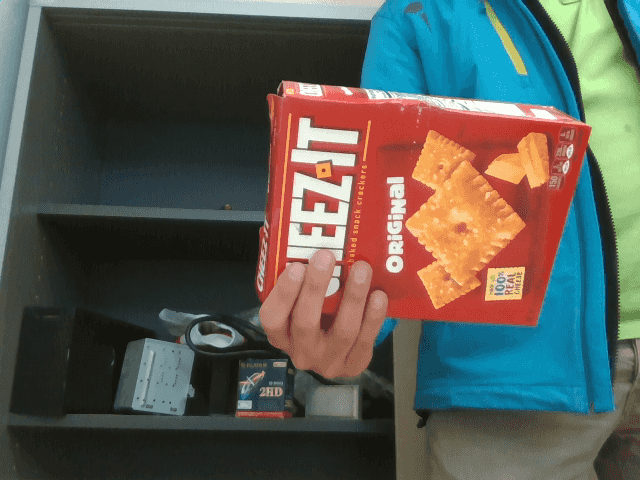}
 \includegraphics[width=0.32\linewidth]{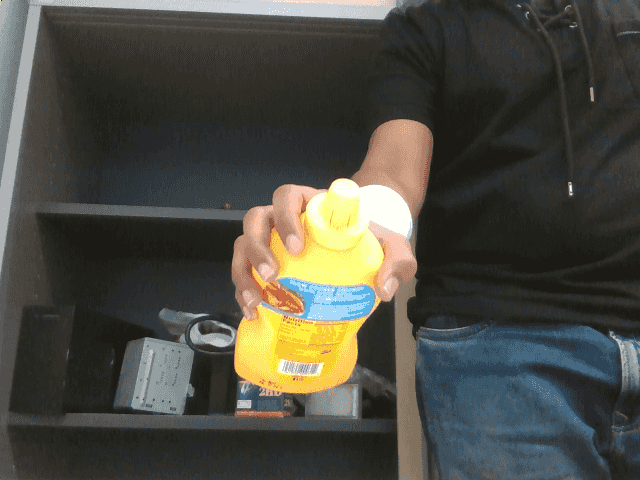}
 \includegraphics[width=0.32\linewidth]{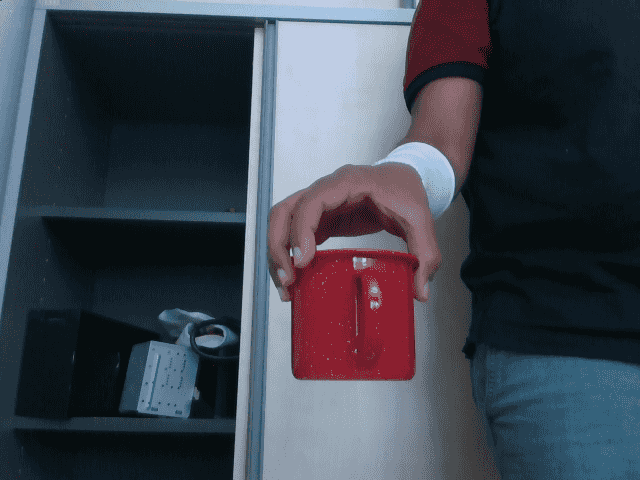}
 \caption{\small The two captured scene backgrounds (top and bottom) in
HO-3D~\cite{hampali:cvpr2020}.}
 \label{fig:ho3d}
\end{figure}

\subsection{Why Generalize Better over HO-3D?}

Following~\ref{sec:diversity} and~\ref{sec:background}, DexYCB generalizes
better due to a better diversity of grasps given a similar diversity of scene
background. This explains the edge of DexYCB in the cross-dataset evaluation in
Sec.~\ref{sec:cross}.

\section{Benchmarking Representative Approaches}

\subsection{Setup Details}

Tab.~\ref{tab:setup} shows the statistics of the four evaluation setups: S0
(default), S1 (unseen subjects), S2 (unseen views), and S3 (unseen grasping).
The first row (all) represents the full dataset, and each sub-table below shows
the statistics of one particular setup, divided into train/val/test splits. For
each split, we list the number of subjects (``\#sub''), objects (``\#obj''),
views (``\#view''), sequences (``\#seq''), and image samples (``\#image'').

We also list the number of object annotations (``\#obj anno'') in each split,
including the full object set (``all'') and the subset with only the grasped
objects (``grasped''). For 6D object pose estimation, we follow the BOP
challenge to speed up the evaluation by evaluating on subsampled keyframes. We
use a subsampling factor of 4. The column ``6D eval'' lists the number of
object annotations in the keyframes of the test split. For object handover, the
test images need to capture a person with an object in hand ready for handover.
Therefore we use the last frame of each video as the test samples, since during
capture the subjects are instructed to hand over the object to someone across
at the end. The column ``handover eval'' lists the number of in-hand object
annotations (which is also the number of the last frames from all videos since
each image contains one in-hand object) in the test split for the handover
evaluation. 

Finally, we list the number of hand annotations (``\#hand anno'') in each
split, including the full hand set (``all'') and the subsets with only right
(``right'') and left (``left'') hands.

\subsection{Qualitative: 2D Object and Keypoint Detection}

\begin{figure*}[t]
 \centering
 \begin{minipage}{1\textwidth}
  \includegraphics[width=0.24\linewidth]{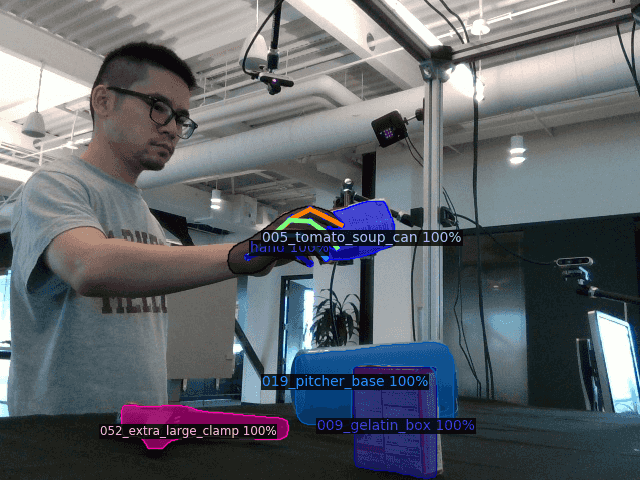}~
  \includegraphics[width=0.24\linewidth]{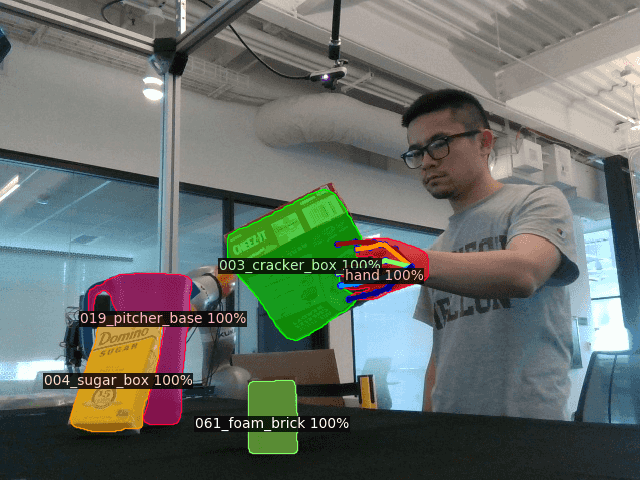}~
  \includegraphics[width=0.24\linewidth]{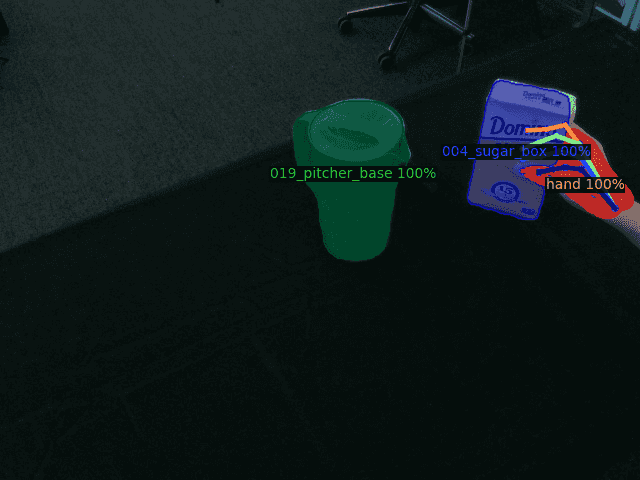}~
  \includegraphics[width=0.24\linewidth]{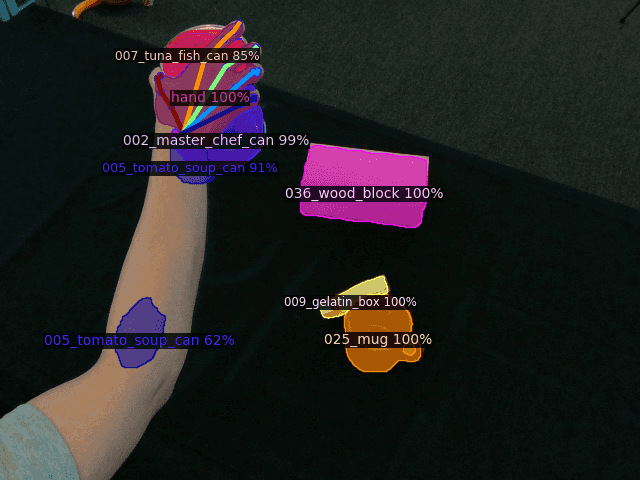}
 \end{minipage}
 \\ \vspace{1mm}
 \begin{minipage}{1\textwidth}
  \includegraphics[width=0.24\linewidth]{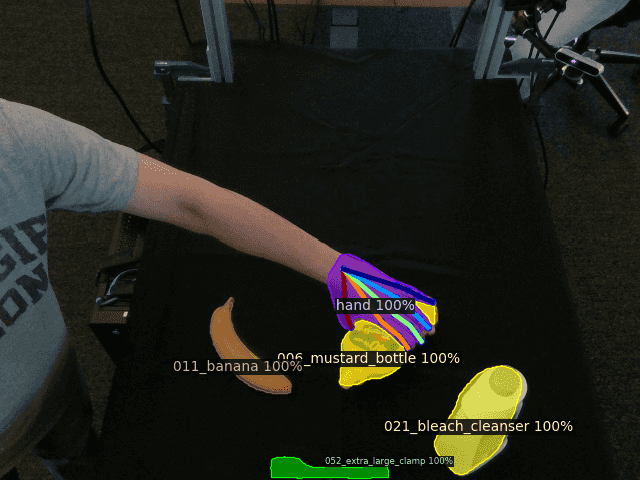}~
  \includegraphics[width=0.24\linewidth]{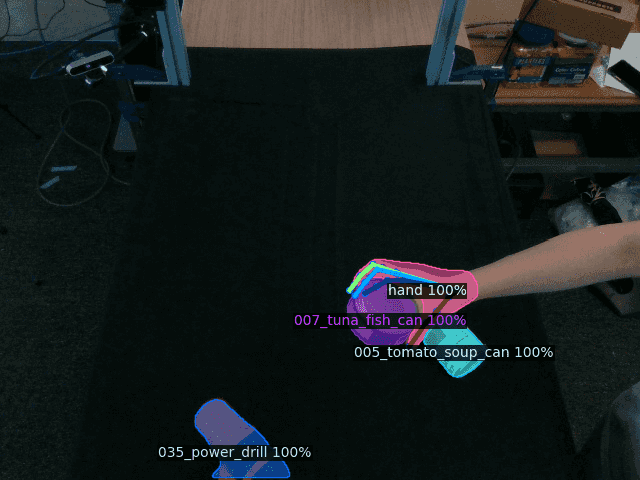}~
  \includegraphics[width=0.24\linewidth]{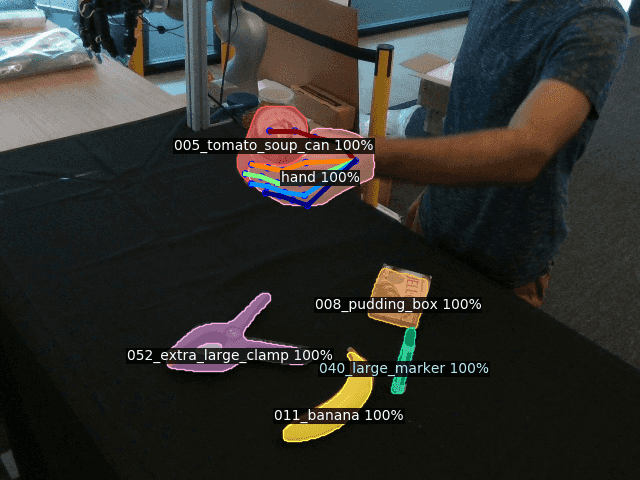}~
  \includegraphics[width=0.24\linewidth]{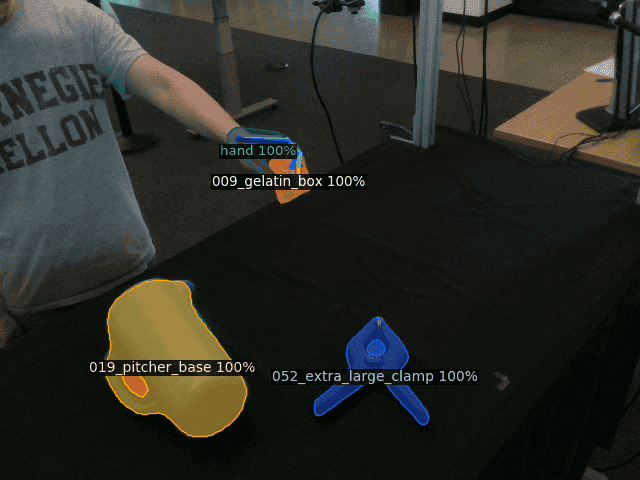}~
 \end{minipage}
 \\ \vspace{1mm}
 \begin{minipage}{1\textwidth}
  \includegraphics[width=0.24\linewidth]{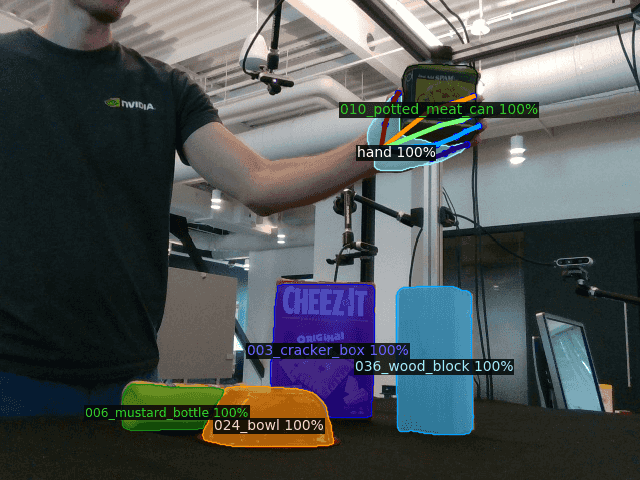}~
  \includegraphics[width=0.24\linewidth]{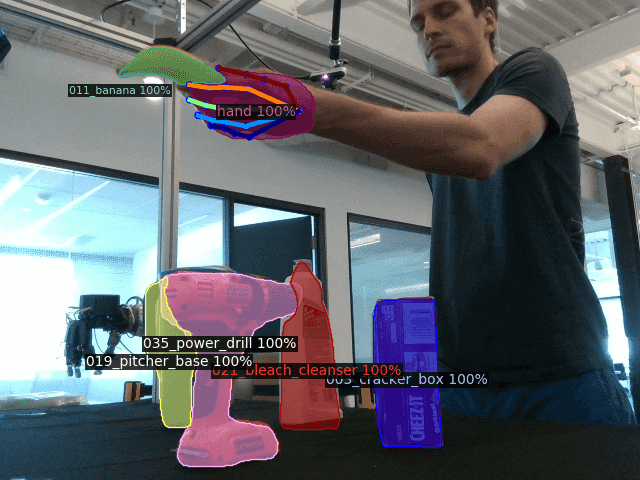}~
  \includegraphics[width=0.24\linewidth]{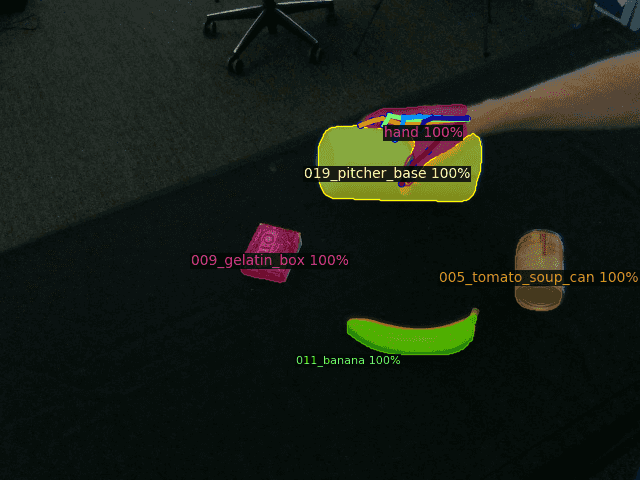}~
  \includegraphics[width=0.24\linewidth]{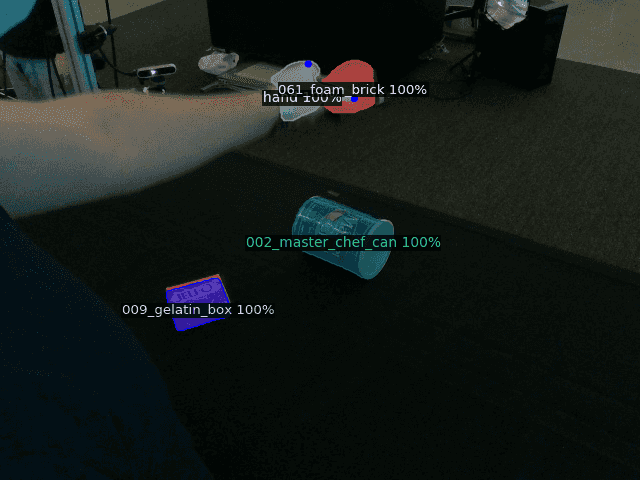}
 \end{minipage}
 \\ \vspace{1mm}
 \begin{minipage}{1\textwidth}
  \includegraphics[width=0.24\linewidth]{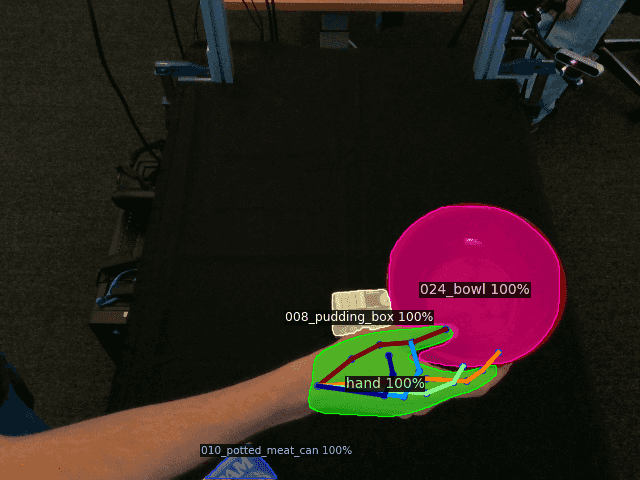}~
  \includegraphics[width=0.24\linewidth]{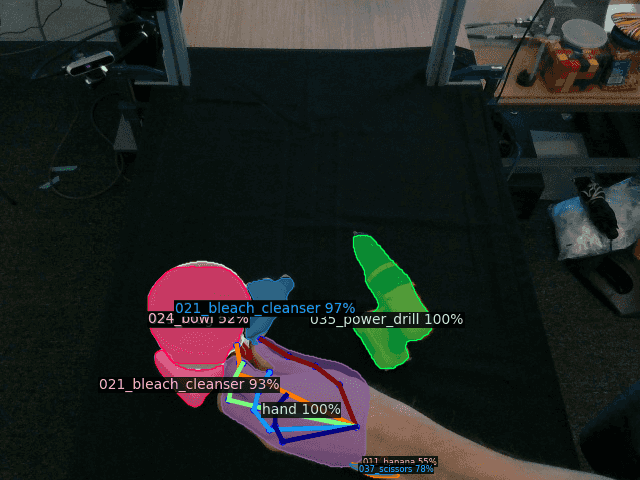}~
  \includegraphics[width=0.24\linewidth]{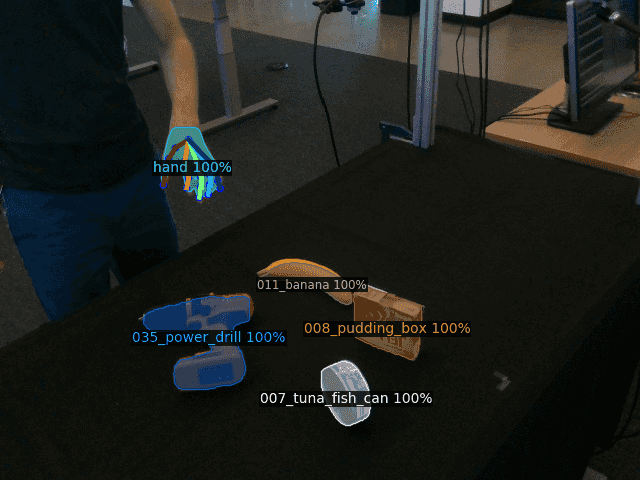}~
  \includegraphics[width=0.24\linewidth]{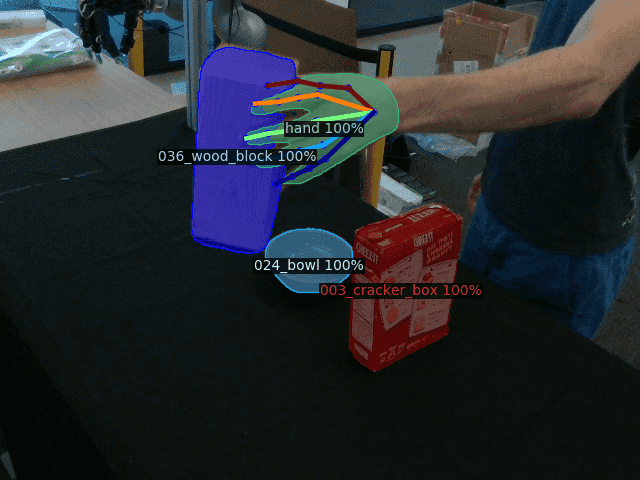}
 \end{minipage}
 \\ \vspace{1mm}
 \begin{minipage}{1\textwidth}
  \includegraphics[width=0.24\linewidth]{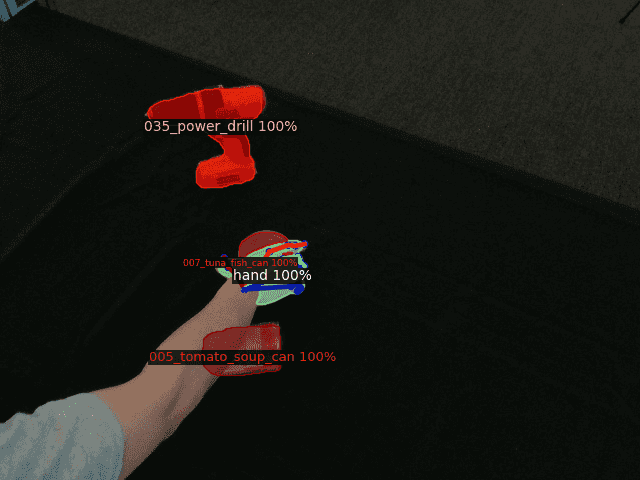}~
  \includegraphics[width=0.24\linewidth]{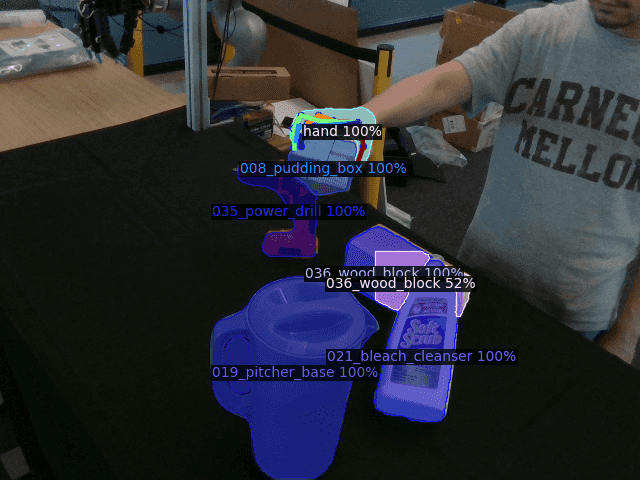}~
  \includegraphics[width=0.24\linewidth]{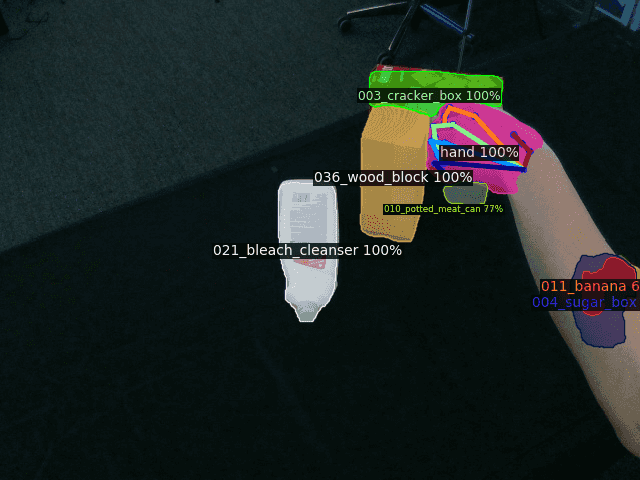}~
  \includegraphics[width=0.24\linewidth]{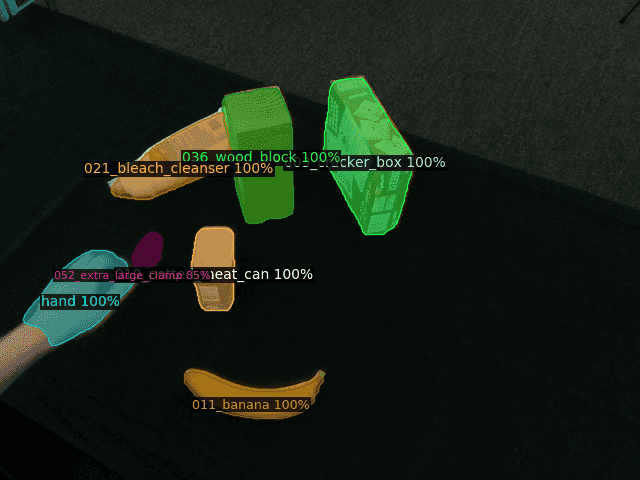}
 \end{minipage}
 \\ \vspace{1mm}
 \begin{minipage}{1\textwidth}
  \includegraphics[width=0.24\linewidth]{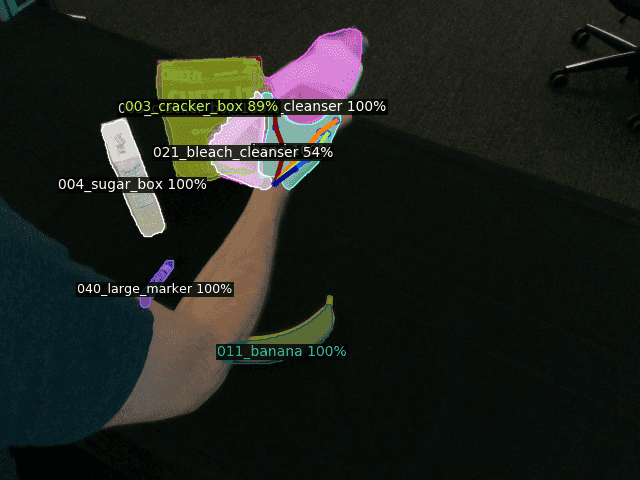}~
  \includegraphics[width=0.24\linewidth]{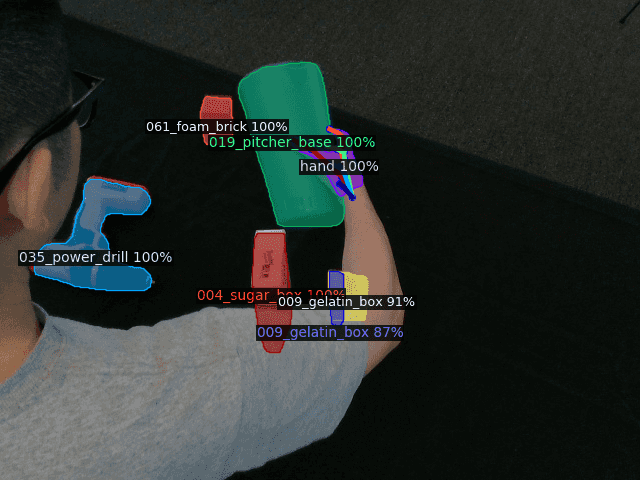}~
  \includegraphics[width=0.24\linewidth]{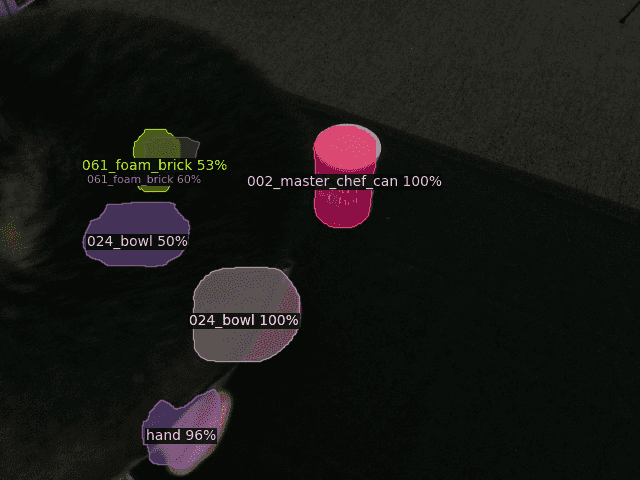}~
  \includegraphics[width=0.24\linewidth]{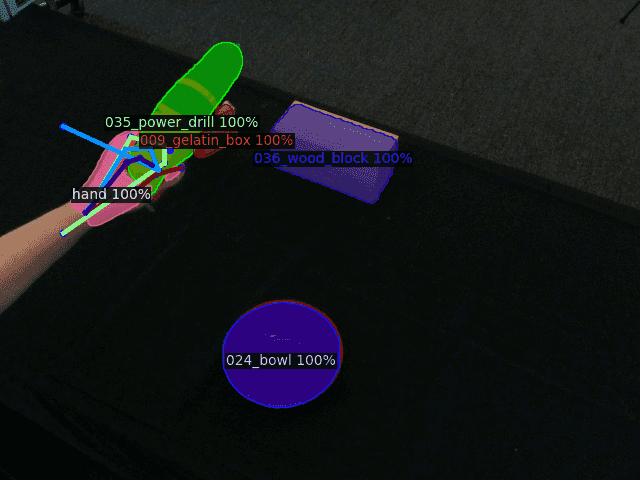}
 \end{minipage}
 \\ \vspace{2mm}
 \caption{\small Qualitative results of 2D object and keypoint detection with
Mask R-CNN (Detectron2))~\cite{he:iccv2017,wu2019detectron2}. The last two rows
highlight failure examples.}
 \label{fig:mask-rcnn}
\end{figure*}

Fig.~\ref{fig:mask-rcnn} shows qualitative results of 2D object detection and
keypoint detection with Mask R-CNN
(Detectron2)~\cite{he:iccv2017,wu2019detectron2} on the S0 setup. We highlight
some failure examples in the last two rows. In many failure cases we see false
object detections due to occlusions either by other objects (e.g.,
``036\_wood\_block'' in row 5 and column 2) or by hand interaction (e.g.,
``021\_bleach\_cleanser'' in row 6 and column 1). We also see inaccurately
detected hand keypoints when the hand is interacting with objects (e.g., row 6
and column 4).

\subsection{Qualitative: 6D Object Pose Estimation}

\begin{figure*}[t]
 \centering
 \begin{minipage}{0.15\textwidth}
  \centering
  Input RGB
 \end{minipage}
 \begin{minipage}{0.84\textwidth}
  \centering
  \includegraphics[width=0.24\linewidth]{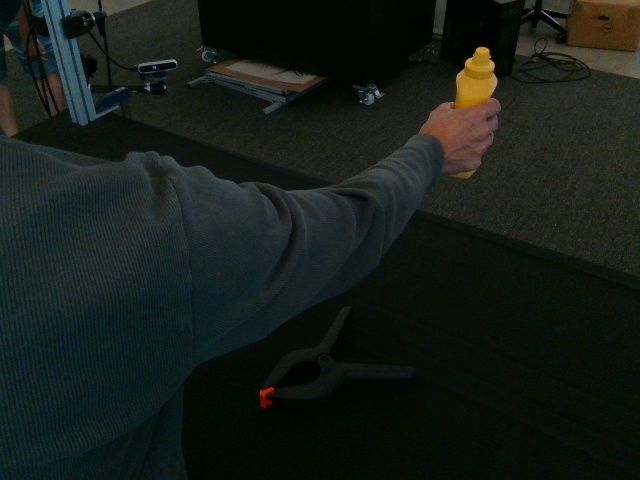}~
  \includegraphics[width=0.24\linewidth]{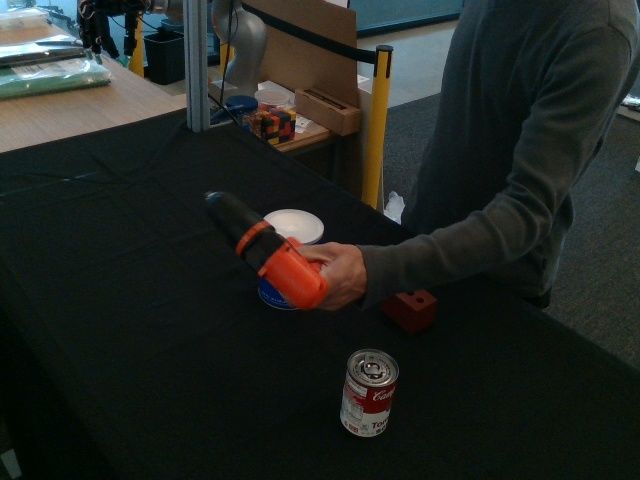}~
  \includegraphics[width=0.24\linewidth]{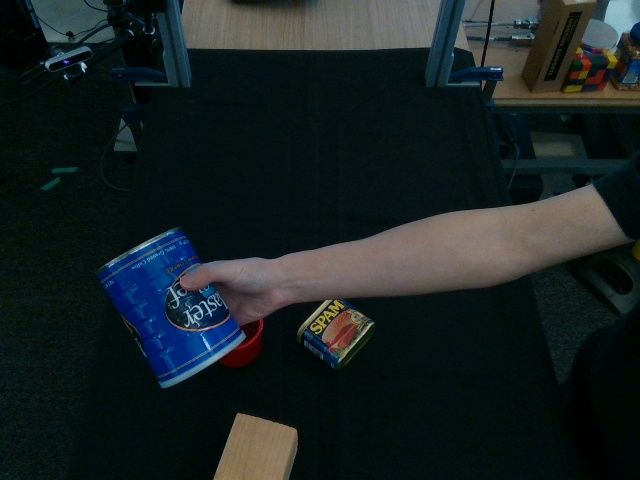}~
  \includegraphics[width=0.24\linewidth]{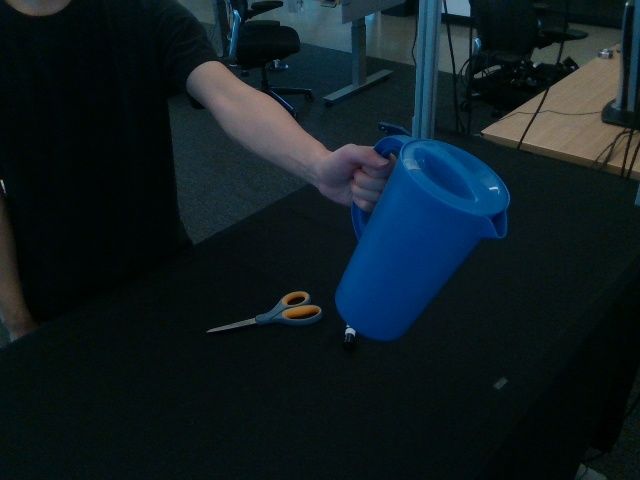}
 \end{minipage}
 \\ \vspace{1mm}
 \begin{minipage}{0.15\textwidth}
  \centering
  PoseCNN~\cite{xiang:rss2018}\\
  RGB
 \end{minipage}
 \begin{minipage}{0.84\textwidth}
  \centering
  \includegraphics[width=0.24\linewidth]{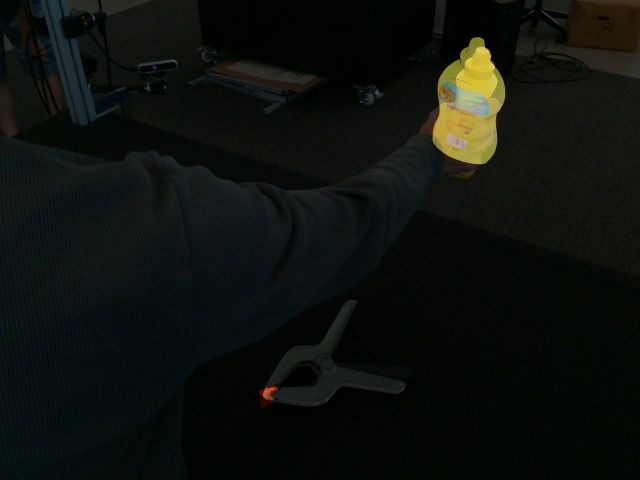}~
  \includegraphics[width=0.24\linewidth]{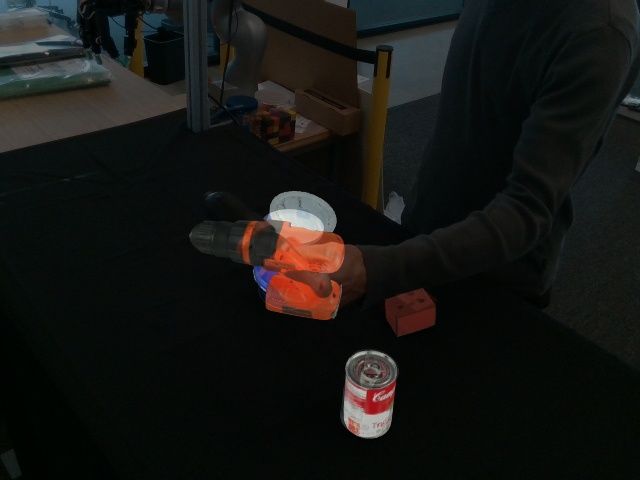}~
  \includegraphics[width=0.24\linewidth]{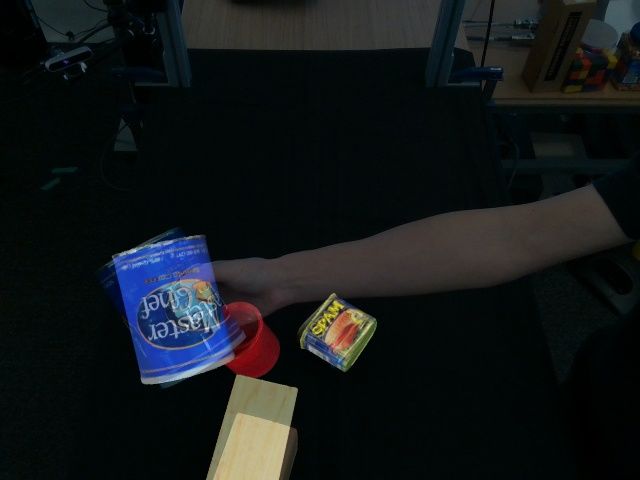}~
  \includegraphics[width=0.24\linewidth]{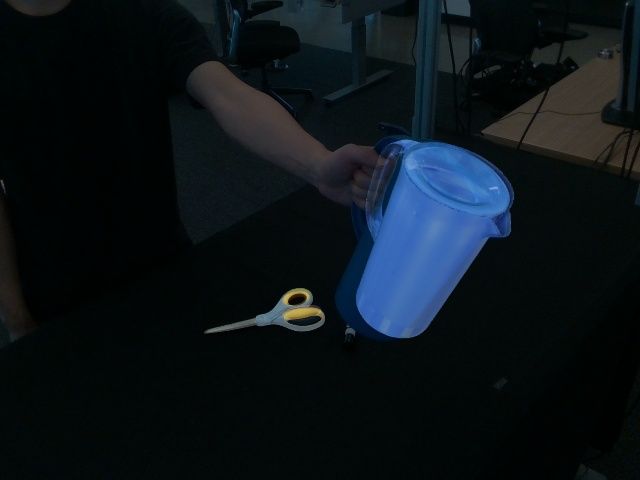}
 \end{minipage}
 \\ \vspace{1mm}
 \begin{minipage}{0.15\textwidth}
  \centering
  PoseCNN~\cite{xiang:rss2018}\\
  + depth ref
 \end{minipage}
 \begin{minipage}{0.84\textwidth}
  \centering
  \includegraphics[width=0.24\linewidth]{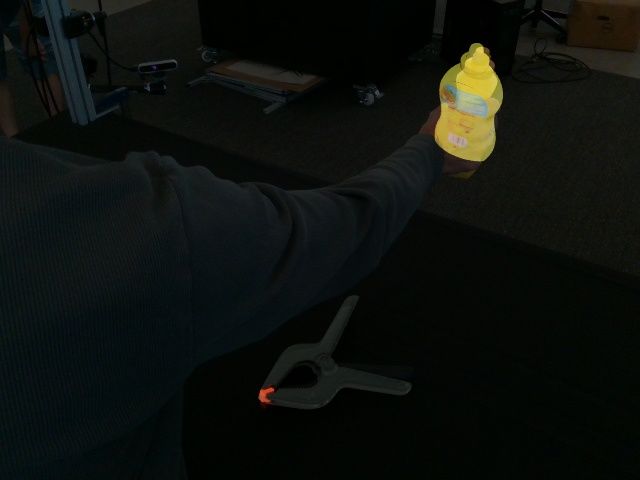}~
  \includegraphics[width=0.24\linewidth]{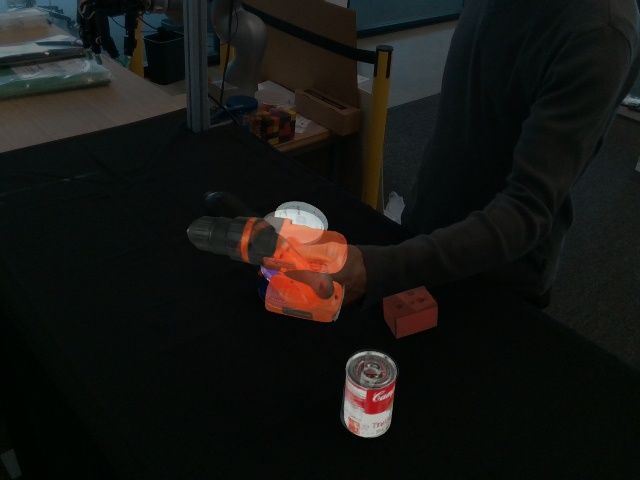}~
  \includegraphics[width=0.24\linewidth]{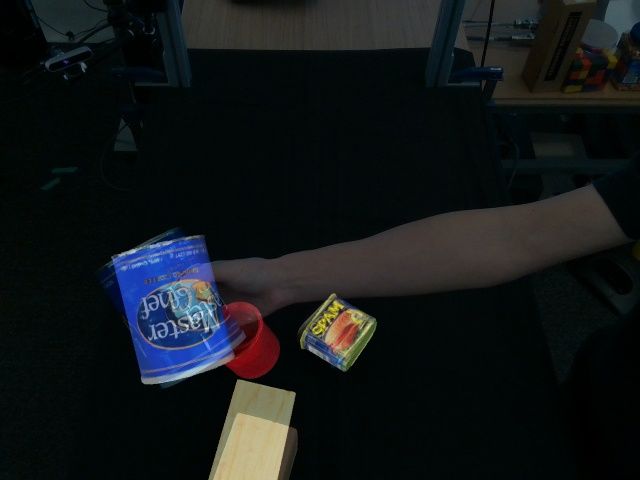}~
  \includegraphics[width=0.24\linewidth]{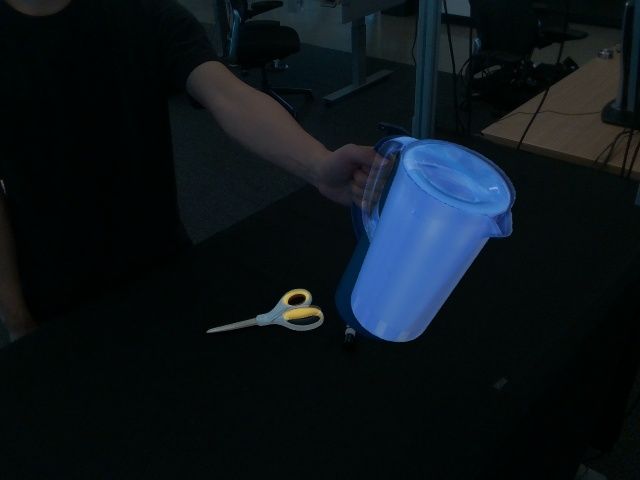}
 \end{minipage}
 \\ \vspace{1mm}
 \begin{minipage}{0.15\textwidth}
  \centering
  DeepIM~\cite{li:eccv2018}\\
  RGB
 \end{minipage}
 \begin{minipage}{0.84\textwidth}
  \centering
  \includegraphics[width=0.24\linewidth]{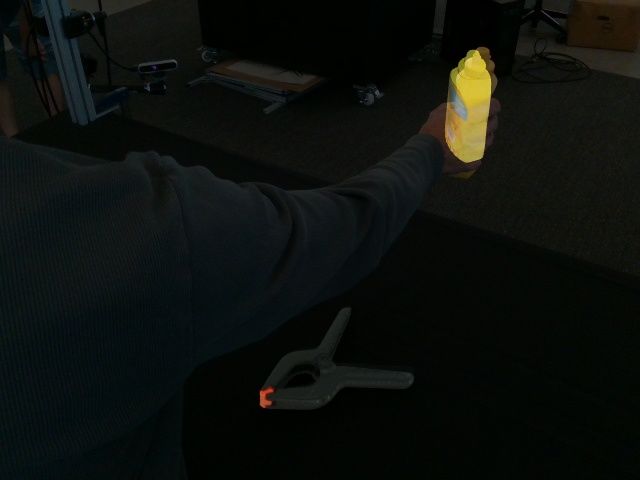}~
  \includegraphics[width=0.24\linewidth]{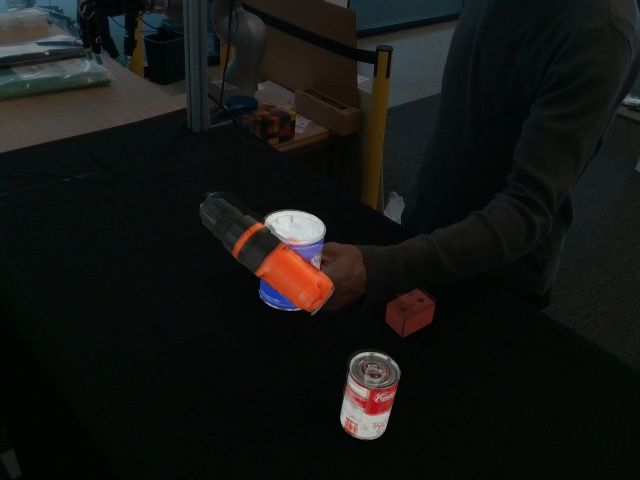}~
  \includegraphics[width=0.24\linewidth]{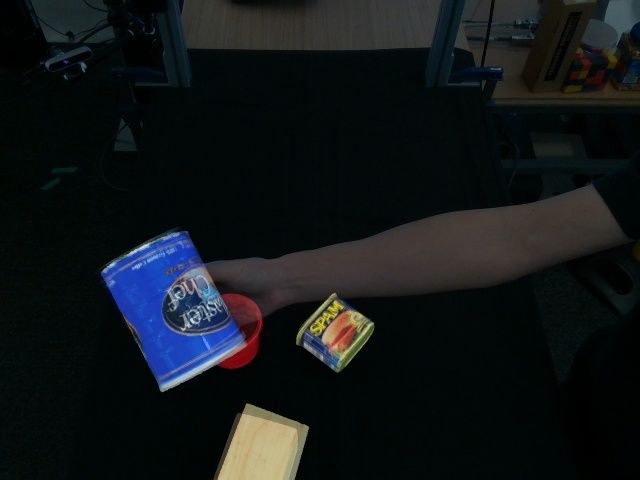}~
  \includegraphics[width=0.24\linewidth]{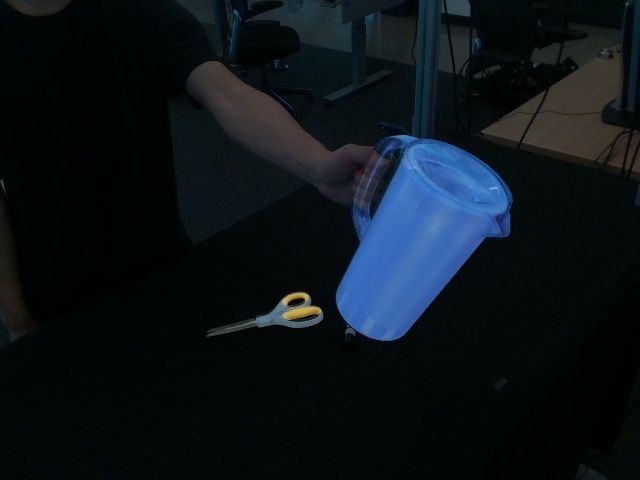}
 \end{minipage}
 \\ \vspace{1mm}
 \begin{minipage}{0.15\textwidth}
  \centering
  DeepIM~\cite{li:eccv2018}\\
  RGB-D
 \end{minipage}
 \begin{minipage}{0.84\textwidth}
  \centering
  \includegraphics[width=0.24\linewidth]{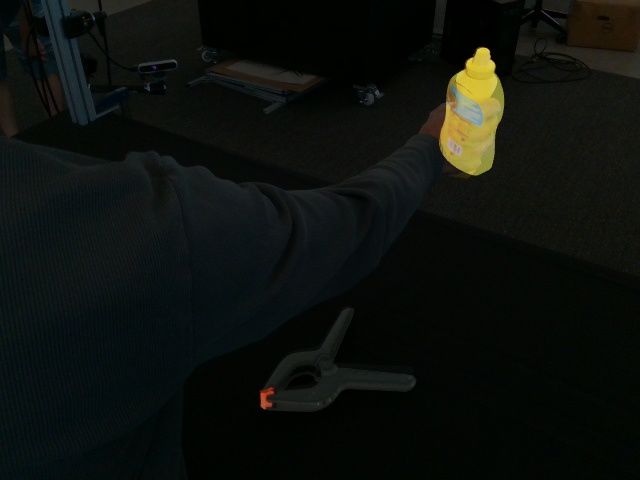}~
  \includegraphics[width=0.24\linewidth]{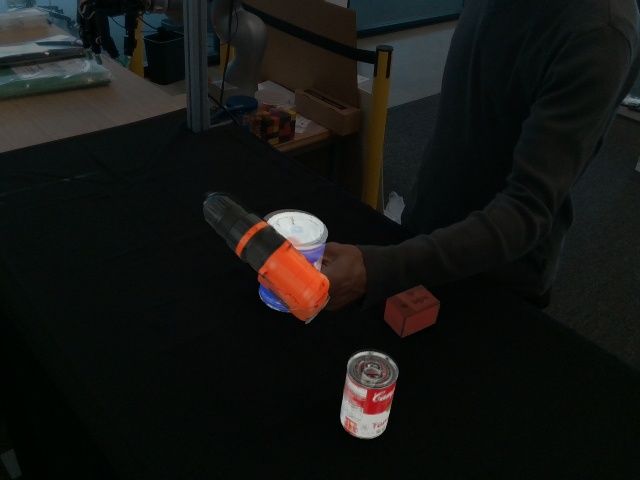}~
  \includegraphics[width=0.24\linewidth]{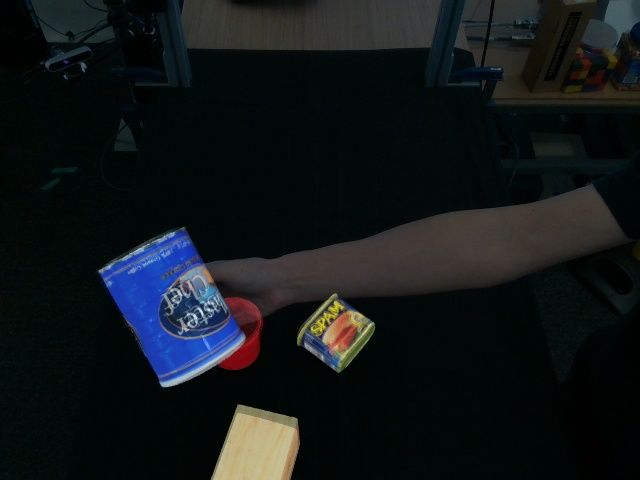}~
  \includegraphics[width=0.24\linewidth]{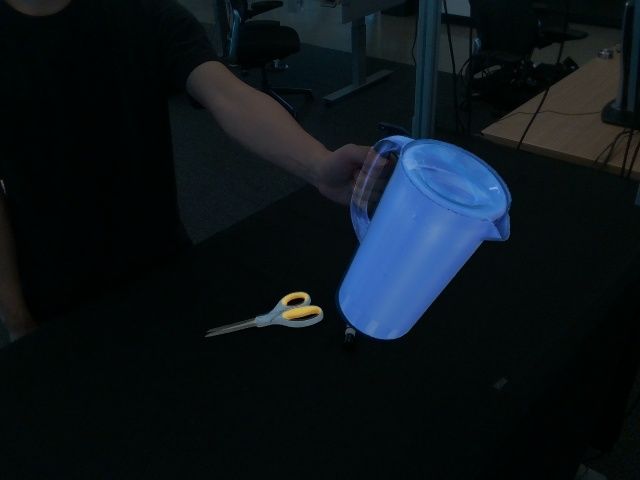}
 \end{minipage}
 \\ \vspace{1mm}
 \begin{minipage}{0.15\textwidth}
  \centering
  PoseRBPF~\cite{deng:rss2019}\\
  RGB
 \end{minipage}
 \begin{minipage}{0.84\textwidth}
  \centering
  \includegraphics[width=0.24\linewidth]{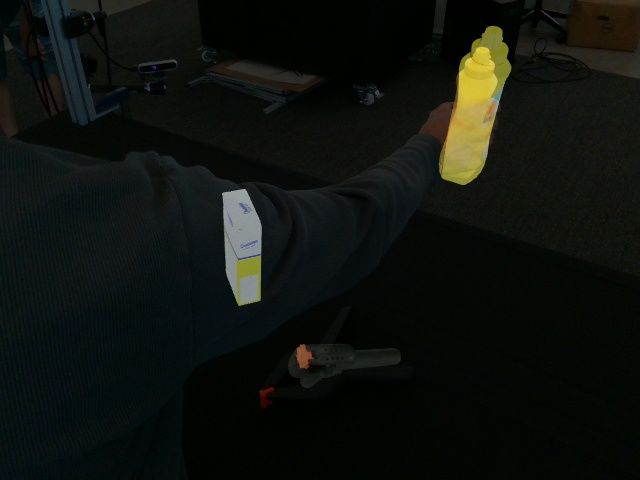}~
  \includegraphics[width=0.24\linewidth]{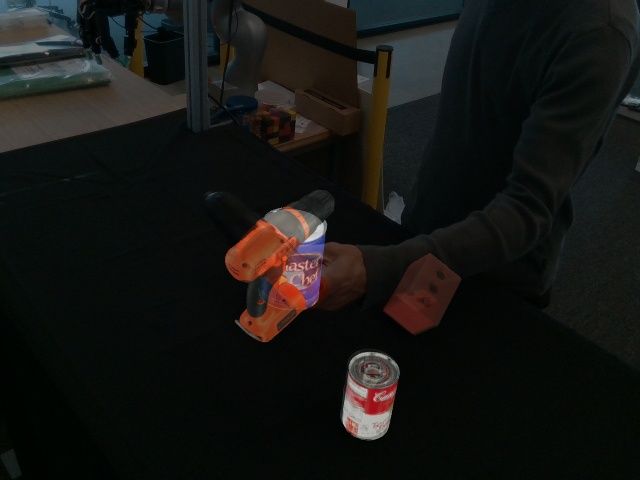}~
  \includegraphics[width=0.24\linewidth]{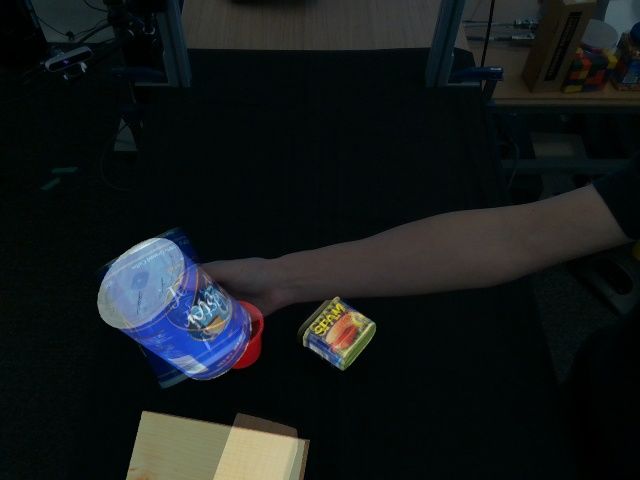}~
  \includegraphics[width=0.24\linewidth]{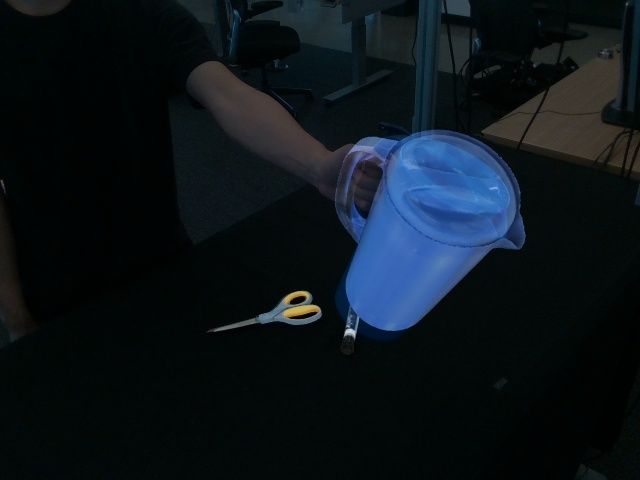}
 \end{minipage}
 \\ \vspace{1mm}
 \begin{minipage}{0.15\textwidth}
  \centering
  PoseRBPF~\cite{deng:rss2019}\\
  RGB-D
 \end{minipage}
 \begin{minipage}{0.84\textwidth}
  \centering
  \includegraphics[width=0.24\linewidth]{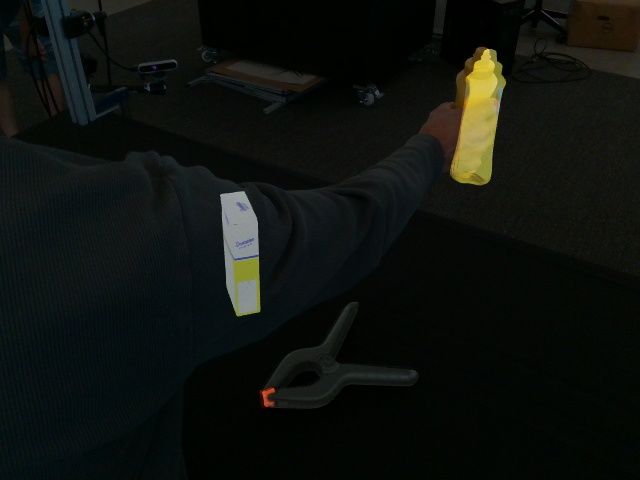}~
  \includegraphics[width=0.24\linewidth]{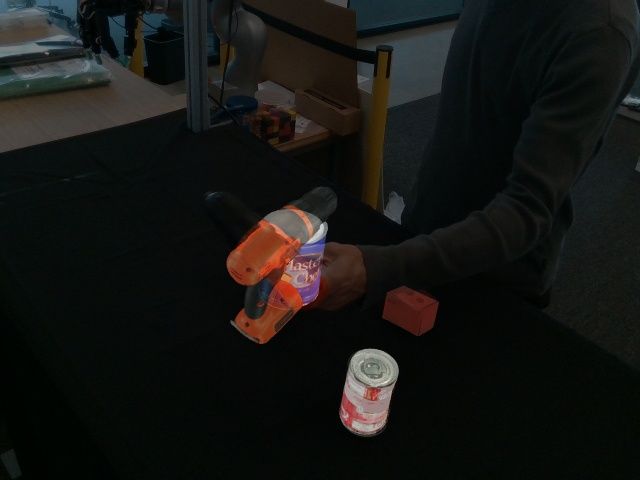}~
  \includegraphics[width=0.24\linewidth]{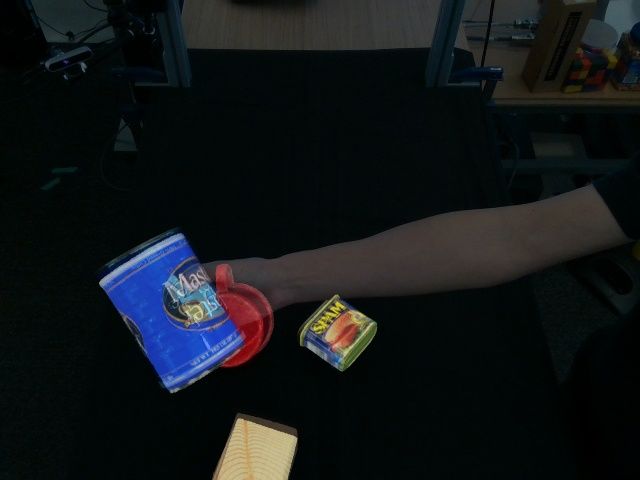}~
  \includegraphics[width=0.24\linewidth]{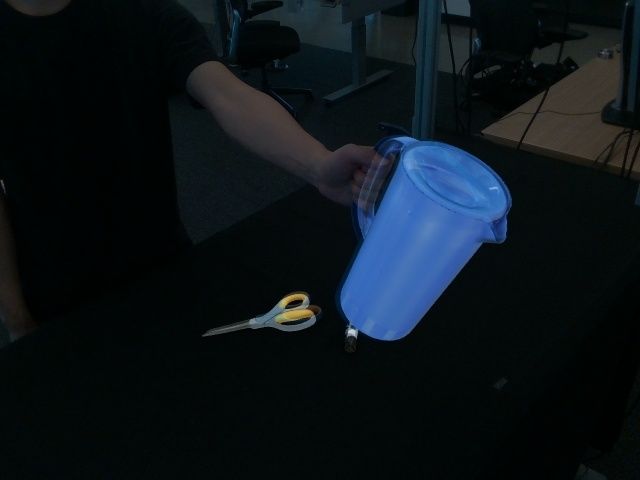}
 \end{minipage}
 \\ \vspace{1mm}
 \begin{minipage}{0.15\textwidth}
  \centering
  CosyPose~\cite{labbe:eccv2020}\\
  RGB
 \end{minipage}
 \begin{minipage}{0.84\textwidth}
  \centering
  \includegraphics[width=0.24\linewidth]{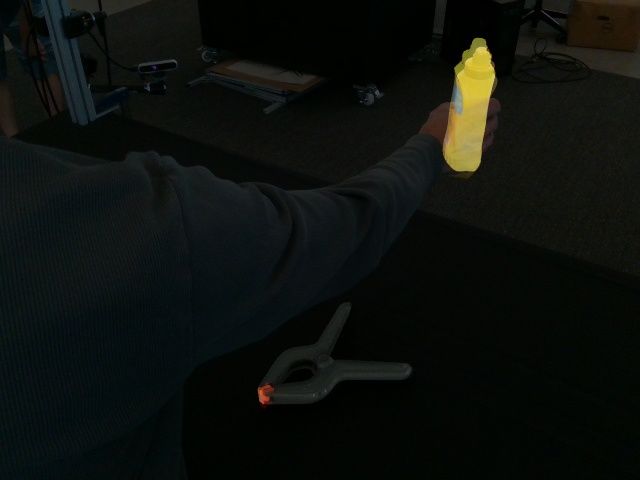}~
  \includegraphics[width=0.24\linewidth]{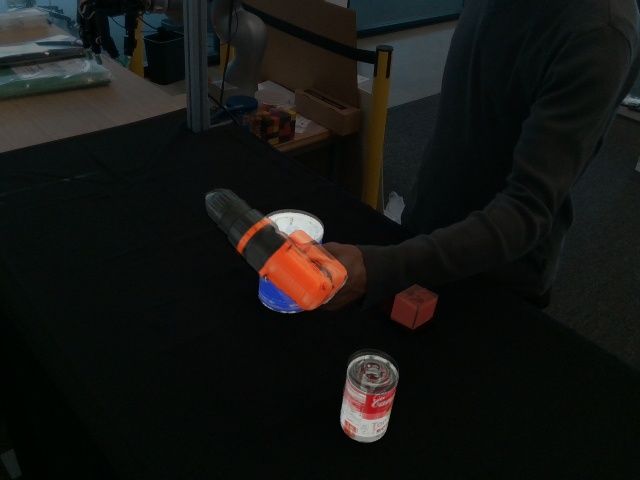}~
  \includegraphics[width=0.24\linewidth]{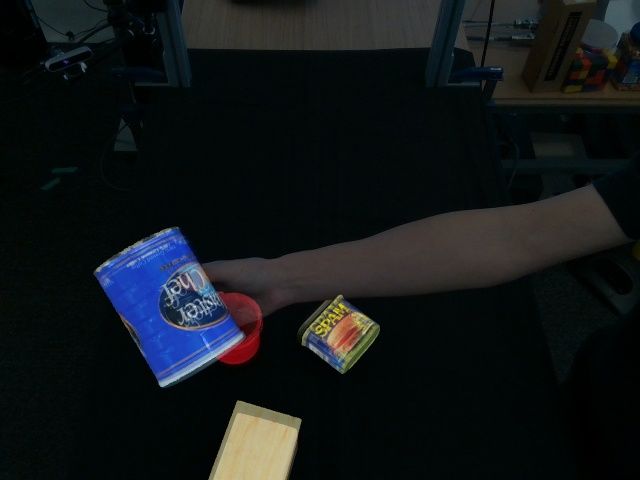}~
  \includegraphics[width=0.24\linewidth]{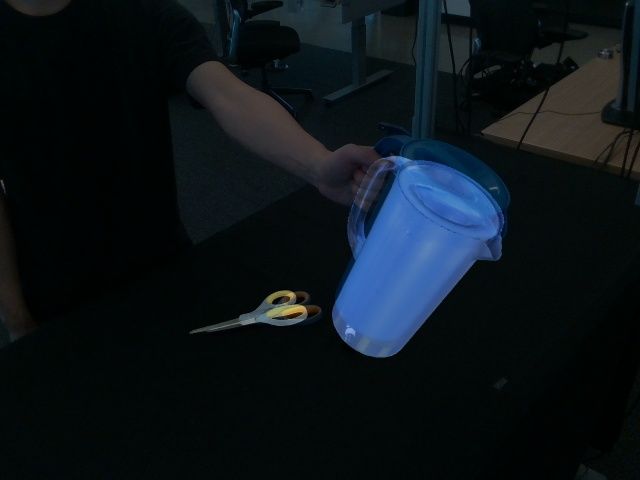}
 \end{minipage}
 \\ \vspace{2mm}
 \caption{\small Qualitative results of 6D object pose estimation. We render
object models given the estimated poses on a darkened input image.}
 \label{fig:bop}
\end{figure*}

Fig.~\ref{fig:bop} shows qualitative results of 6D object pose estimation on
the S1 setup. The first row shows the input RGB images and the following rows
show the estimated pose from each representative approach. We render object
models given the estimated poses and overlay them on top of a darkened input
image. We can see the challenge when the object is severely occluded by the
hand (e.g., the leftmost example of PoseCNN~\cite{xiang:rss2018} (RGB)). We
also see that refinement based approaches like DeepIM~\cite{li:eccv2018} and
CosyPose~\cite{labbe:eccv2020} are able to improve upon their coarse estimate
input (i.e., PoseCNN (RGB)) and generate more accurate final predictions (e.g.,
the second-left example of DeepIM (RGB)).

\begin{table*}[t]
 \centering
 \small
 \begin{tabular}{l|ccccccc}
                           & \multicolumn{7}{c}{S0 (default)} \\ \cline{2-8}
                           & \multicolumn{2}{c}{PoseCNN~\cite{xiang:rss2018}} & \multicolumn{2}{c}{DeepIM~\cite{li:eccv2018}} & \multicolumn{2}{c}{PoseRBPF~\cite{deng:rss2019}} & CosyPose~\cite{labbe:eccv2020} \\
                           & RGB & + depth ref & RGB & RGB-D & RGB & RGB-D & RGB \\
  \hline
  002\_master\_chef\_can   & 47.47 & 51.04 & 63.53 & 71.69          & 36.70 & 55.23          & \textbf{78.97} \\
  003\_cracker\_box        & 61.04 & 64.44 & 79.82 & 86.53          & 48.54 & 68.47          & \textbf{88.74} \\
  004\_sugar\_box          & 45.11 & 49.90 & 59.30 & 70.98          & 34.44 & 58.95          & \textbf{78.45} \\
  005\_tomato\_soup\_can   & 36.68 & 46.23 & 52.84 & \textbf{61.28} & 29.54 & 42.60          & 57.89          \\
  006\_mustard\_bottle     & 52.56 & 56.10 & 60.92 & 71.00          & 33.86 & 59.20          & \textbf{71.57} \\
  007\_tuna\_fish\_can     & 32.70 & 36.37 & 39.78 & \textbf{46.81} & 19.14 & 35.03          & 46.31          \\
  008\_pudding\_box        & 44.24 & 51.72 & 56.55 & 68.26          & 29.64 & 52.34          & \textbf{68.52} \\
  009\_gelatin\_box        & 46.62 & 56.15 & 62.49 & \textbf{72.60} & 32.08 & 49.88          & 70.38          \\
  010\_potted\_meat\_can   & 37.41 & 43.42 & 55.48 & \textbf{63.33} & 32.98 & 44.39          & 59.23          \\
  011\_banana              & 38.33 & 42.41 & 46.47 & \textbf{53.30} & 22.53 & 42.96          & 38.01          \\
  019\_pitcher\_base       & 53.49 & 56.39 & 60.01 & \textbf{71.27} & 18.66 & 59.32          & 55.43          \\
  021\_bleach\_cleanser    & 49.41 & 54.87 & 59.13 & \textbf{71.73} & 32.03 & 60.40          & 68.80          \\
  024\_bowl                & 57.42 & 58.90 & 63.92 & 73.15          & 34.35 & 61.07          & \textbf{77.12} \\
  025\_mug                 & 40.68 & 43.52 & 49.17 & \textbf{58.27} & 21.88 & 37.69          & 54.70          \\
  035\_power\_drill        & 47.93 & 51.59 & 62.10 & \textbf{72.30} & 37.54 & 58.86          & 52.02          \\
  036\_wood\_block         & 40.11 & 46.76 & 51.77 & 68.28          & 25.29 & 53.23          & \textbf{68.46} \\
  037\_scissors            & 21.93 & 24.11 & 27.57 & \textbf{33.59} & 22.32 & 32.31          & 23.82          \\
  040\_large\_marker       & 35.09 & 42.49 & 35.35 & 49.04          & 25.05 & 36.03          & \textbf{53.61} \\
  052\_extra\_large\_clamp & 30.48 & 34.75 & 39.94 & 49.96          & 22.07 & 36.74          & \textbf{52.18} \\
  061\_foam\_brick         & 12.80 & 15.96 & 17.50 & 25.82          & 18.33 & \textbf{35.10} & 34.34          \\
  \hline
  all                      & 41.65 & 46.40 & 52.25 & \textbf{62.03} & 28.90 & 49.09          & 59.99 
 \end{tabular}
 \vspace{-2mm}
 \caption{\small 6D object pose estimation results of representative approaches
in AR (\%) on S0.}
 \label{tab:6d:sota:s0}
\end{table*}

\subsection{Qualitative: 3D Hand Pose Estimation}

\begin{figure*}[t]
 \centering
 \includegraphics[width=0.16\linewidth]{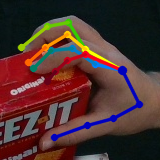}~
 \includegraphics[width=0.16\linewidth,trim=12cm 0cm 17cm 5cm ,clip]{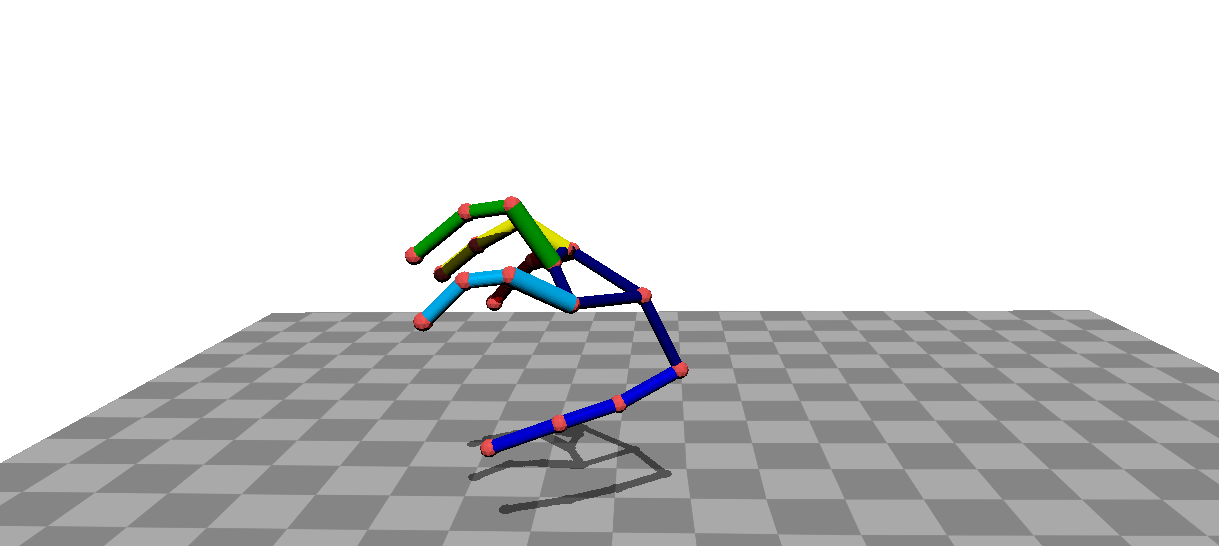}~
 \includegraphics[width=0.16\linewidth,trim=10cm 5cm 23cm 5cm ,clip]{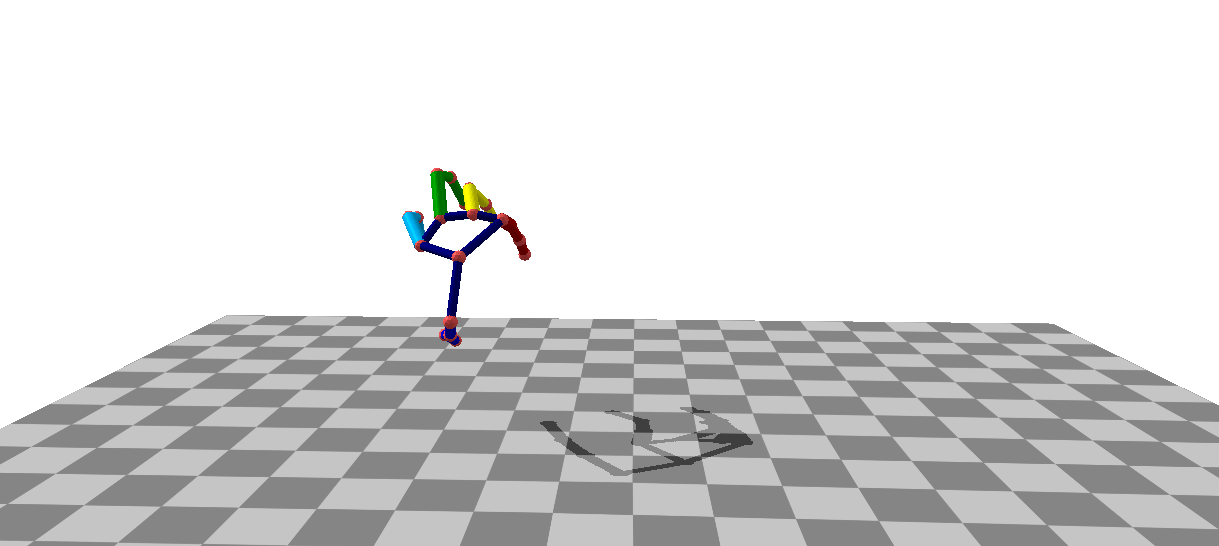}~
 \includegraphics[width=0.16\linewidth]{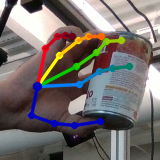}~
 \includegraphics[width=0.16\linewidth,trim=17cm 0cm 12cm 5cm ,clip]{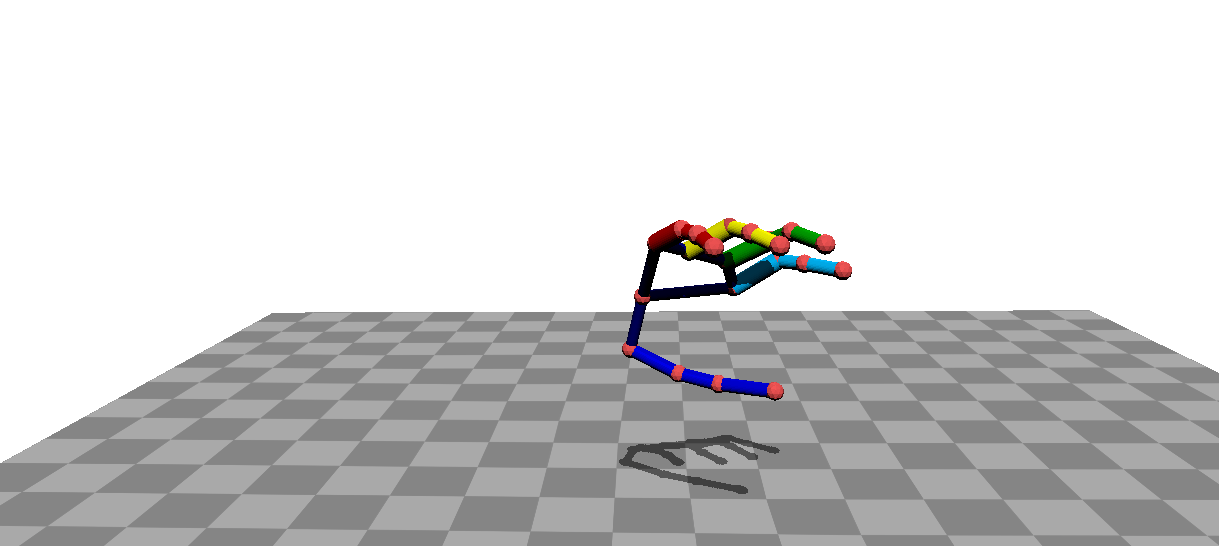}~
 \includegraphics[width=0.16\linewidth,trim=10cm 5cm 23cm 5cm ,clip]{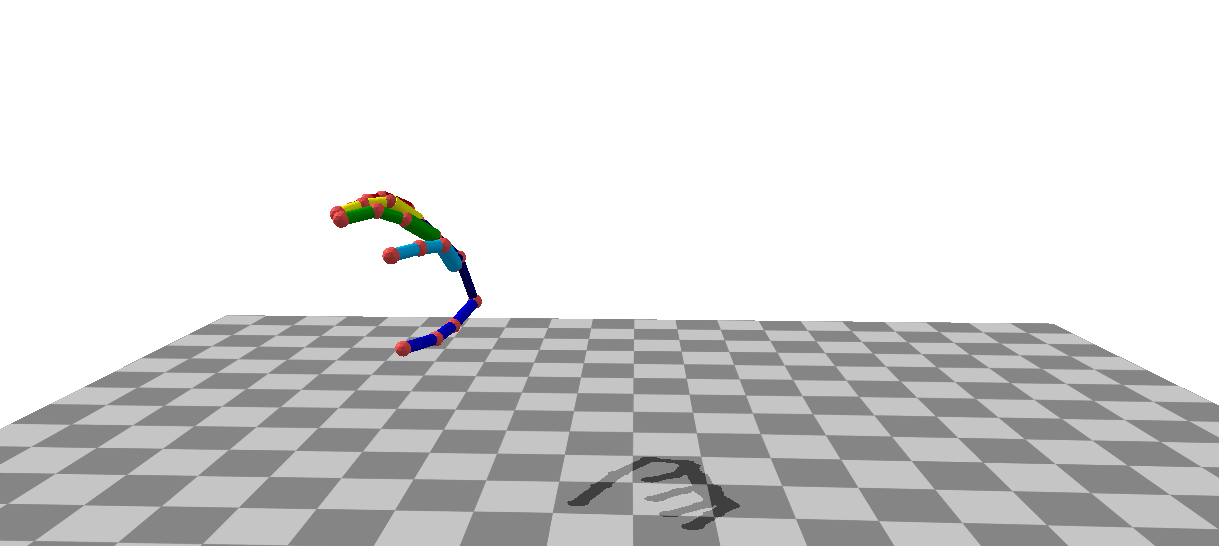}
 \\ \vspace{1mm}
 \includegraphics[width=0.16\linewidth]{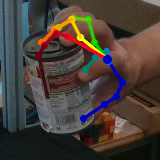}~
 \includegraphics[width=0.16\linewidth,trim=12cm 0cm 17cm 5cm ,clip]{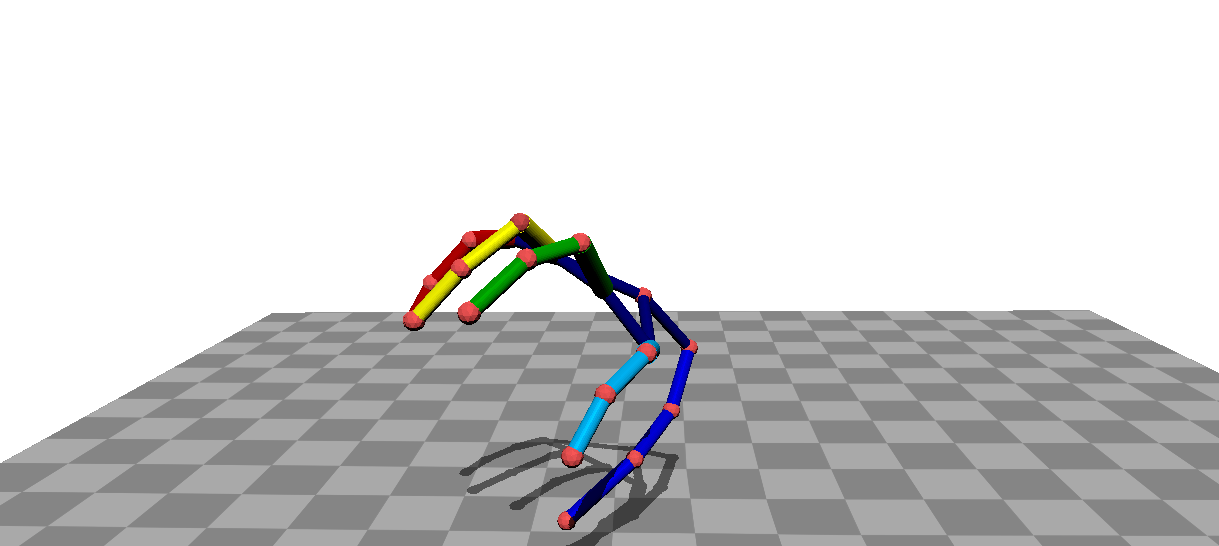}~
 \includegraphics[width=0.16\linewidth,trim=10cm 5cm 23cm 5cm ,clip]{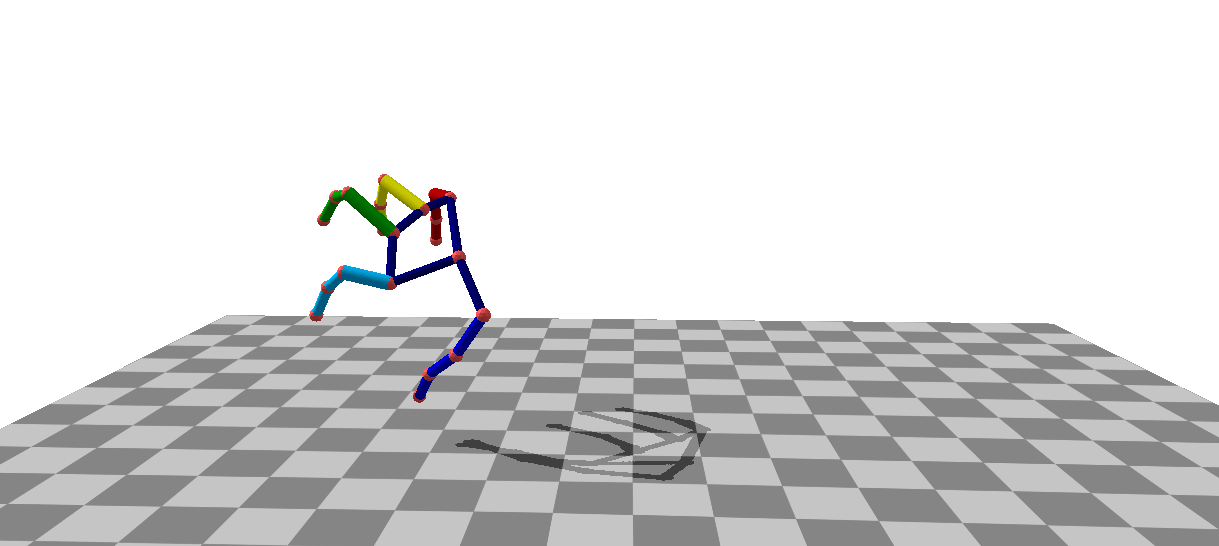}~
 \includegraphics[width=0.16\linewidth]{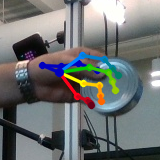}~
 \includegraphics[width=0.16\linewidth,trim=17cm 0cm 12cm 5cm ,clip]{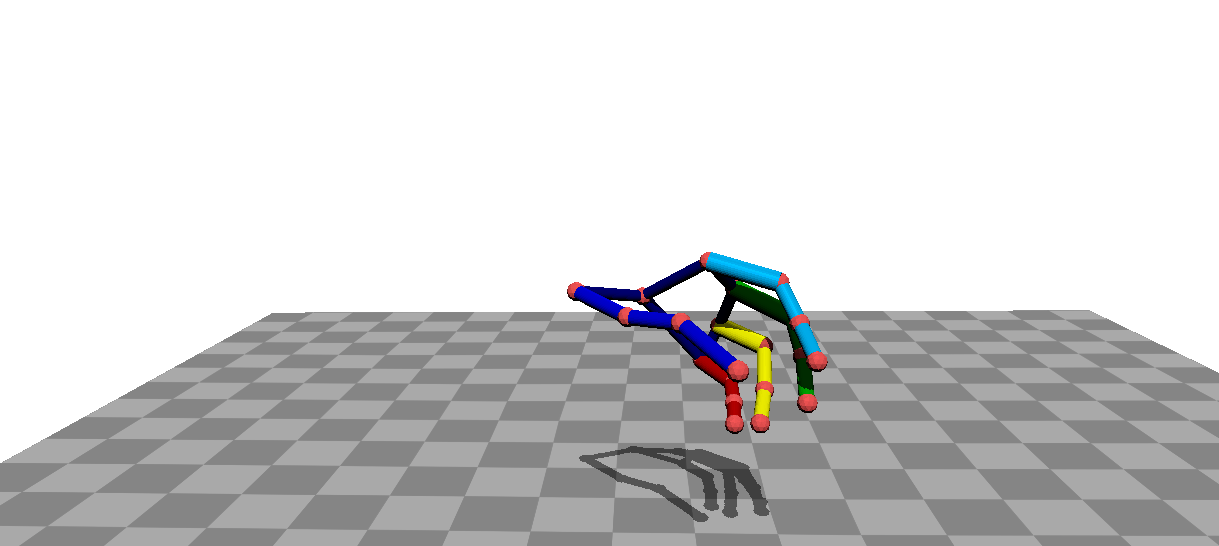}~
 \includegraphics[width=0.16\linewidth,trim=10cm 5cm 23cm 5cm ,clip]{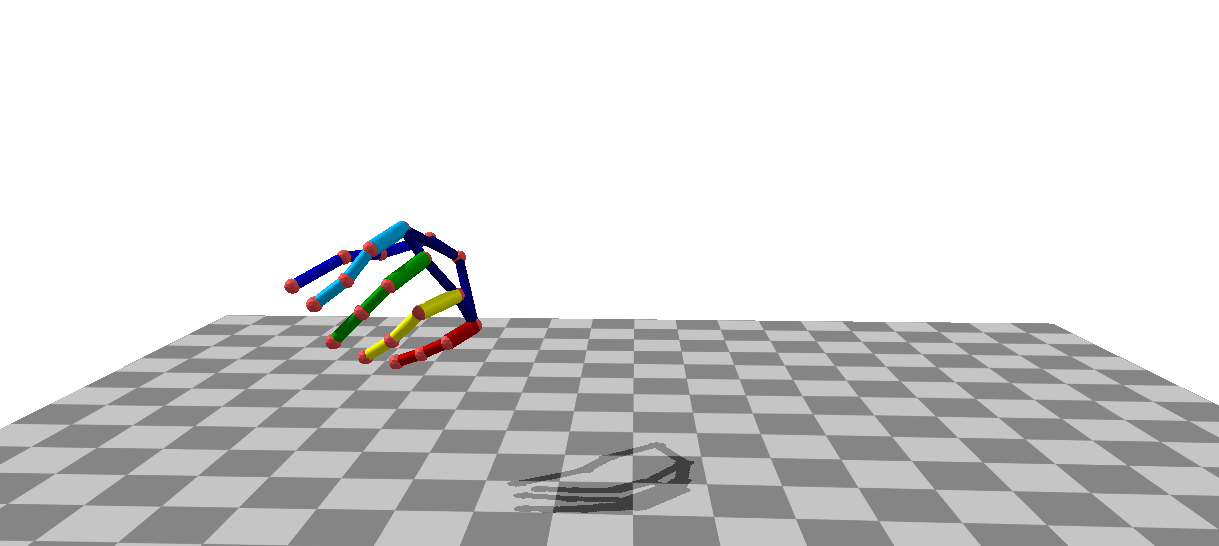}
 \\ \vspace{1mm}
 \includegraphics[width=0.16\linewidth]{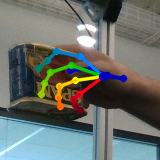}~
 \includegraphics[width=0.16\linewidth,trim=12cm 0cm 17cm 5cm ,clip]{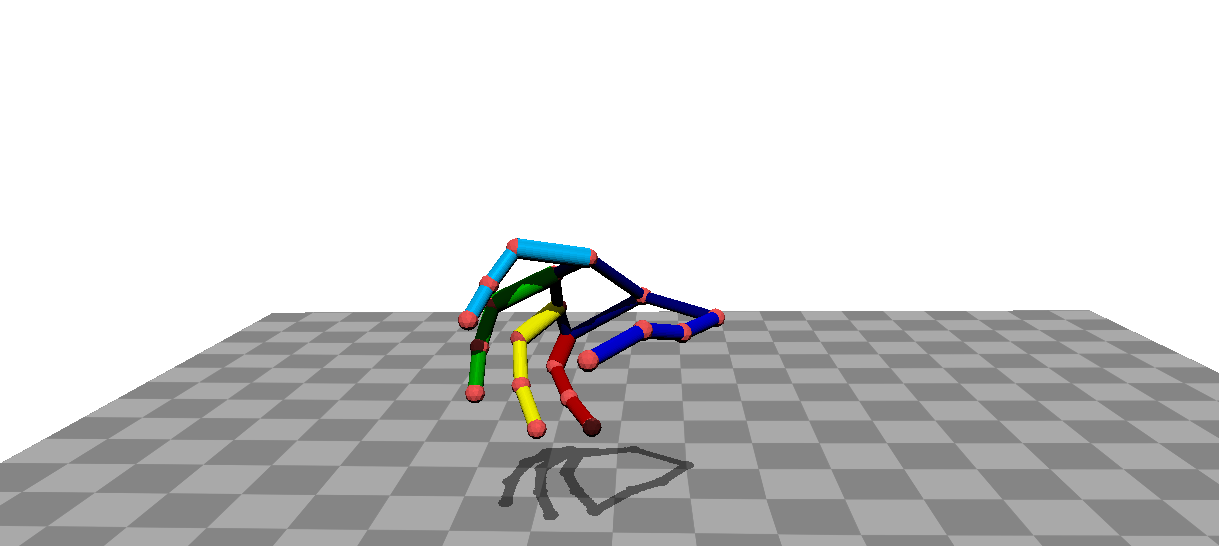}~
 \includegraphics[width=0.16\linewidth,trim=10cm 5cm 23cm 5cm ,clip]{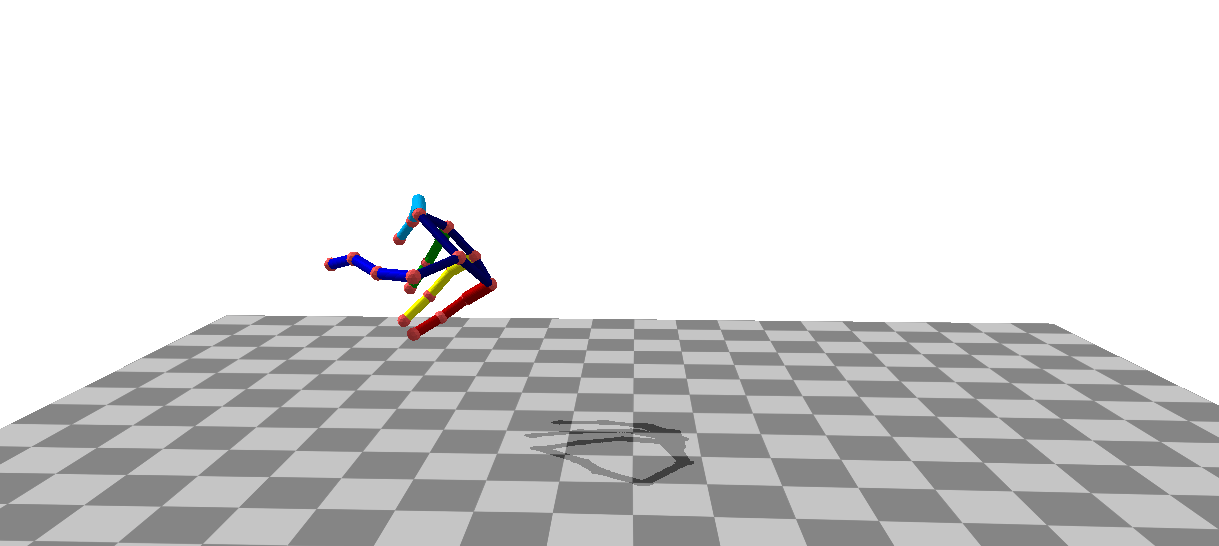}~
 \includegraphics[width=0.16\linewidth]{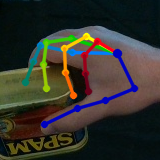}~
 \includegraphics[width=0.16\linewidth,trim=12cm 0cm 17cm 5cm ,clip]{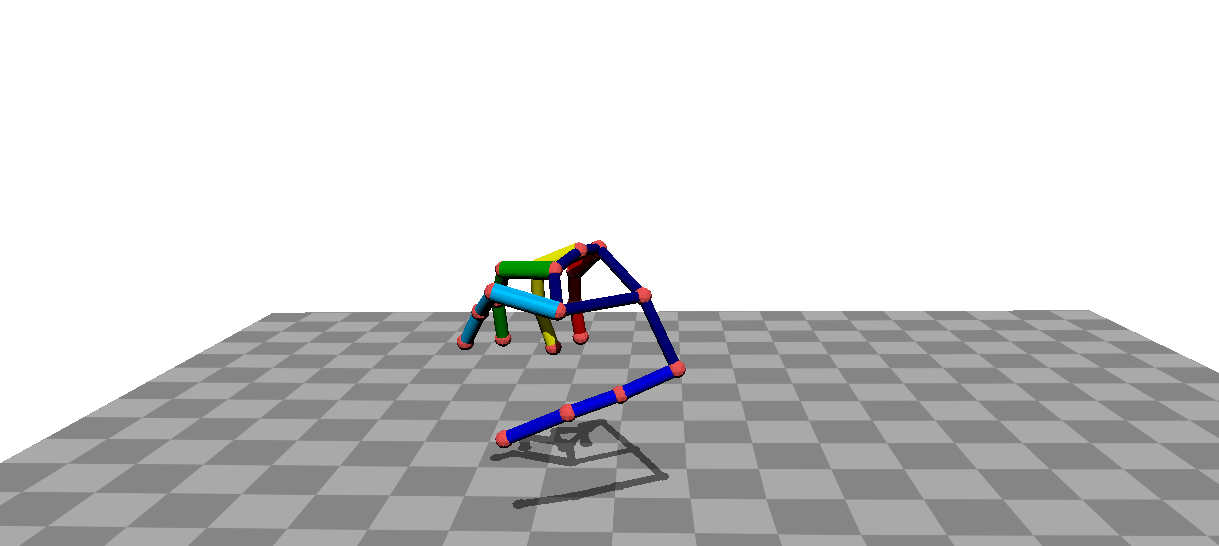}~
 \includegraphics[width=0.16\linewidth,trim=10cm 5cm 23cm 5cm ,clip]{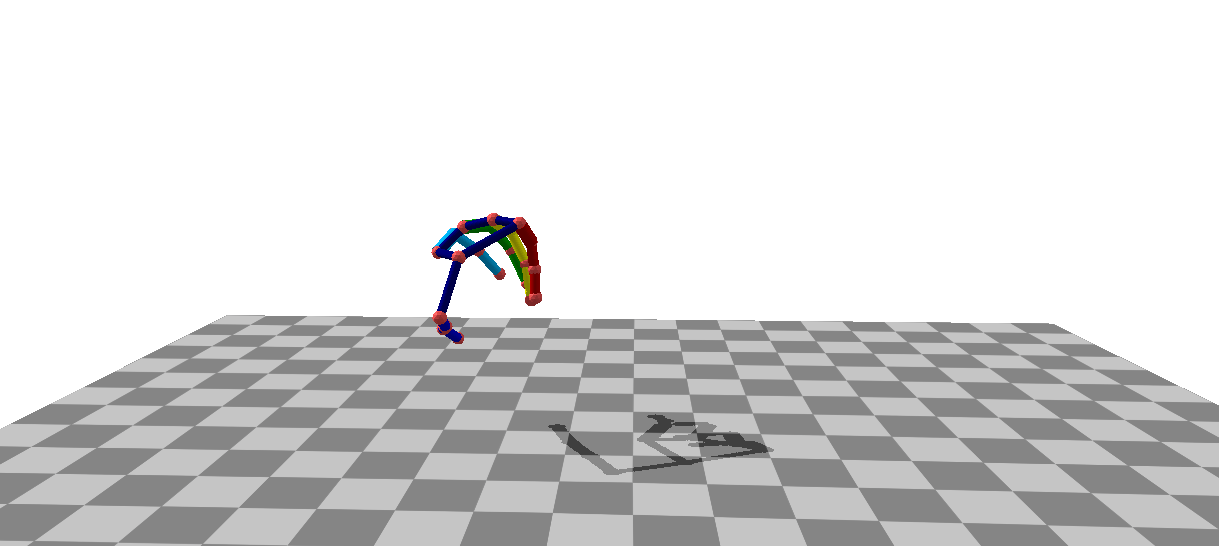}
 \\ \vspace{1mm}
 \includegraphics[width=0.16\linewidth]{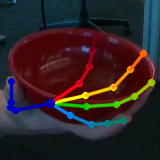}~
 \includegraphics[width=0.16\linewidth,trim=17cm 0cm 10cm 3cm ,clip]{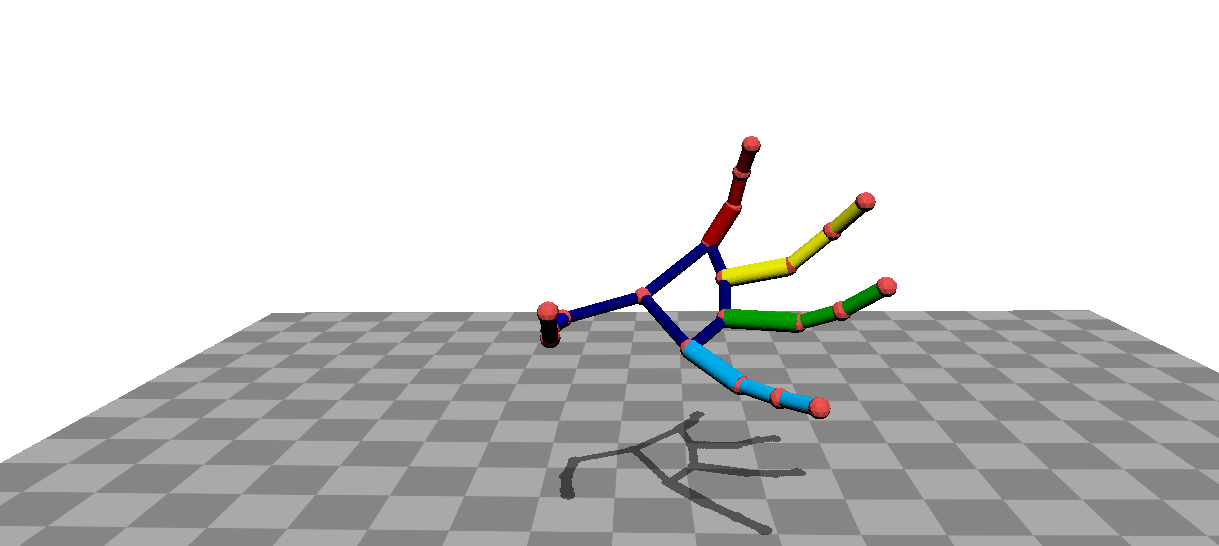}~
 \includegraphics[width=0.16\linewidth,trim=8cm 5cm 23cm 2cm ,clip]{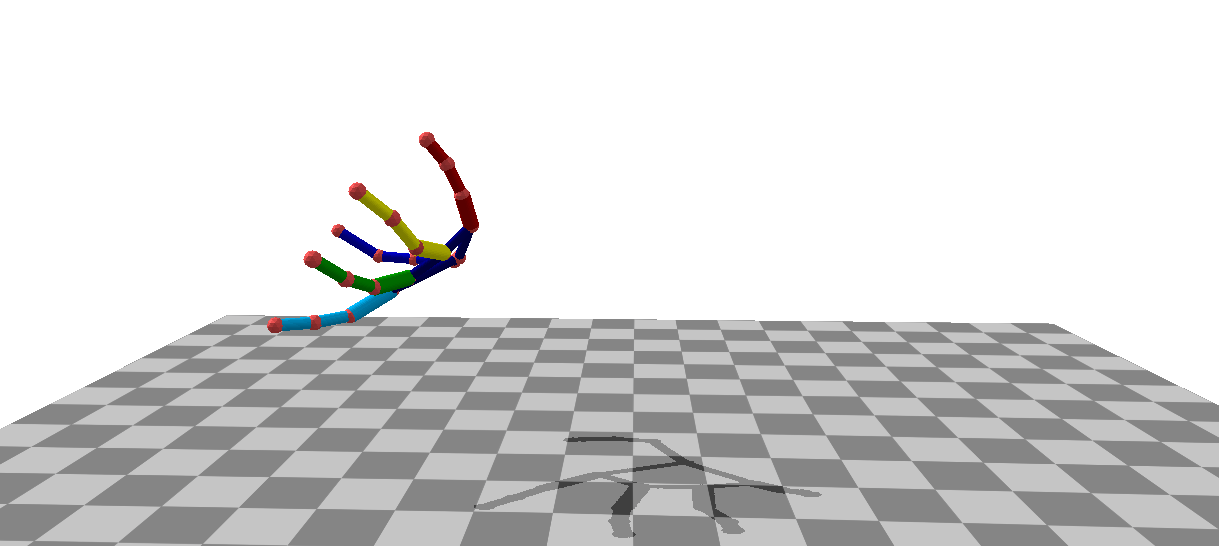}~
 \includegraphics[width=0.16\linewidth]{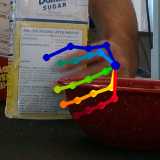}~
 \includegraphics[width=0.16\linewidth,trim=12cm 0cm 17cm 5cm ,clip]{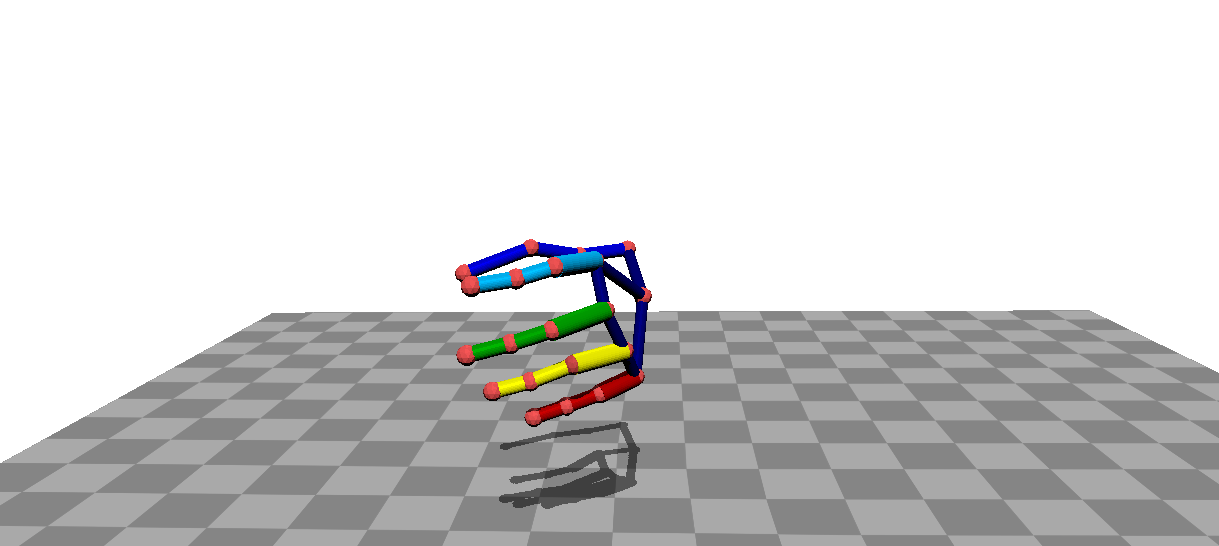}~
 \includegraphics[width=0.16\linewidth,trim=10cm 5cm 23cm 5cm ,clip]{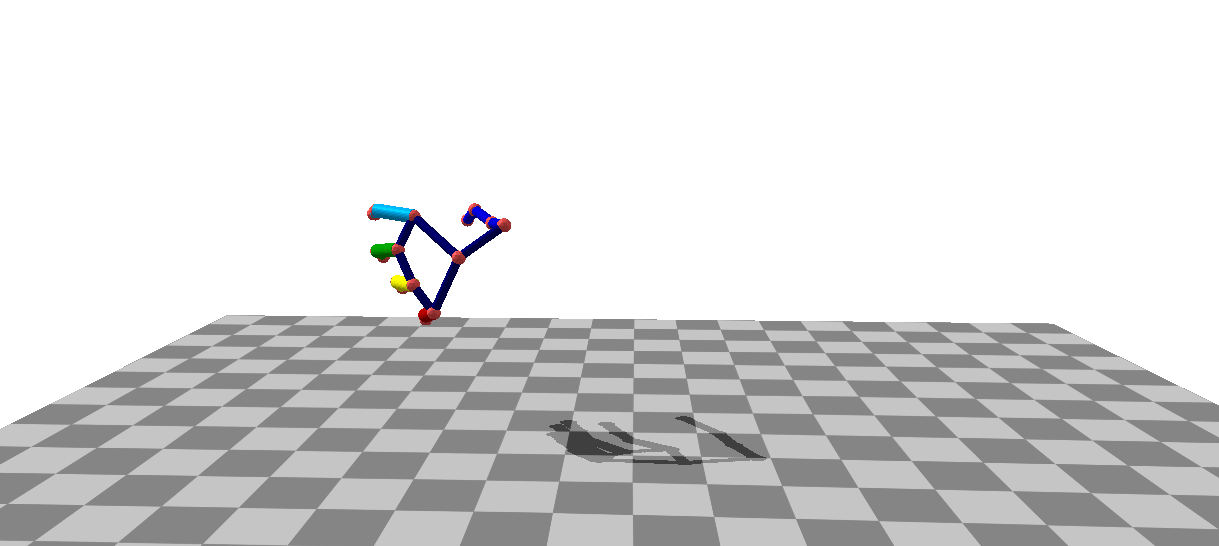}
 \\ \vspace{1mm}
 \includegraphics[width=0.16\linewidth]{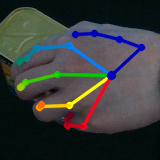}~
 \includegraphics[width=0.16\linewidth,trim=12cm 0cm 17cm 5cm ,clip]{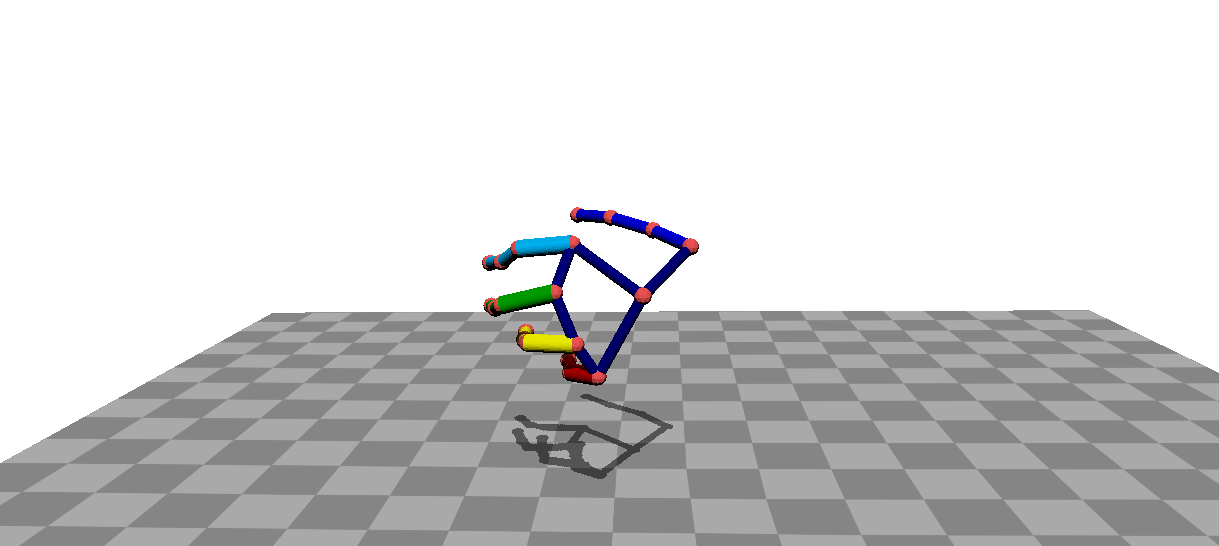}~
 \includegraphics[width=0.16\linewidth,trim=15cm 5cm 19cm 5cm ,clip]{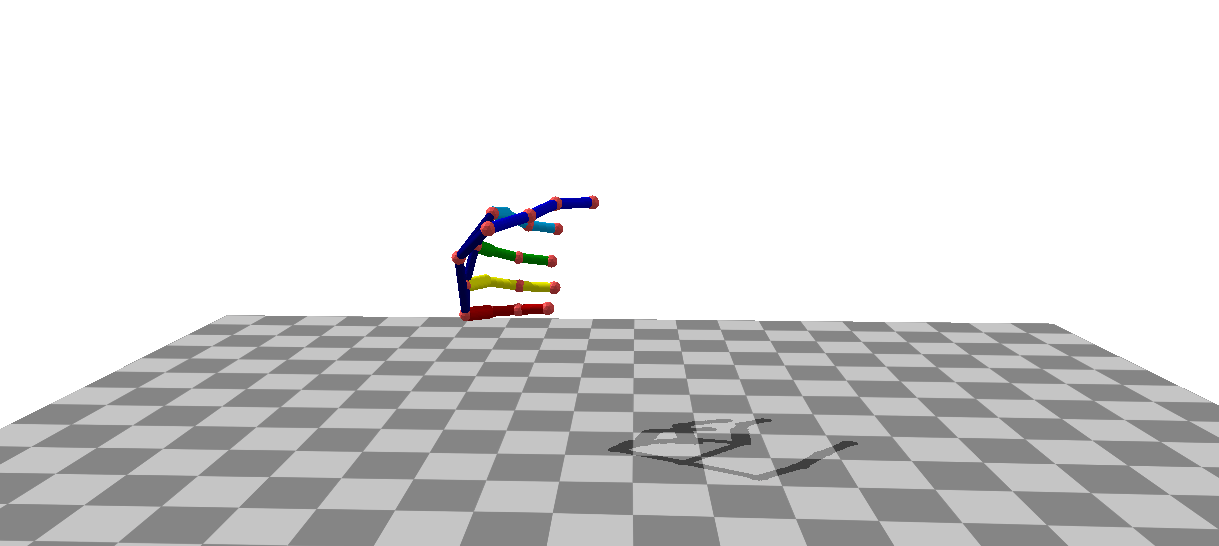}~
 \includegraphics[width=0.16\linewidth]{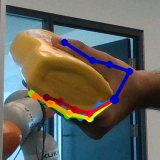}~
 \includegraphics[width=0.16\linewidth,trim=12cm 0cm 17cm 5cm ,clip]{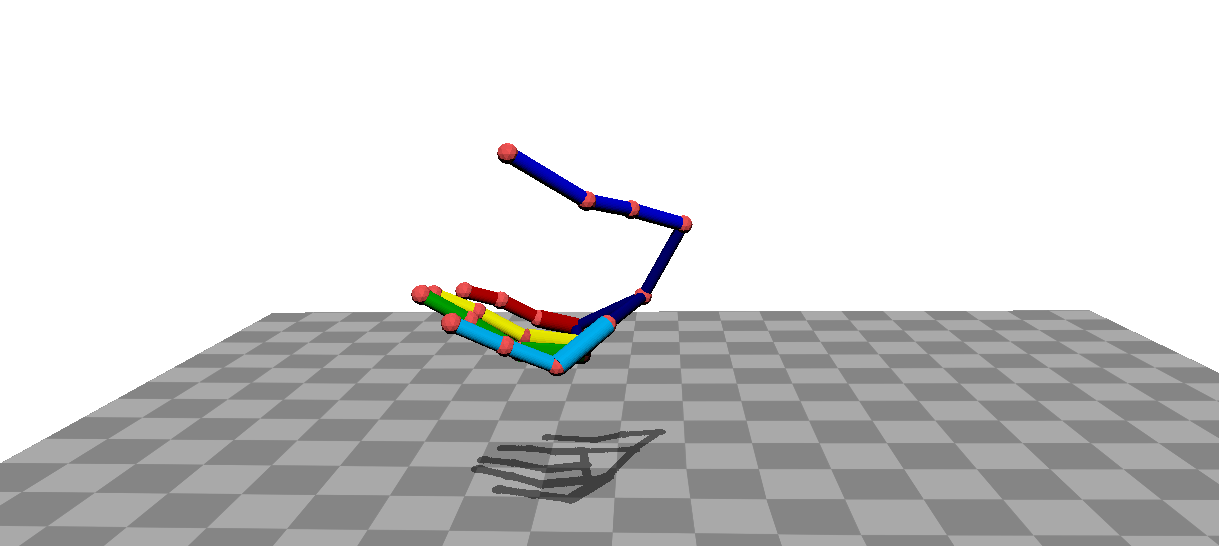}~
 \includegraphics[width=0.16\linewidth,trim=10cm 5cm 23cm 5cm ,clip]{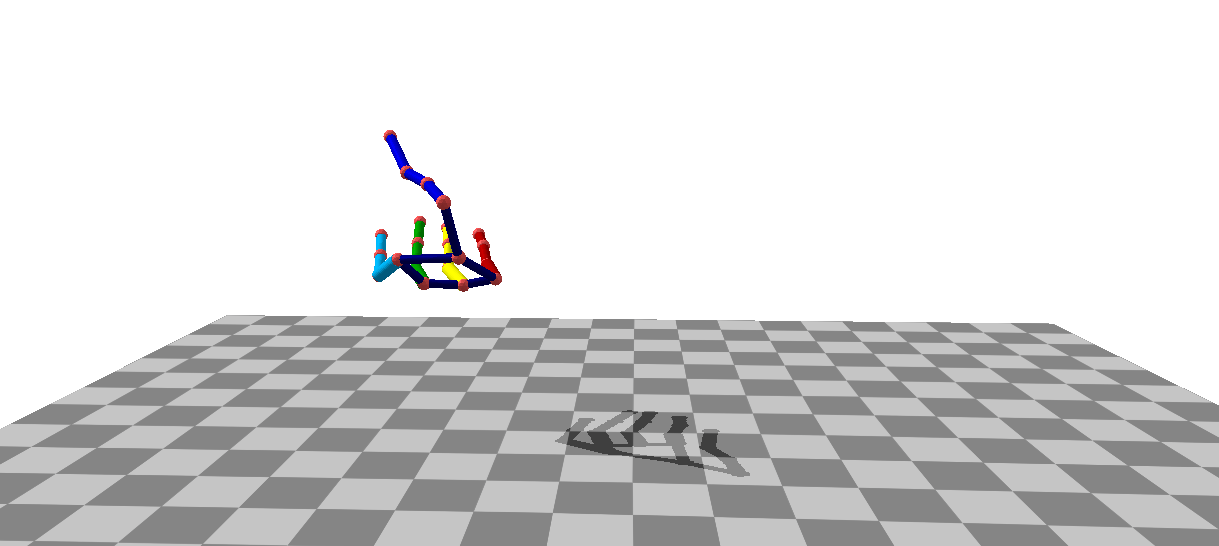}
 \\ \vspace{-3mm}
 \includegraphics[width=0.16\linewidth]{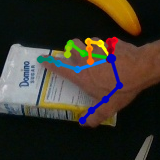}~
 \includegraphics[width=0.16\linewidth,trim=12cm 0cm 17cm 5cm ,clip]{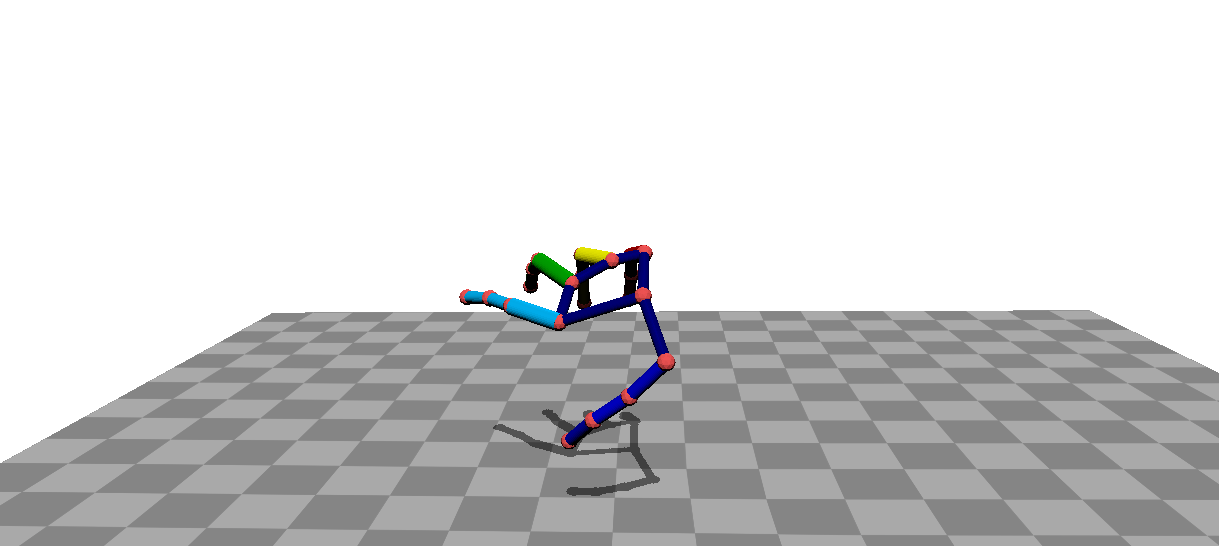}~
 \includegraphics[width=0.16\linewidth,trim=12cm 5cm 19cm 0cm ,clip]{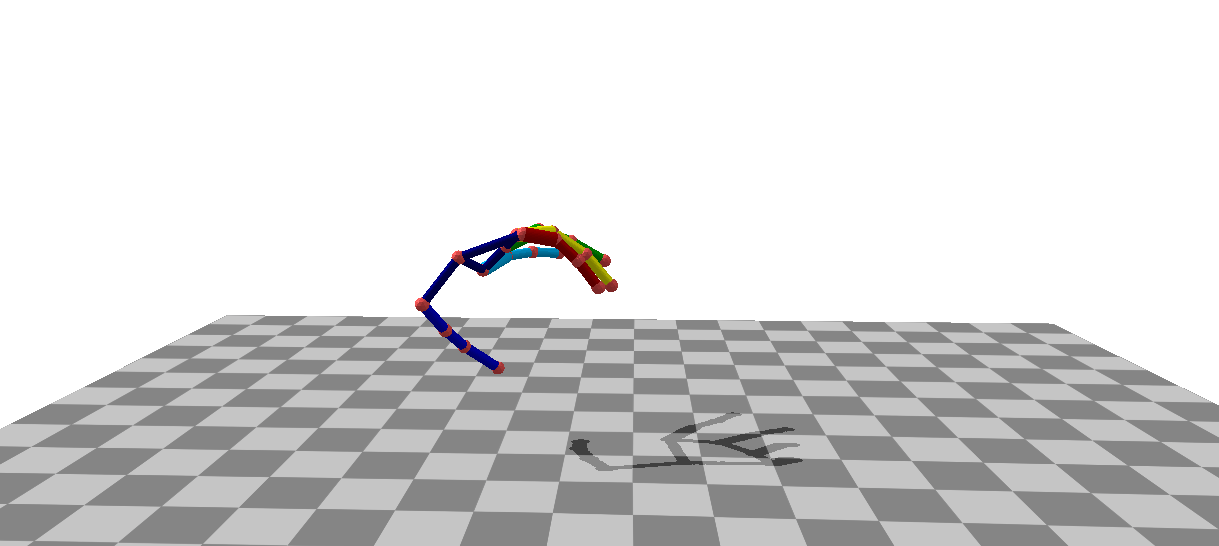}~
 \includegraphics[width=0.16\linewidth]{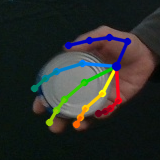}~
 \includegraphics[width=0.16\linewidth,trim=12cm 0cm 17cm 7cm ,clip]{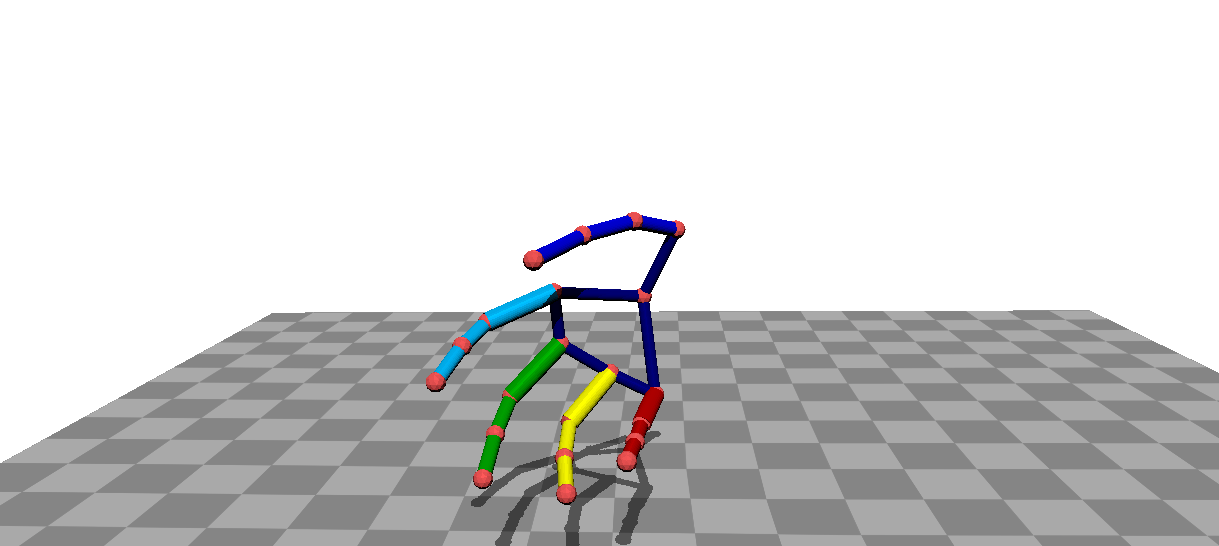}~
 \includegraphics[width=0.16\linewidth,trim=10cm 5cm 23cm 5cm ,clip]{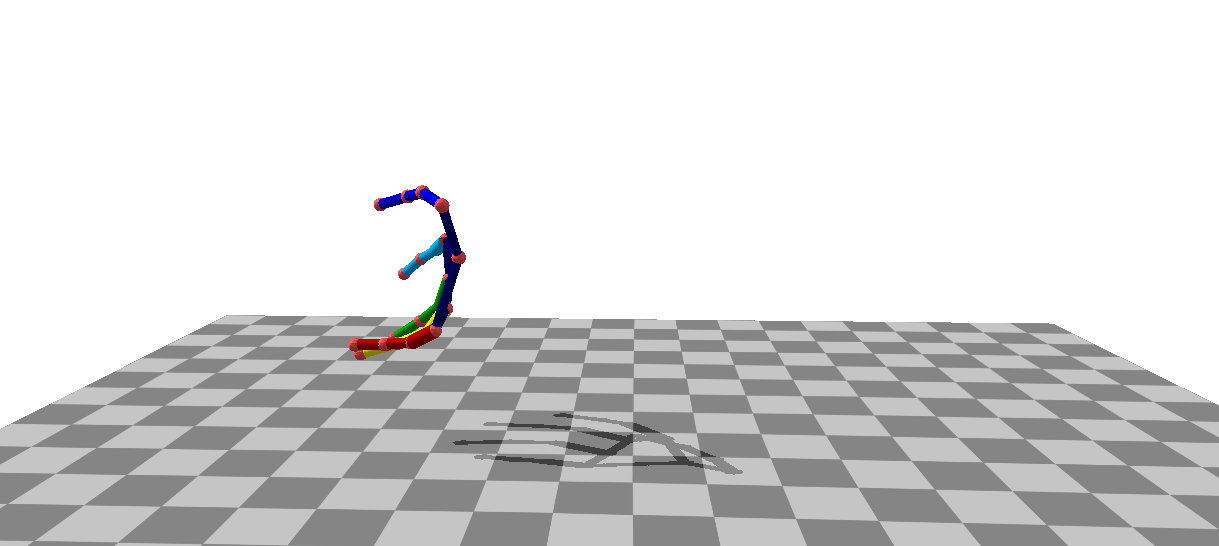}
 \\ \vspace{1mm}
 \includegraphics[width=0.16\linewidth]{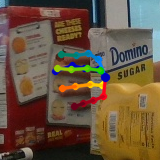}~
 \includegraphics[width=0.16\linewidth,trim=12cm 0cm 17cm 5cm ,clip]{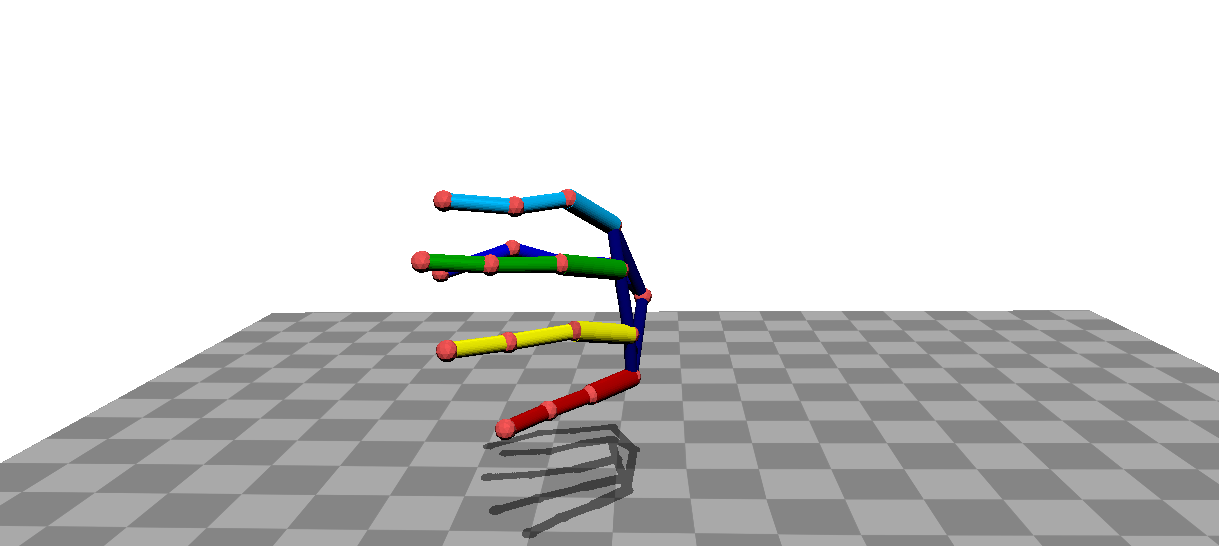}~
 \includegraphics[width=0.16\linewidth,trim=10cm 5cm 23cm 5cm ,clip]{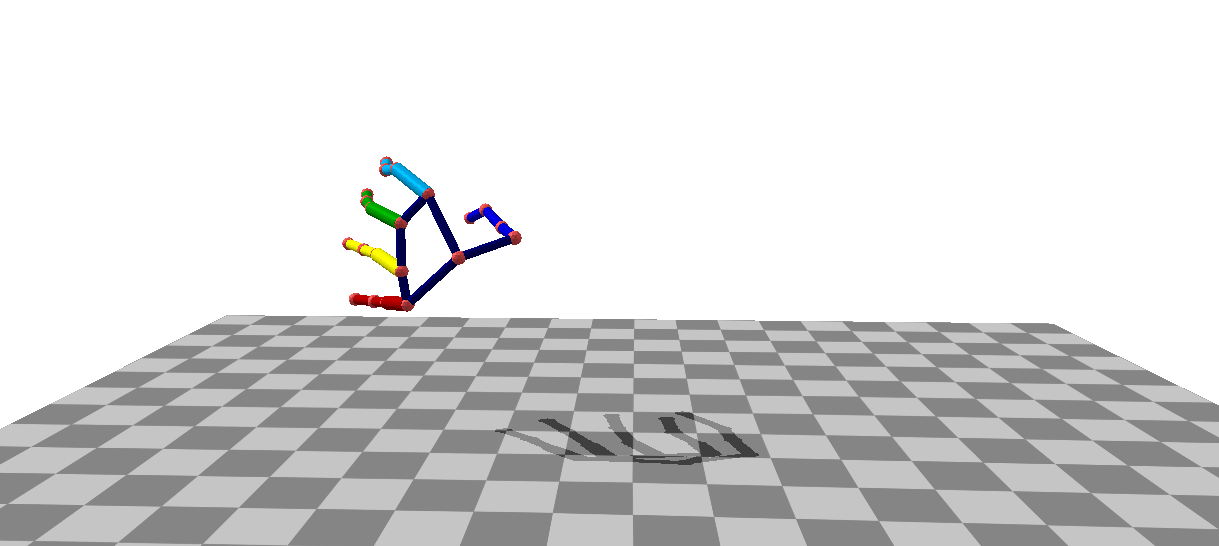}~
 \includegraphics[width=0.16\linewidth]{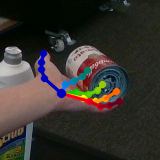}~
 \includegraphics[width=0.16\linewidth,trim=17cm 3cm 12cm 2cm ,clip]{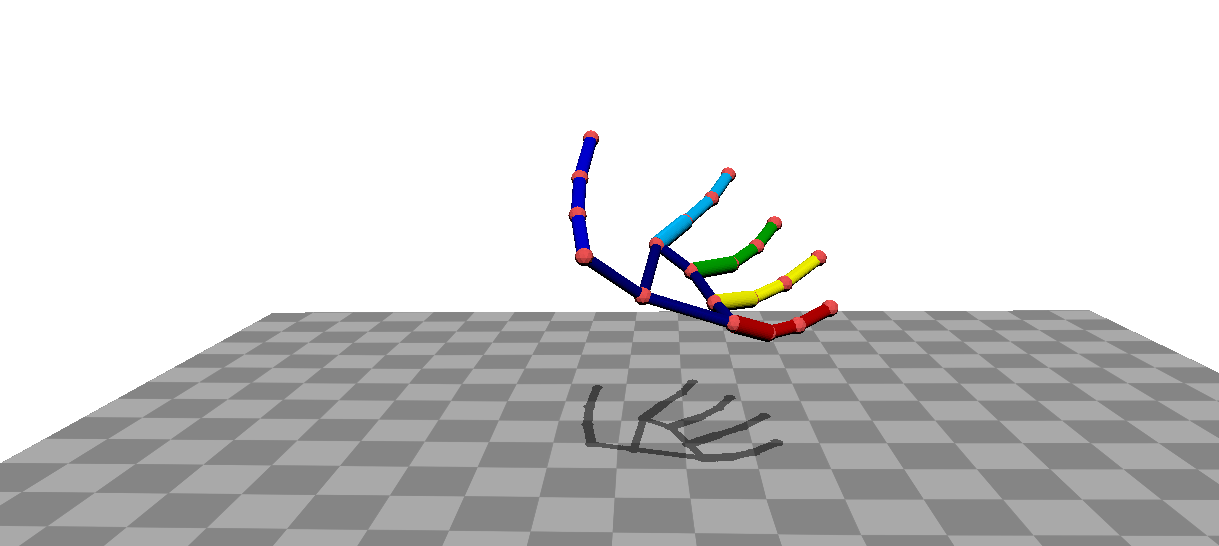}~
 \includegraphics[width=0.16\linewidth,trim=12cm 5cm 19cm 2cm ,clip]{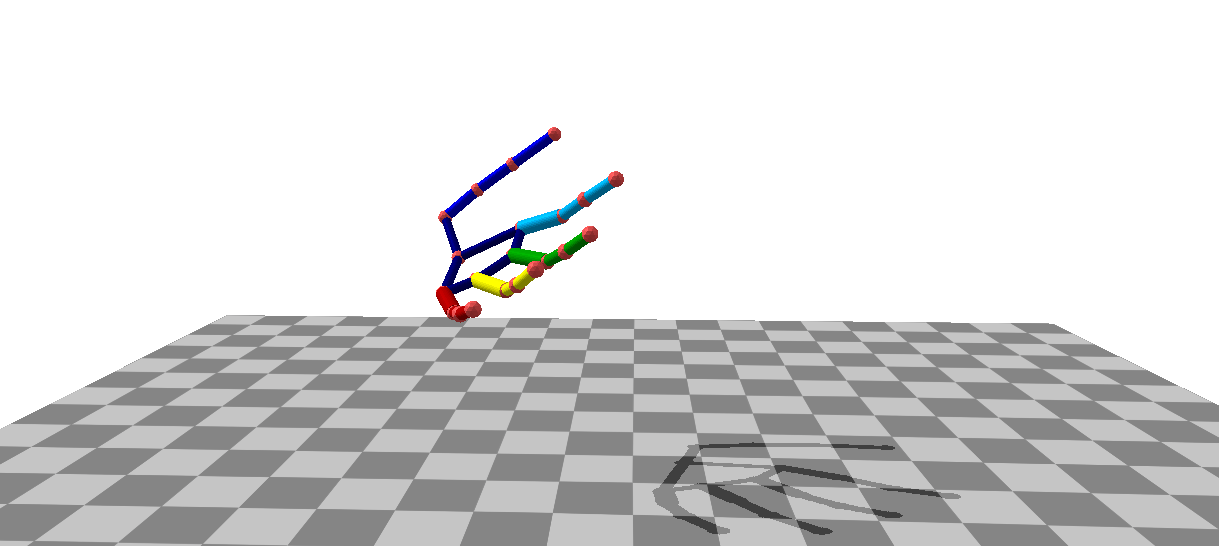}
 \caption{\small Qualitative results of the predicted 3D hand pose using Spurr 
et al.'s method~\cite{spurr:eccv2020} (HRNet32). For each image we visualize 3D
pose from both front and side views.}
 \label{fig:hand}
\end{figure*}

Fig.~\ref{fig:hand} shows qualitative results of 3D hand pose estimation using
Spurr et al.'s method~\cite{spurr:eccv2020} (HRNet32) on the S0 setup. The
method is able to generate sensible articulated pose even under object
occlusions. This is consistent with the quantitative results reported in
Tab.~\ref{tab:3d}, where the mean per joint position error
after Procrustes alignment is less than 1cm (6.83mm). As suggested in that
table, the major source of error comes from translation, rotate, and scale
(Absolute), rather than articulation. Nonetheless, we still see errors in local
articulation when objects are in close contact (e.g., the middle finger in row
2 and column 1) and when the fingers are largely occluded by the held object
(e.g., the index finger in row 6 and column 4).

\subsection{Full Quantitative: 6D Object Pose Estimation}

For 6D object pose estimation, since in Tab.~\ref{tab:6d:sota} we
report benchmark results only on S1, we now include the results on the other
three setups. Tab.~\ref{tab:6d:sota:s0},~\ref{tab:6d:sota:s2}, and
~\ref{tab:6d:sota:s3} show the results in AR on S0, S2, and S3, respectively.
Overall, we observe a similar trend as on S1 (see Sec.~\ref{sec:6d}
), where DeepIM (RGB-D)~\cite{li:eccv2018} and
CosyPose~\cite{labbe:eccv2020} are the two most competitive approaches in
estimation accuracy.

\begin{table*}[t]
 \centering
 \small
 \begin{tabular}{l|ccccccc}
                           & \multicolumn{7}{c}{S2 (unseen views)} \\ \cline{2-8}
                           & \multicolumn{2}{c}{PoseCNN~\cite{xiang:rss2018}} & \multicolumn{2}{c}{DeepIM~\cite{li:eccv2018}} & \multicolumn{2}{c}{PoseRBPF~\cite{deng:rss2019}} & CosyPose~\cite{labbe:eccv2020} \\
                           & RGB & + depth ref & RGB & RGB-D & RGB & RGB-D & RGB \\
  \hline
  002\_master\_chef\_can   & 51.36 & 57.42 & 68.86 & 72.87          & 37.64 & 63.04          & \textbf{81.94} \\
  003\_cracker\_box        & 60.65 & 67.45 & 84.10 & 89.34          & 54.25 & 74.63          & \textbf{90.66} \\
  004\_sugar\_box          & 51.73 & 60.06 & 69.68 & 76.23          & 47.15 & 71.17          & \textbf{83.11} \\
  005\_tomato\_soup\_can   & 45.44 & 52.56 & 55.36 & \textbf{65.60} & 32.62 & 47.60          & 61.52          \\
  006\_mustard\_bottle     & 51.96 & 58.77 & 64.04 & 71.64          & 40.81 & 65.02          & \textbf{73.55} \\
  007\_tuna\_fish\_can     & 31.96 & 39.20 & 44.21 & 51.25          & 26.09 & 44.43          & \textbf{55.65} \\
  008\_pudding\_box        & 50.81 & 61.41 & 64.94 & 73.88          & 36.66 & 62.11          & \textbf{77.75} \\
  009\_gelatin\_box        & 44.38 & 52.93 & 59.64 & 70.02          & 37.51 & 47.61          & \textbf{72.51} \\
  010\_potted\_meat\_can   & 45.08 & 54.94 & 66.58 & \textbf{70.41} & 40.13 & 51.26          & 68.60          \\
  011\_banana              & 44.42 & 49.09 & 48.16 & \textbf{58.30} & 24.85 & 49.07          & 42.50          \\
  019\_pitcher\_base       & 53.51 & 56.73 & 61.36 & \textbf{71.98} & 26.37 & 62.30          & 56.77          \\
  021\_bleach\_cleanser    & 56.03 & 62.24 & 66.20 & \textbf{76.89} & 38.13 & 67.77          & 73.86          \\
  024\_bowl                & 58.15 & 62.15 & 67.75 & 72.60          & 40.76 & 70.66          & \textbf{80.42} \\
  025\_mug                 & 43.70 & 46.79 & 52.47 & 58.65          & 26.81 & 48.77          & \textbf{59.09} \\
  035\_power\_drill        & 50.79 & 57.14 & 68.25 & \textbf{75.54} & 40.00 & 60.70          & 54.66          \\
  036\_wood\_block         & 49.50 & 57.34 & 57.19 & \textbf{75.54} & 33.19 & 60.41          & 72.40          \\
  037\_scissors            & 25.90 & 28.21 & 31.42 & 36.68          & 27.86 & \textbf{39.38} & 26.90          \\
  040\_large\_marker       & 38.19 & 48.47 & 38.21 & 56.44          & 27.65 & 35.72          & \textbf{58.60} \\
  052\_extra\_large\_clamp & 34.42 & 40.21 & 42.24 & 52.72          & 28.25 & 46.33          & \textbf{58.73} \\
  061\_foam\_brick         & 15.75 & 19.11 & 20.10 & 28.39          & 26.19 & \textbf{40.90} & 31.58          \\
  \hline
  all                      & 45.18 & 51.61 & 56.54 & \textbf{65.25} & 34.65 & 55.44          & 64.04 
 \end{tabular}
 \vspace{-2mm}
 \caption{\small 6D object pose estimation results of representative approaches
in AR (\%) on S2.}
 \label{tab:6d:sota:s2}
\end{table*}

\begin{table*}[t]
 \centering
 \small
 \begin{tabular}{l|ccccccc}
                           & \multicolumn{7}{c}{S3 (unseen grasping)} \\ \cline{2-8}
                           & \multicolumn{2}{c}{PoseCNN~\cite{xiang:rss2018}} & \multicolumn{2}{c}{DeepIM~\cite{li:eccv2018}} & \multicolumn{2}{c}{PoseRBPF~\cite{deng:rss2019}} & CosyPose~\cite{labbe:eccv2020} \\
                           & RGB & + depth ref & RGB & RGB-D & RGB & RGB-D & RGB \\
  \hline
  002\_master\_chef\_can   & --    & --    & --    & --             & --    & --    & --             \\
  003\_cracker\_box        & --    & --    & --    & --             & --    & --    & --             \\ 
  004\_sugar\_box          & --    & --    & --    & --             & --    & --    & --             \\ 
  005\_tomato\_soup\_can   & --    & --    & --    & --             & --    & --    & --             \\ 
  006\_mustard\_bottle     & --    & --    & --    & --             & --    & --    & --             \\ 
  007\_tuna\_fish\_can     & --    & --    & --    & --             & --    & --    & --             \\ 
  008\_pudding\_box        & --    & --    & --    & --             & --    & --    & --             \\ 
  009\_gelatin\_box        & 33.07 & 43.43 & 47.68 & 54.13          & 29.95 & 47.71 & \textbf{58.29} \\ 
  010\_potted\_meat\_can   & --    & --    & --    & --             & --    & --    & --             \\ 
  011\_banana              & --    & --    & --    & --             & --    & --    & --             \\ 
  019\_pitcher\_base       & --    & --    & --    & --             & --    & --    & --             \\ 
  021\_bleach\_cleanser    & 38.52 & 45.32 & 48.32 & 60.18          & 25.35 & 58.31 & \textbf{63.04} \\ 
  024\_bowl                & --    & --    & --    & --             & --    & --    & --             \\ 
  025\_mug                 & --    & --    & --    & --             & --    & --    & --             \\ 
  035\_power\_drill        & --    & --    & --    & --             & --    & --    & --             \\ 
  036\_wood\_block         & 40.53 & 48.17 & 50.54 & \textbf{65.74} & 27.30 & 55.19 & 65.31          \\ 
  037\_scissors            & --    & --    & --    & --             & --    & --    & --             \\ 
  040\_large\_marker       & --    & --    & --    & --             & --    & --    & --             \\ 
  052\_extra\_large\_clamp & --    & --    & --    & --             & --    & --    & --             \\ 
  061\_foam\_brick         & --    & --    & --    & --             & --    & --    & --             \\ 
  \hline
  all                      & 37.41 & 45.66 & 48.86 & 60.07          & 27.52 & 53.78 & \textbf{62.25} \\ 
 \end{tabular}
 \vspace{-2mm}
 \caption{\small 6D object pose estimation results of representative approaches
in AR (\%) on S3.}
 \label{tab:6d:sota:s3}
\end{table*}

\begin{figure*}[t]
 \centering
 \captionsetup[subfigure]{labelformat=empty}
 \begin{subfigure}[c]{0.32\linewidth}
  \centering
  \includegraphics[width=1.00\linewidth]{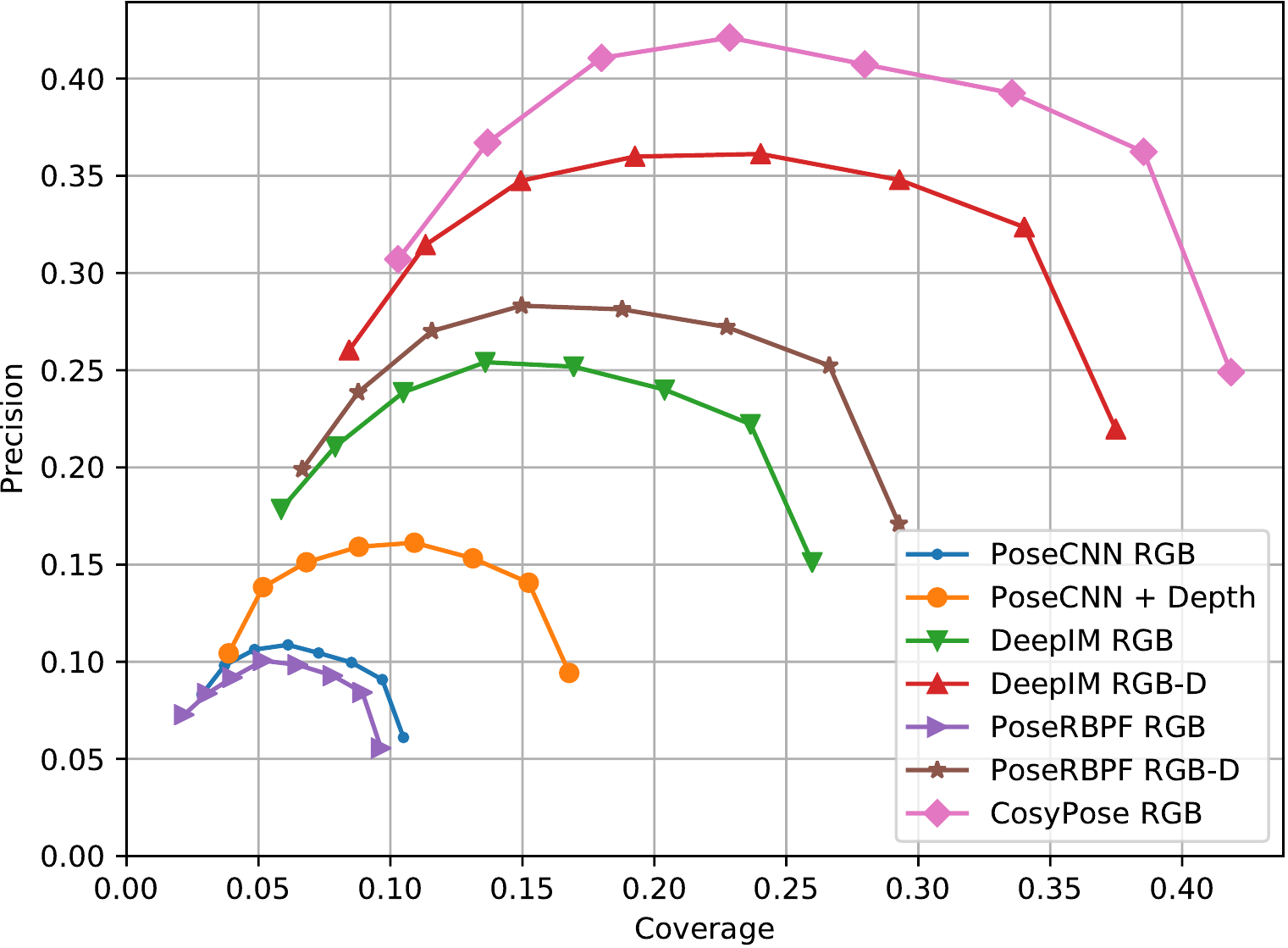}
  \caption{S0 (default)}
 \end{subfigure}
 ~~
 \begin{subfigure}[c]{0.32\linewidth}
  \centering
  \includegraphics[width=1.00\linewidth]{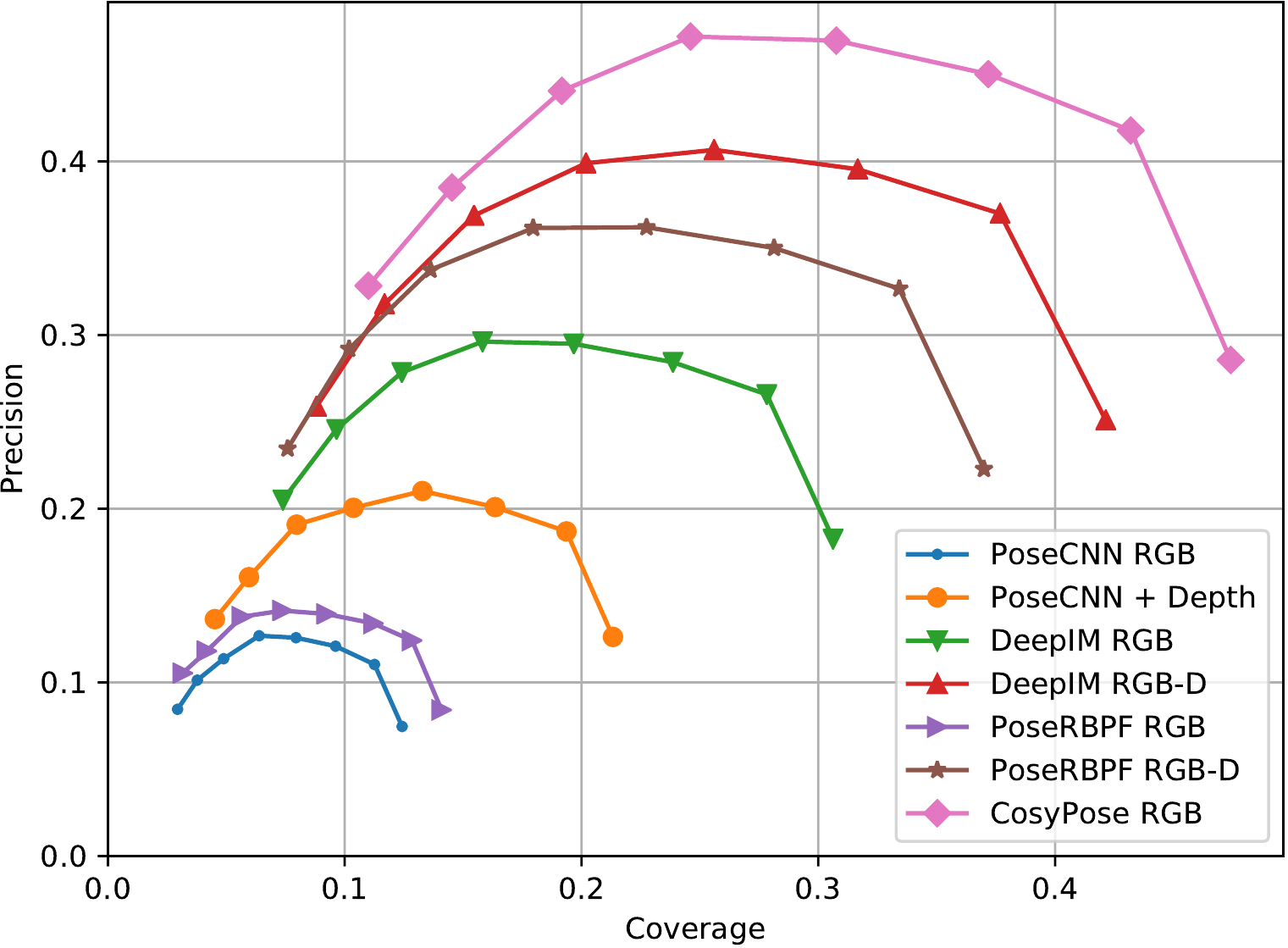}
  \caption{S2 (unseen views)}
 \end{subfigure}
 ~~
 \begin{subfigure}[c]{0.32\linewidth}
  \centering
  \includegraphics[width=1.00\linewidth]{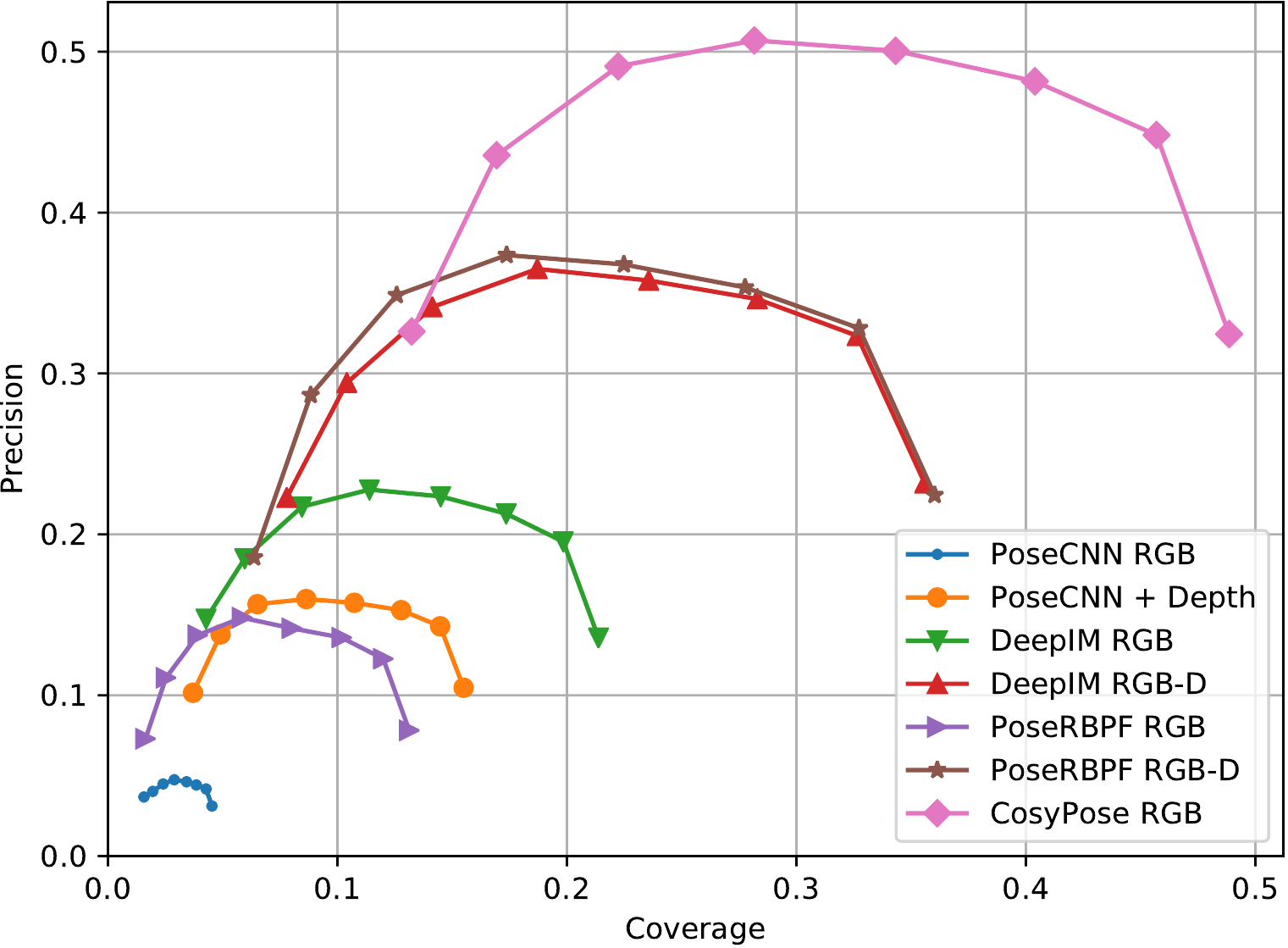}
  \caption{S3 (unseen grasping)}
 \end{subfigure}
 \caption{\small Precision-coverage curves for grasp generation on S0, S2, and
S3.}
 \label{fig:curve:supp}
\end{figure*}

\section{Safe Human-to-Robot Object Handover}

\subsection{Pre-Generated Grasps}

\begin{figure*}[t]
 \centering
 \includegraphics[width=0.24\linewidth]{100grasps-003_cracker_box.png}~
 \includegraphics[width=0.24\linewidth]{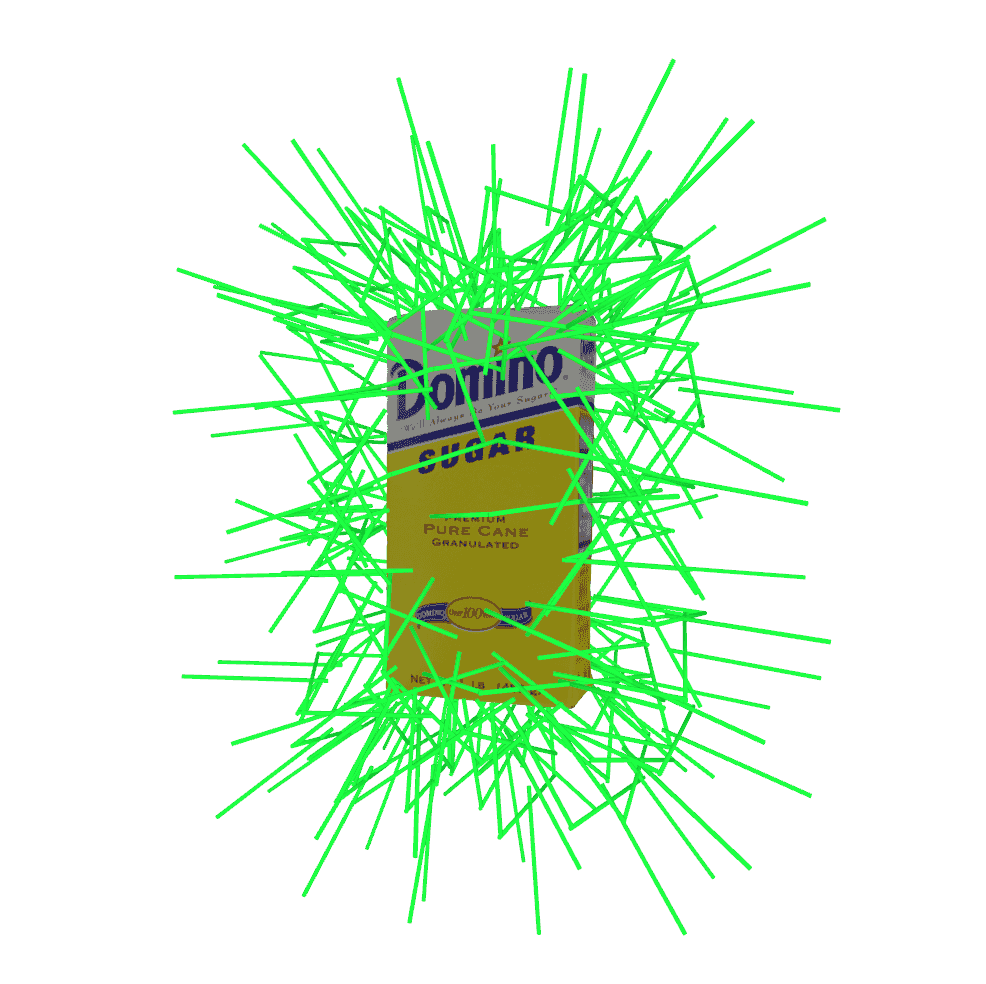}~
 \includegraphics[width=0.24\linewidth]{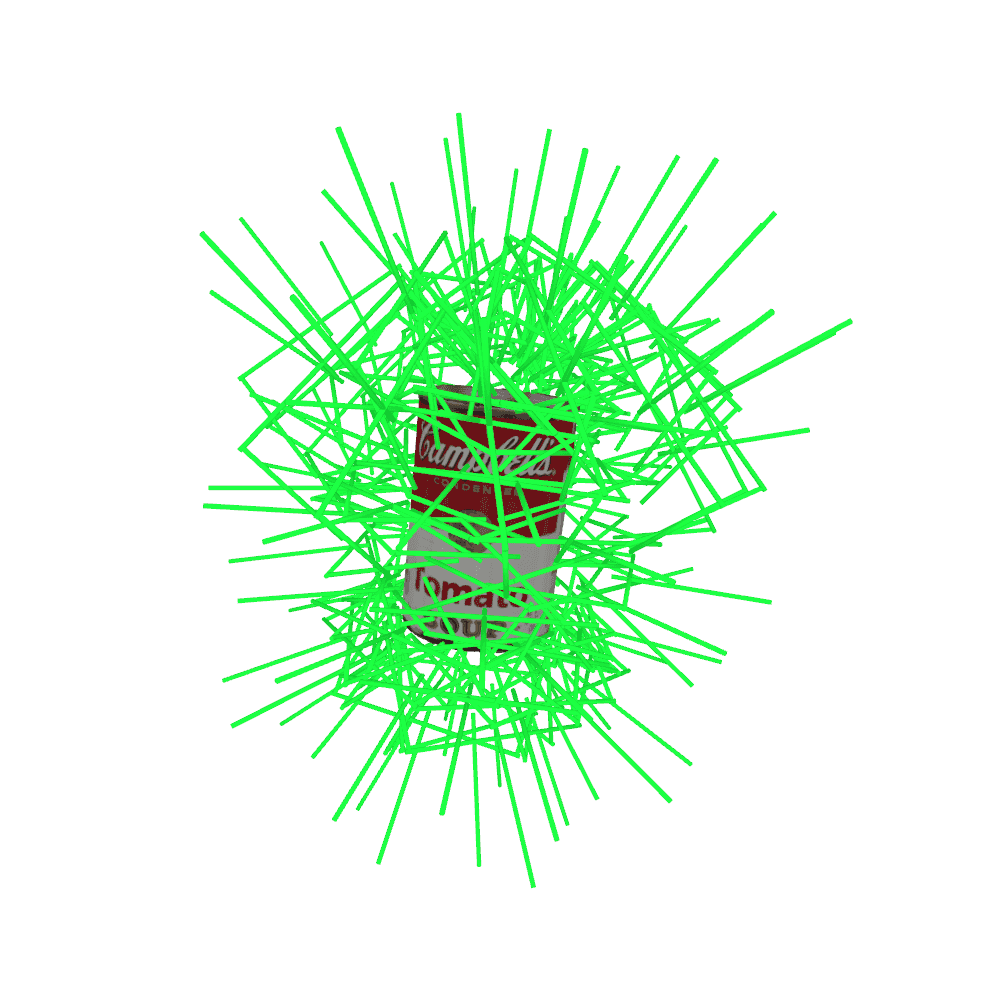}~
 \includegraphics[width=0.24\linewidth]{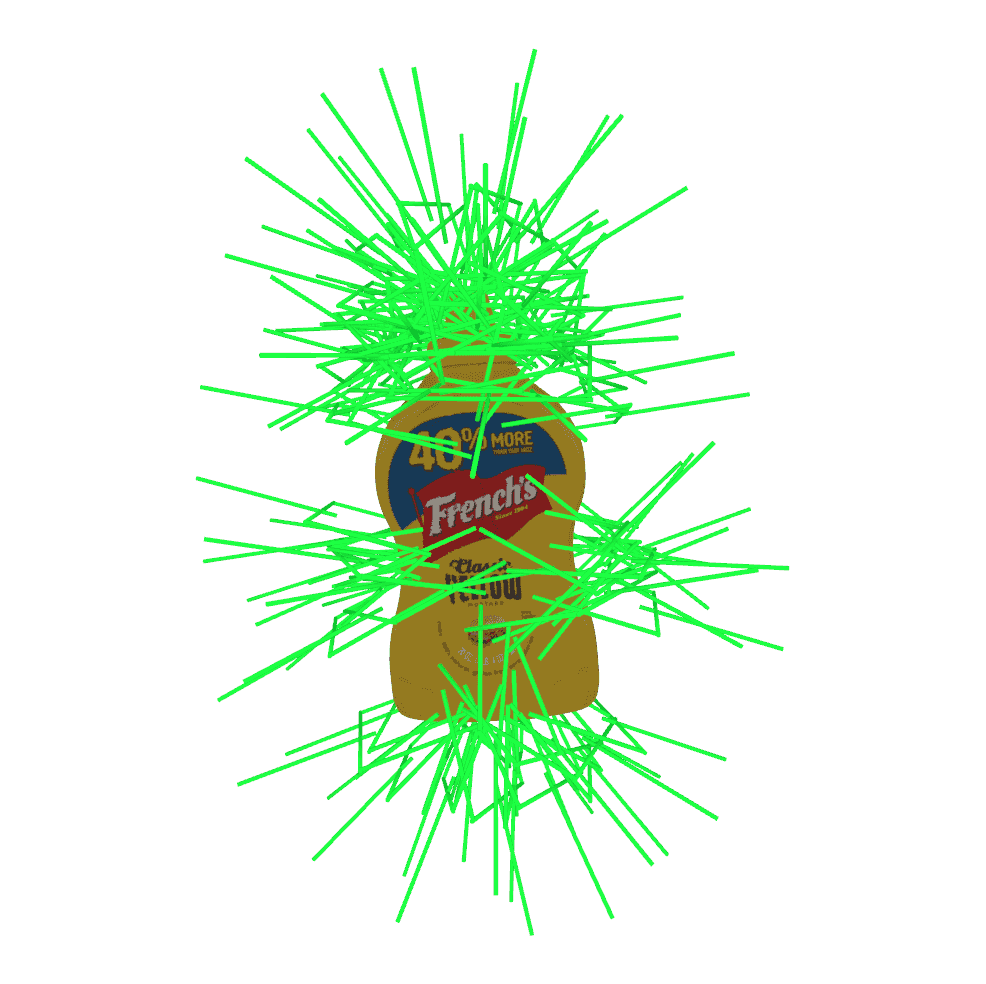}
 \\ \vspace{1mm}
 \includegraphics[width=0.24\linewidth]{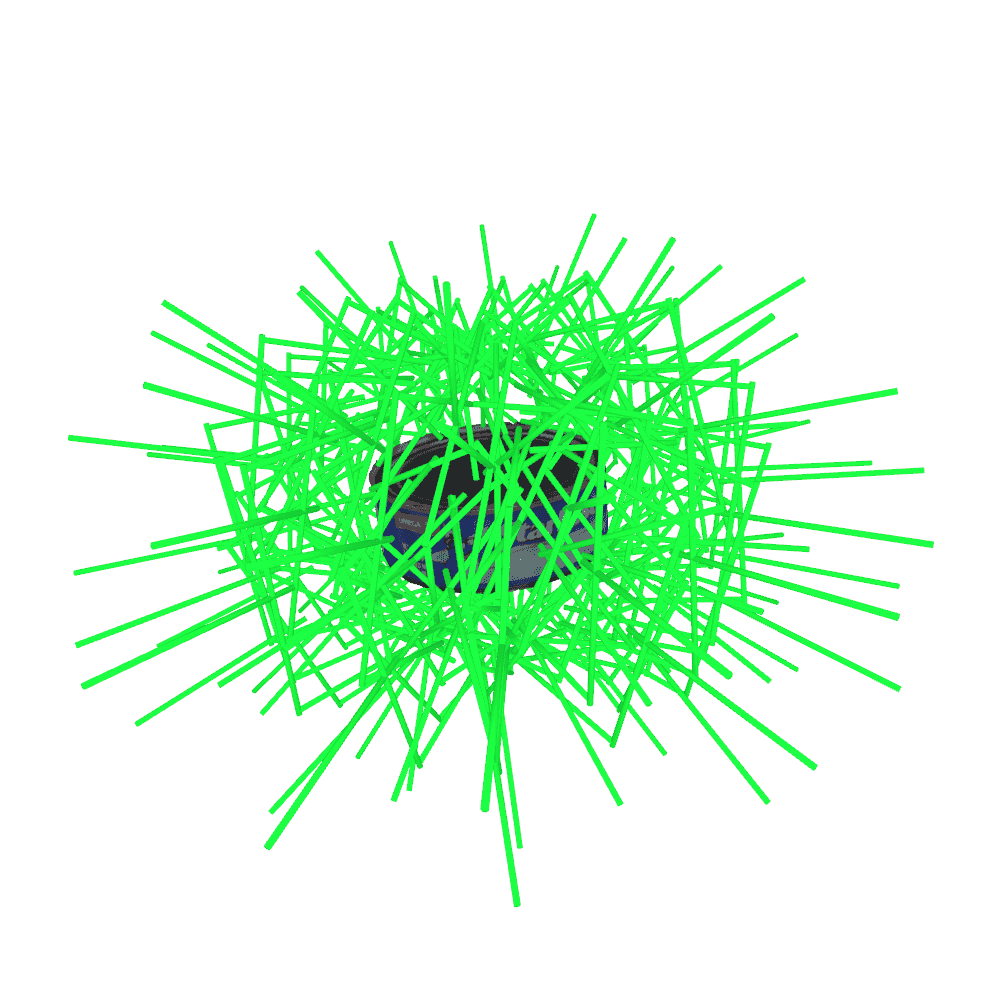}~
 \includegraphics[width=0.24\linewidth]{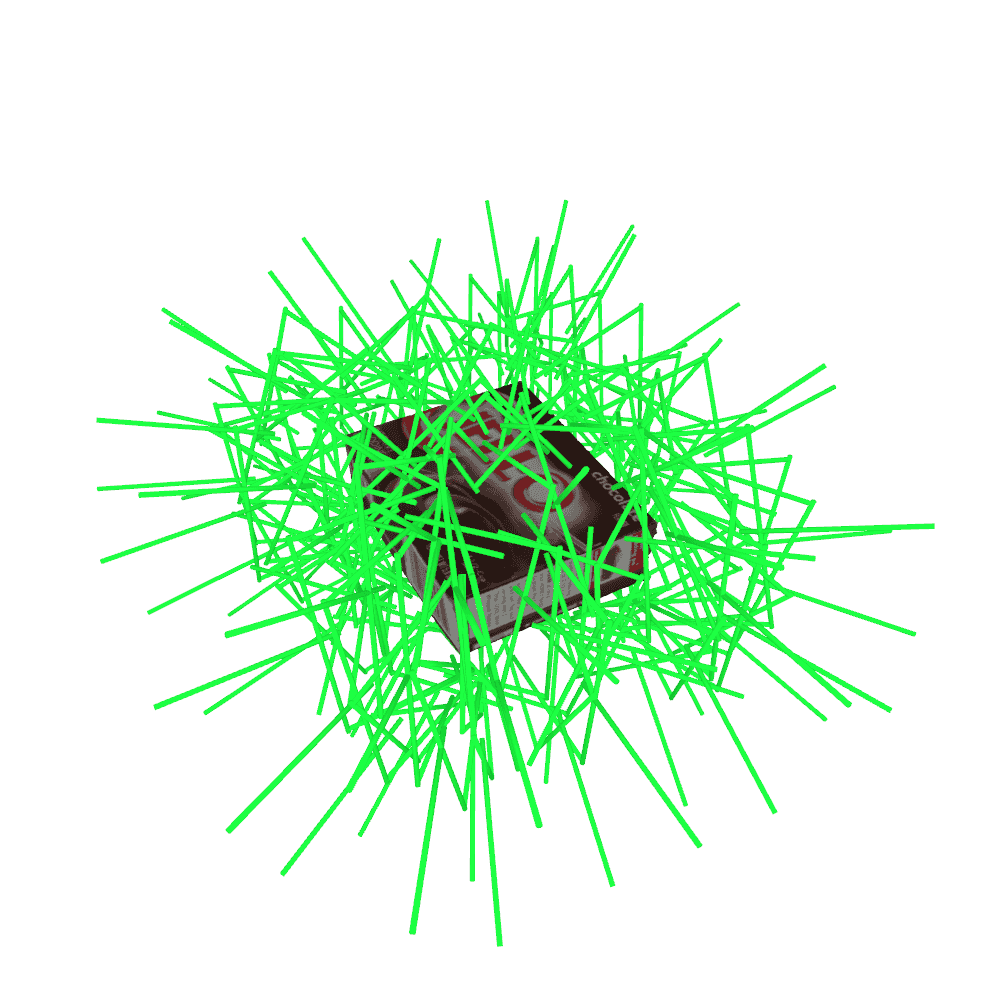}~
 \includegraphics[width=0.24\linewidth]{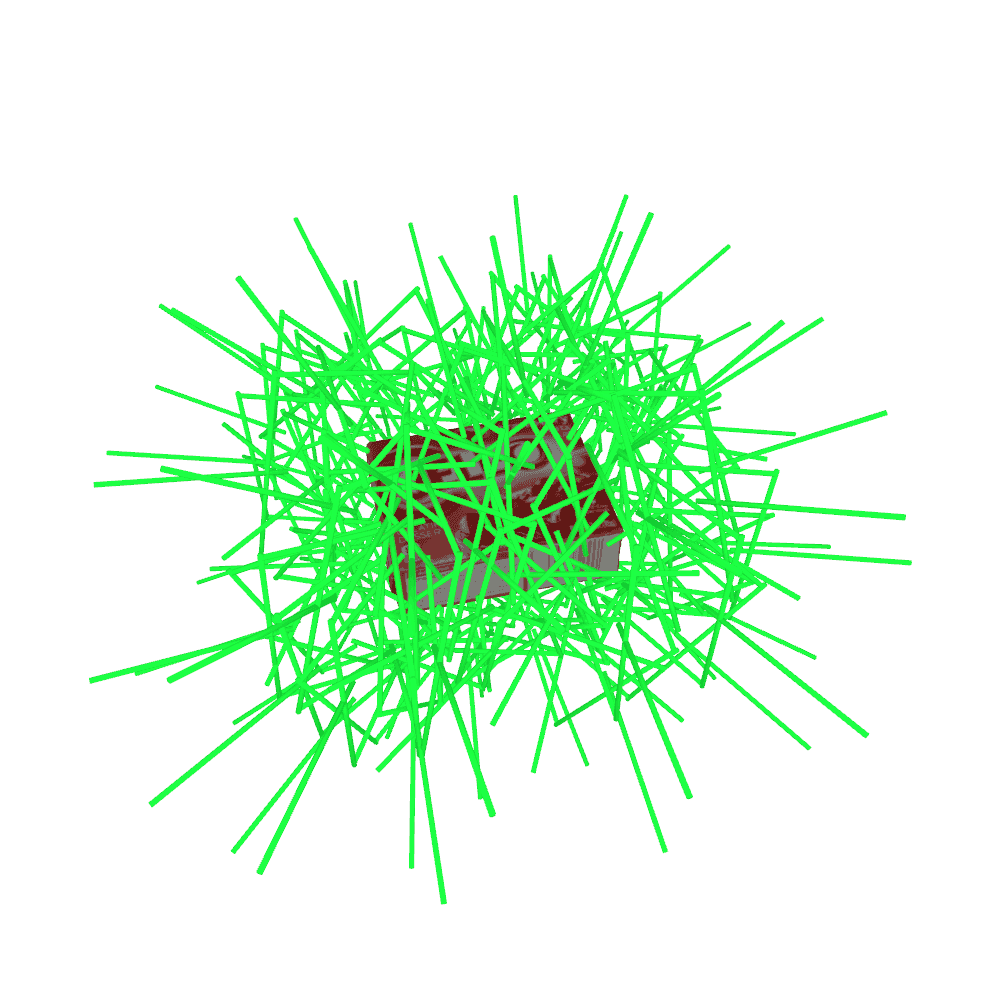}~
 \includegraphics[width=0.24\linewidth]{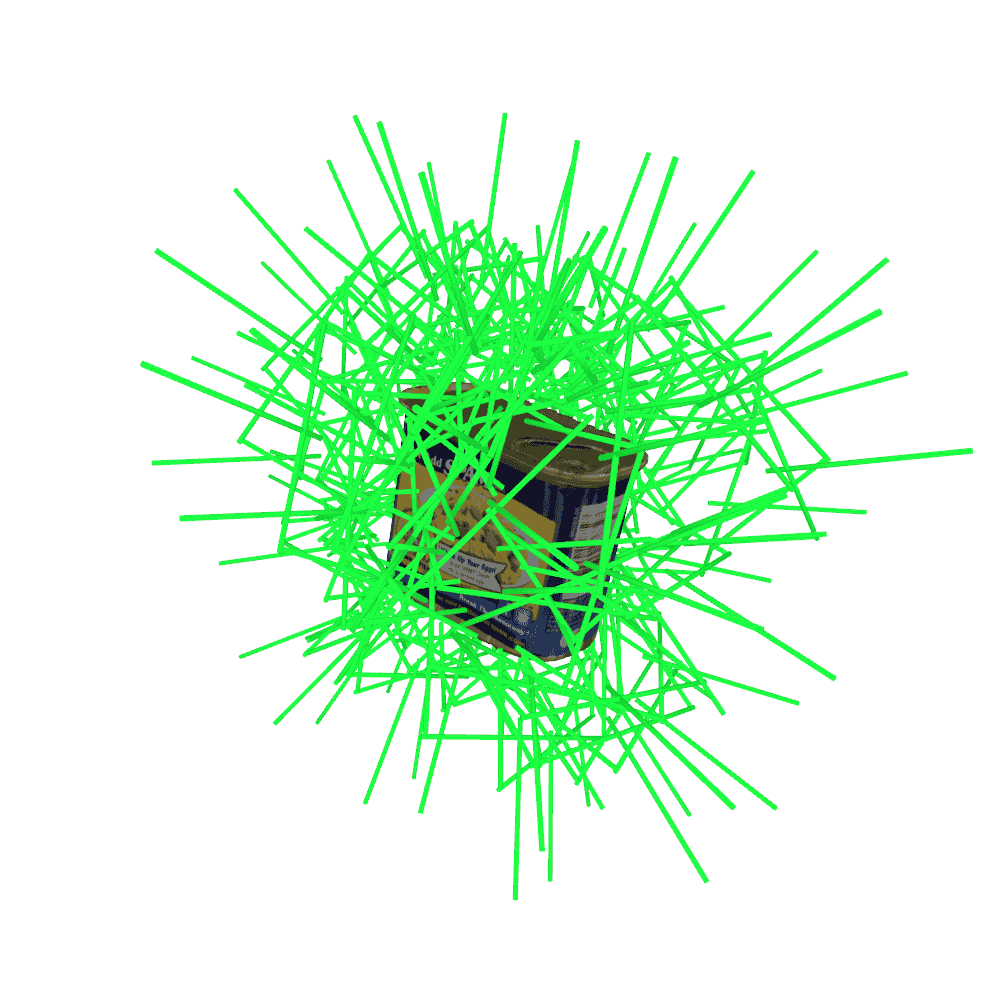}
 \\ \vspace{1mm}
 \includegraphics[width=0.24\linewidth]{100grasps-011_banana.png}~
 \includegraphics[width=0.24\linewidth]{100grasps-019_pitcher_base.png}~
 \includegraphics[width=0.24\linewidth]{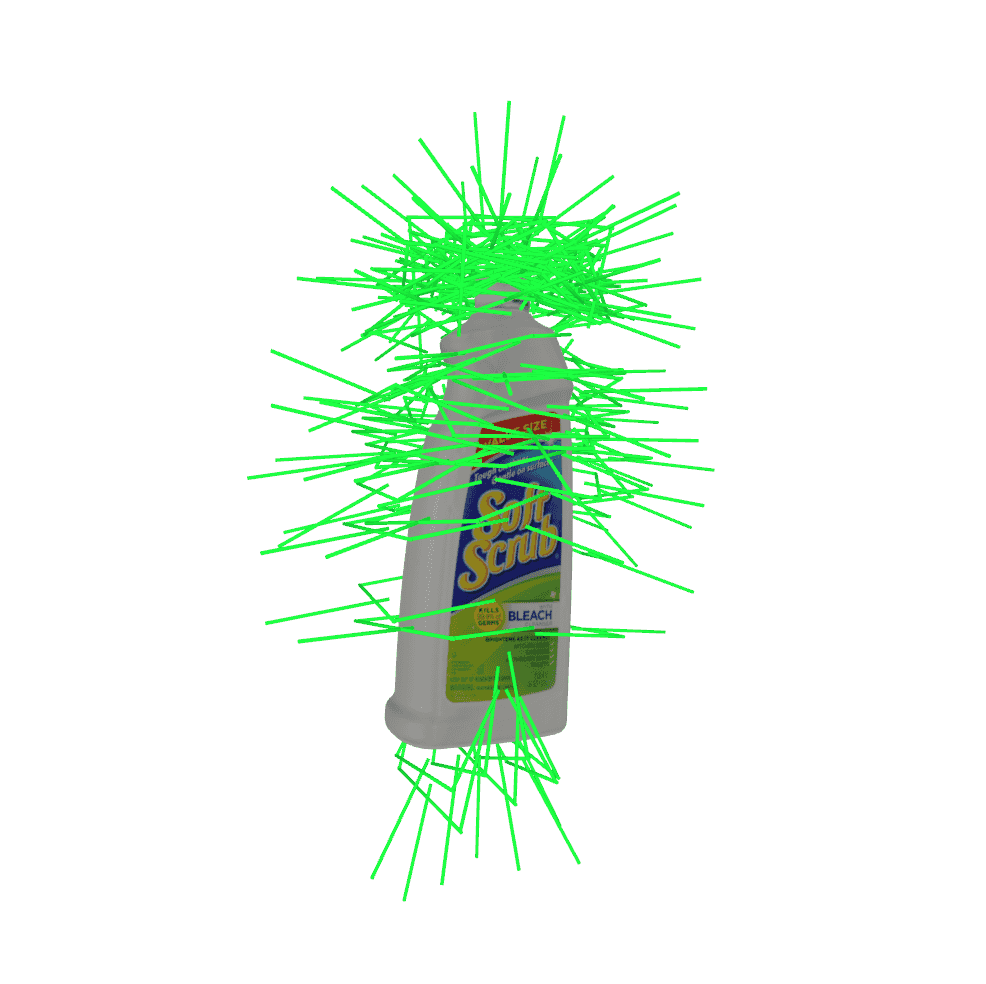}~
 \includegraphics[width=0.24\linewidth]{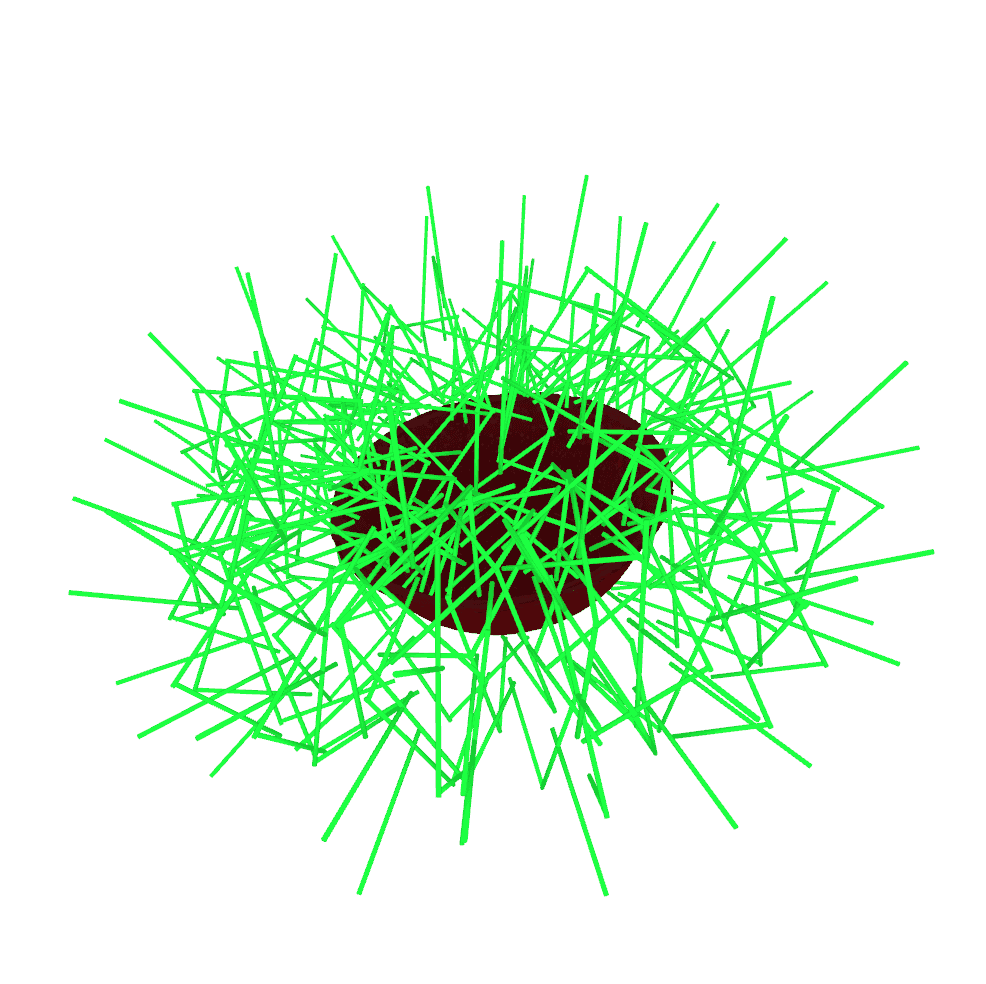}
 \\ \vspace{1mm}
 \includegraphics[width=0.24\linewidth]{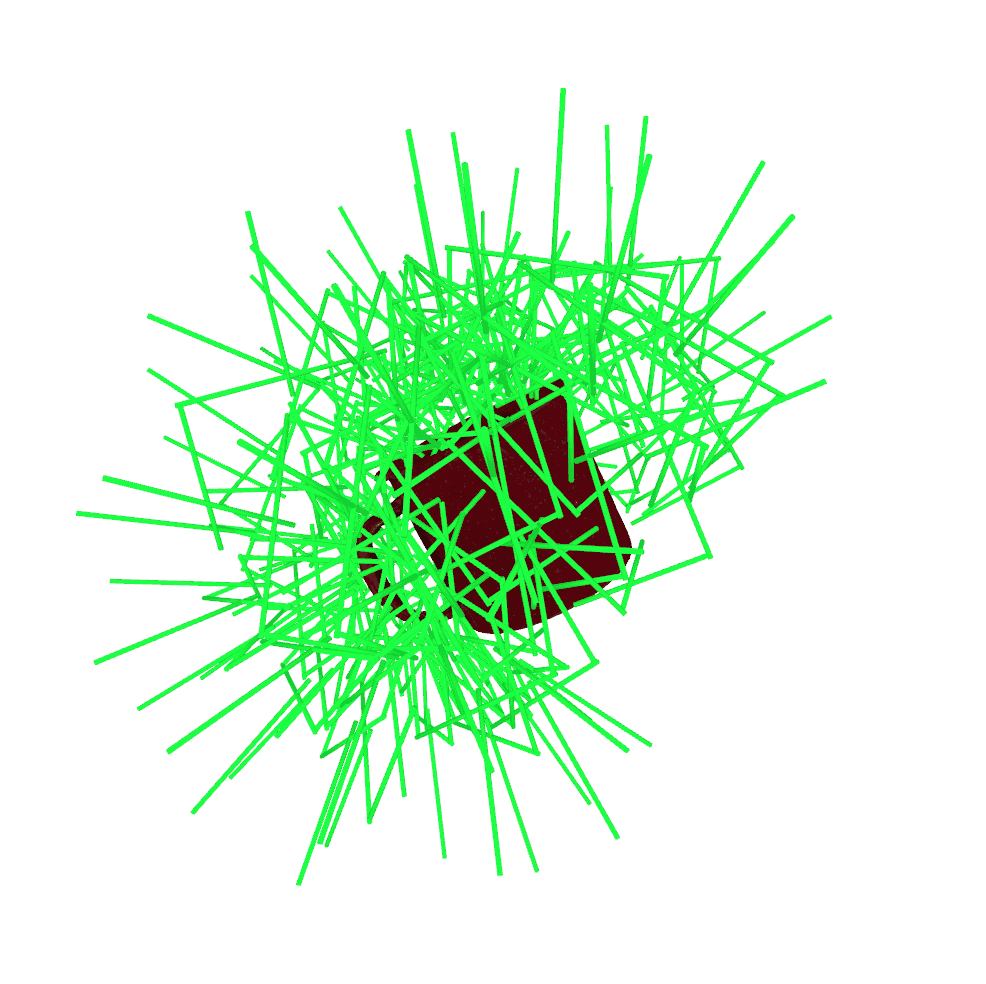}~
 \includegraphics[width=0.24\linewidth]{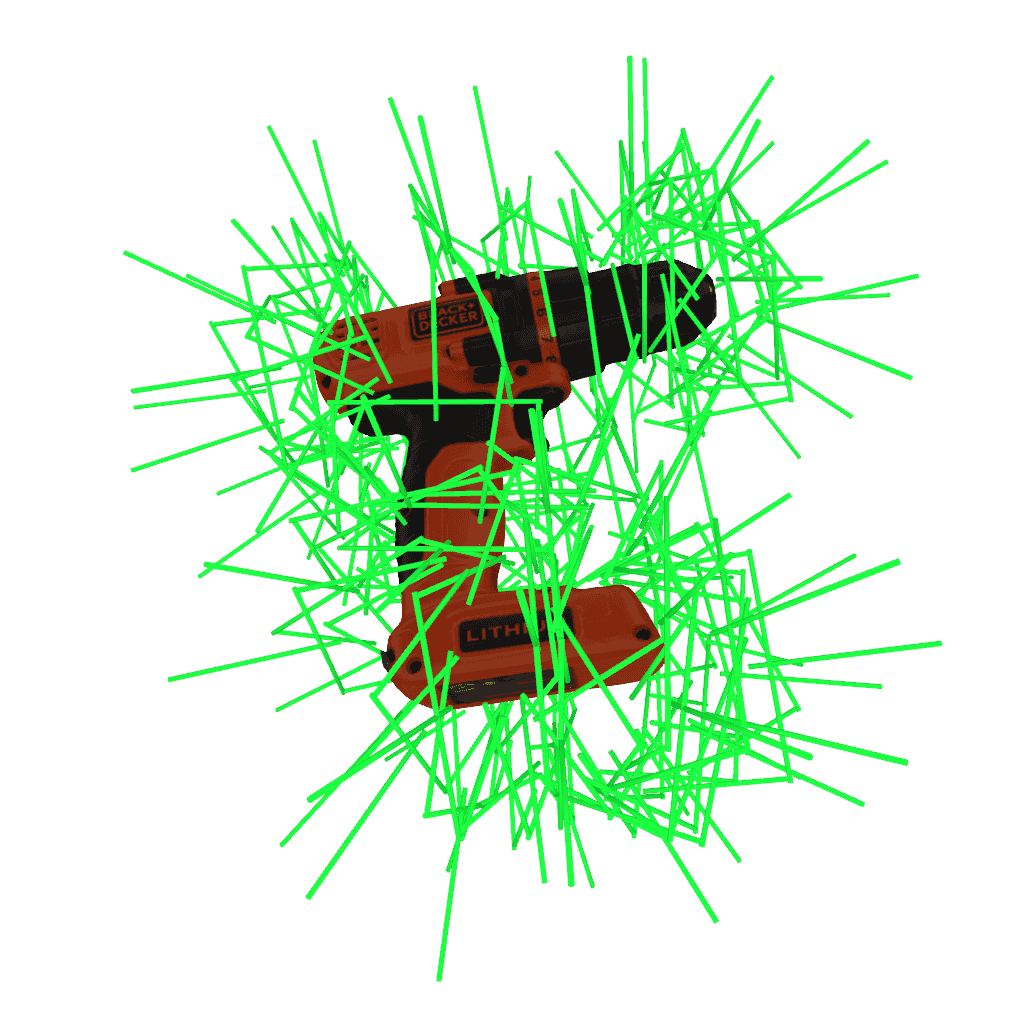}~
 \includegraphics[width=0.24\linewidth]{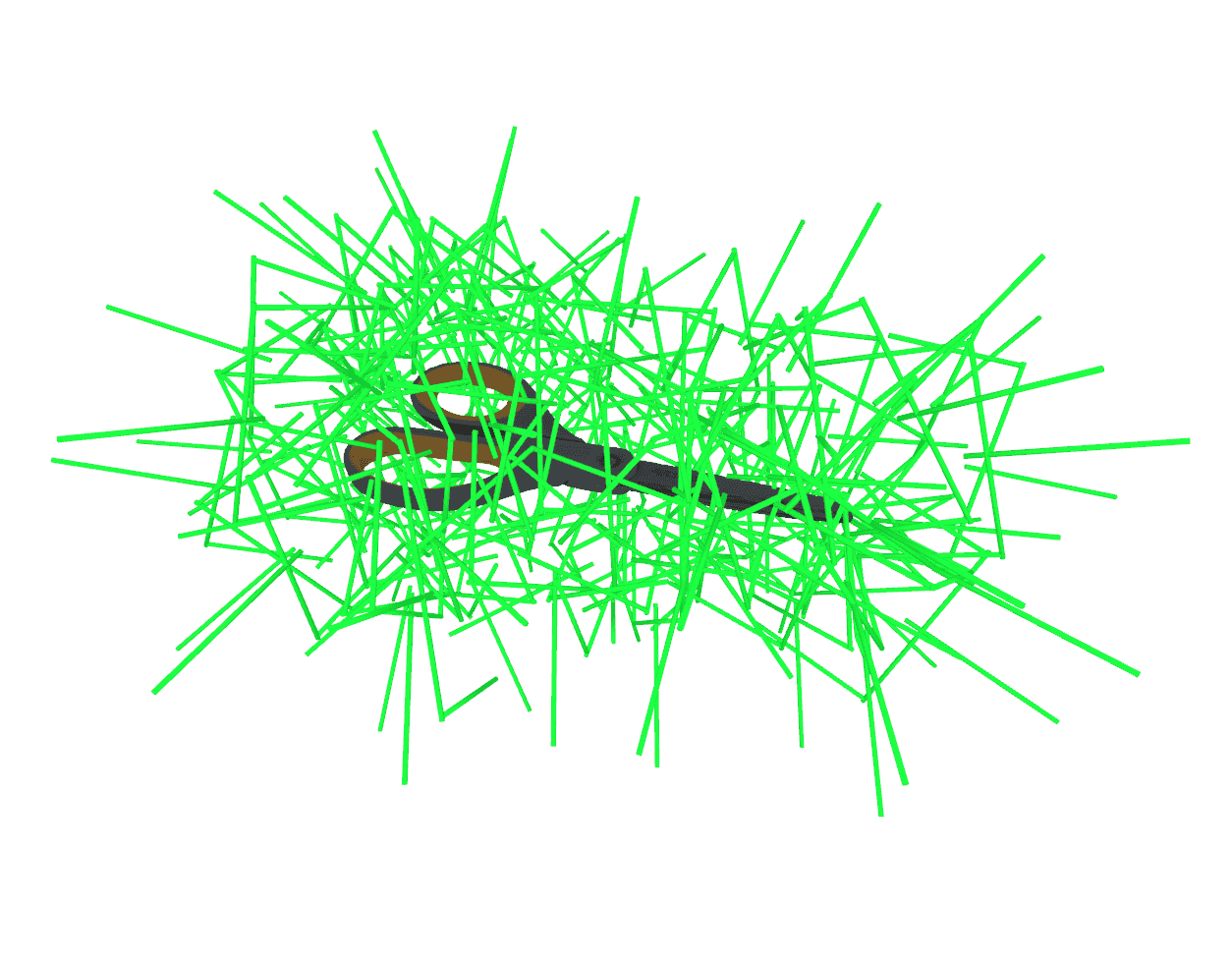}~
 \includegraphics[width=0.24\linewidth]{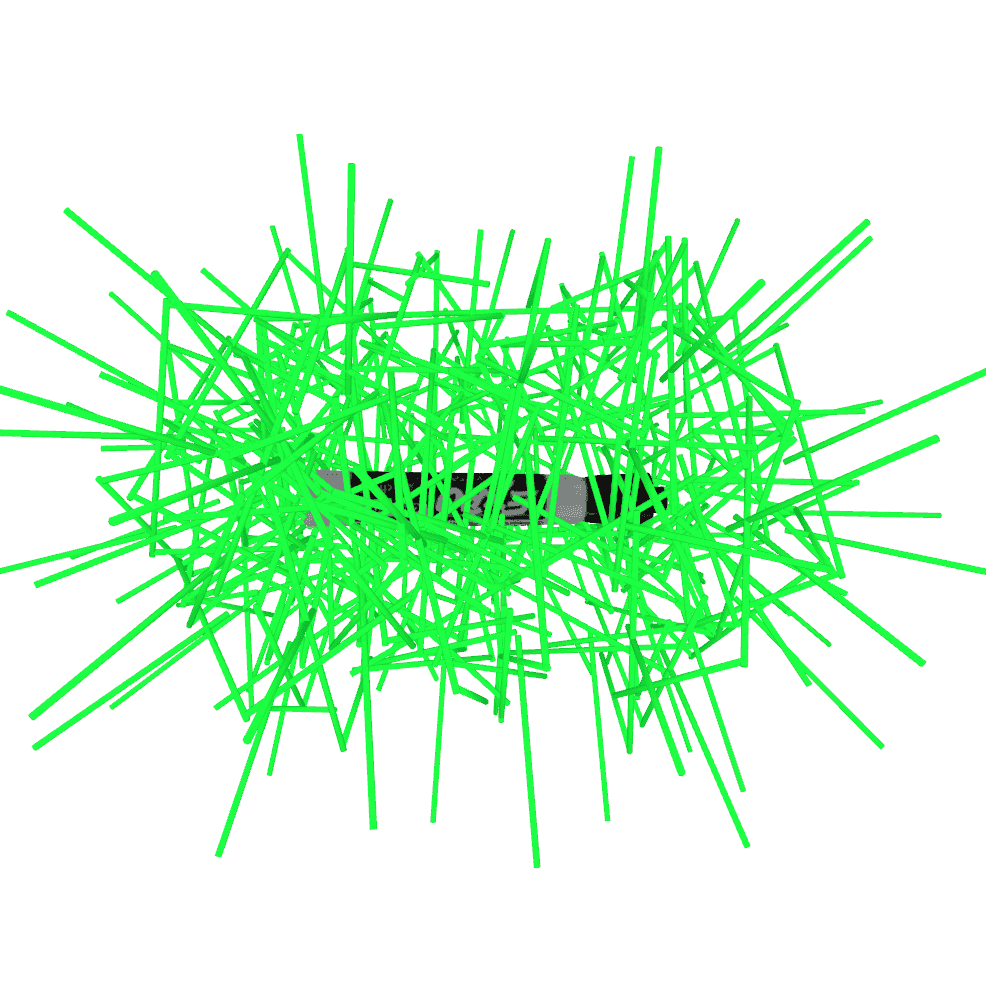}
 \\ \vspace{1mm}
 \includegraphics[width=0.24\linewidth]{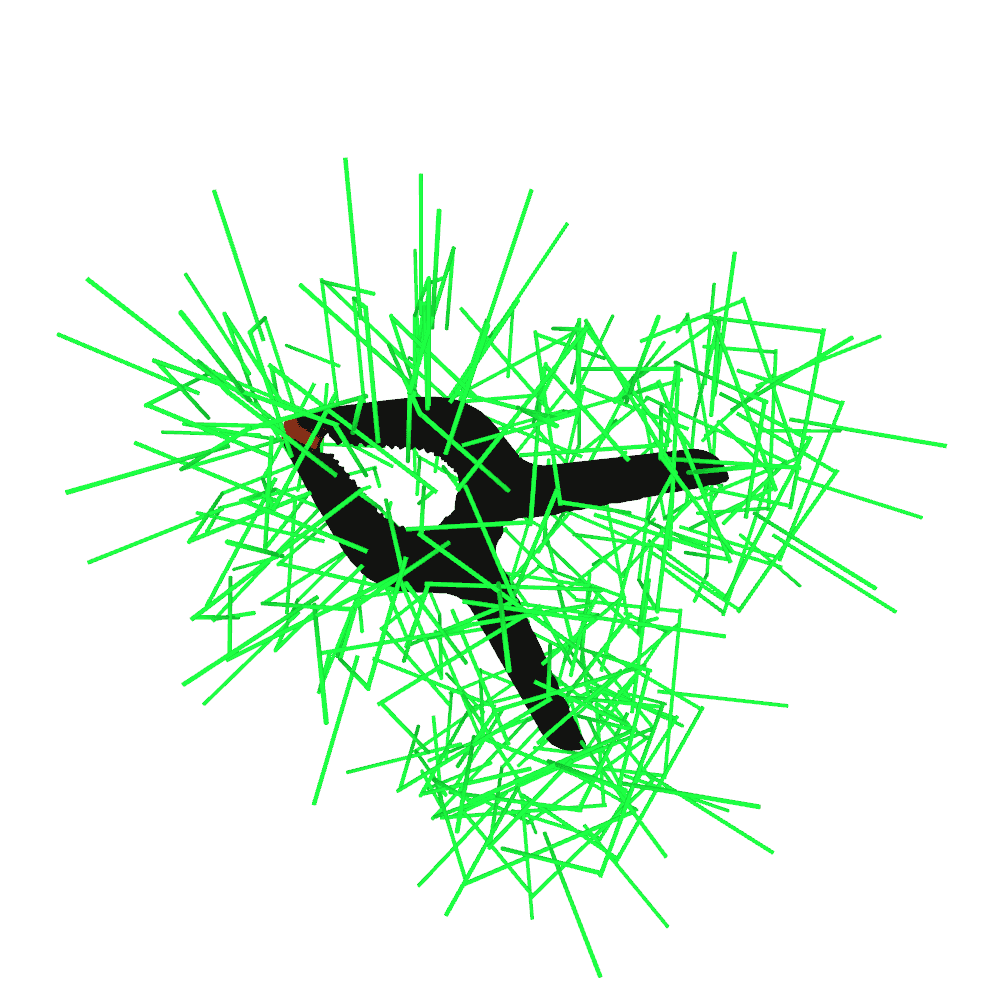}~
 \includegraphics[width=0.24\linewidth]{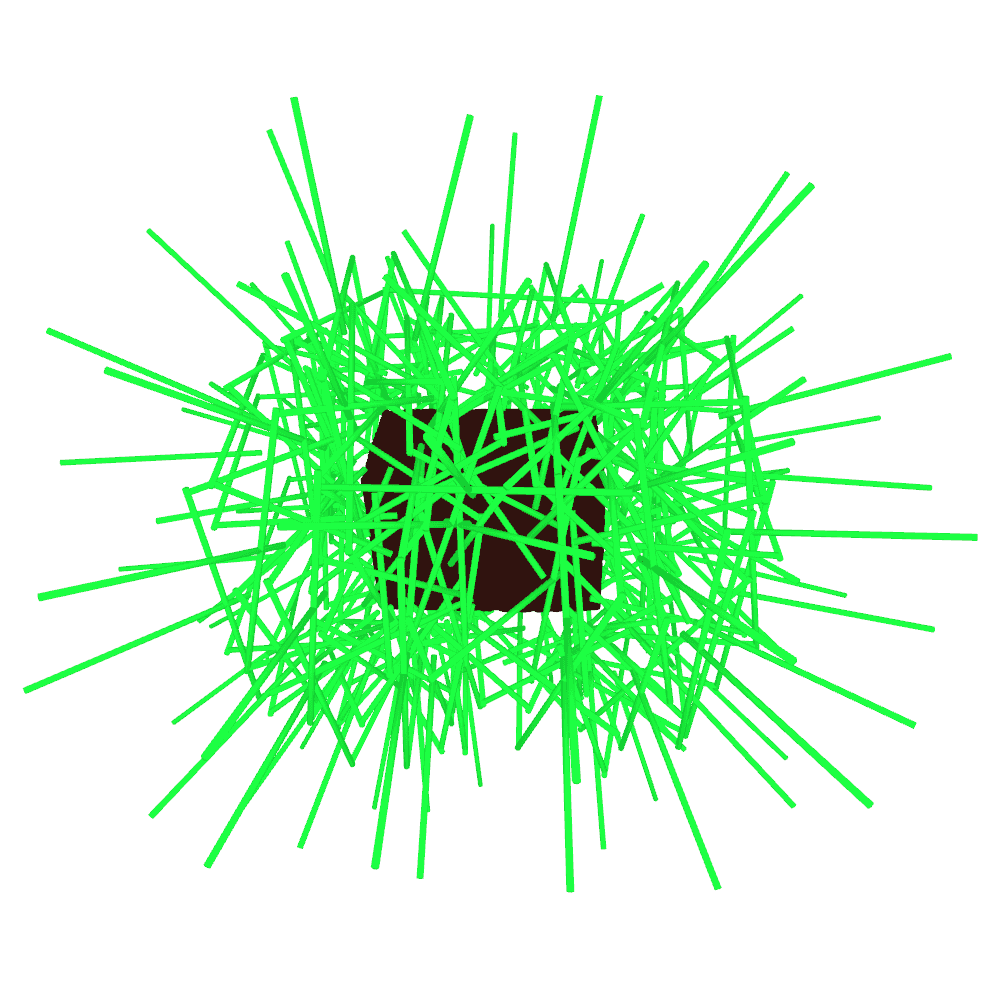}
 \caption{\small The 100 pre-generated grasps for each object.}
 \label{fig:100-grasp}
\end{figure*}

Fig.~\ref{fig:100-grasp} visualizes the 100 grasps for each object used in our
evaluation. These grasps are sampled from the pre-generated grasps for YCB
objects in~\cite{eppner:isrr2019}. Note that here we only show the grasps for
18 out of 20 objects in DexYCB, since the remaining two objects
(``002\_master\_chef\_can'' and ``036\_wood\_block'') do not have any feasible
grasps using the gripper of choice (i.e., Franka Panda).

\subsection{Additional Qualitative Results}

Fig.~\ref{fig:grasp} shows additional qualitative results of the predicted
grasps for human-to-robot object handover. Interestingly, larger objects like
``003\_cracker\_box'' (row 1 and column 1) are less tolerant to errors in pose
estimation, since most successful grasps requires the gripper to be fully open
and barely fitting the object. Therefore a slight error in the estimated pose
will cause the gripper to collide with the object. At the same time, errors in
object pose estimation may cause the gripper to miss the grasp especially on
smaller objects (e.g., gray grasps for ``037\_scissors'' in row 4 and column
3). On the other hand, a hand that is miss or partially detected due to
occlusion may cause a potential pinch by the gripper (e.g., red grasps for
``061\_foam\_brick'' in row 5 and column 3).

\subsection{Full Quantitative Results}

Since in Fig.~\ref{fig:curve} we show the results of grasp
generation only on S1, we now include the results on the other three setups.
Fig.~\ref{fig:curve:supp} shows the precision-coverage curves on S0, S2, and S3,
respectively. Overall, similar to on S1, we can see that more accurate object
pose estimation leads to better performance in grasp generation.

It is worth noting that although both DeepIM (RGB-D)~\cite{li:eccv2018} and
CosyPose~\cite{labbe:eccv2020} achieved very close performance on 6D object
pose (e.g. 57.54 versus 57.43 AR in Tab.~\ref{tab:6d:sota}), the
later performs significantly better on grasp generation (e.g. see
Fig.~\ref{fig:curve}). This is because that DeepIM (RGB-D)
maintains a competitive average recall by only outperforming CosyPose on a
smaller set of objects (e.g. 7 on S1) but with larger margins, whereas CosyPose
shows an edge over DeepIM (RGB-D) on more objects (e.g. 13 on S1). This shows
that a higher performance on 6D pose metrics like AR does not necessarily
translate to a higher performance on downstream robotics tasks like object
handover.

\section{Results on In-the-Wild Images}

\begin{figure*}
 \centering
 \includegraphics[height=0.244\linewidth]{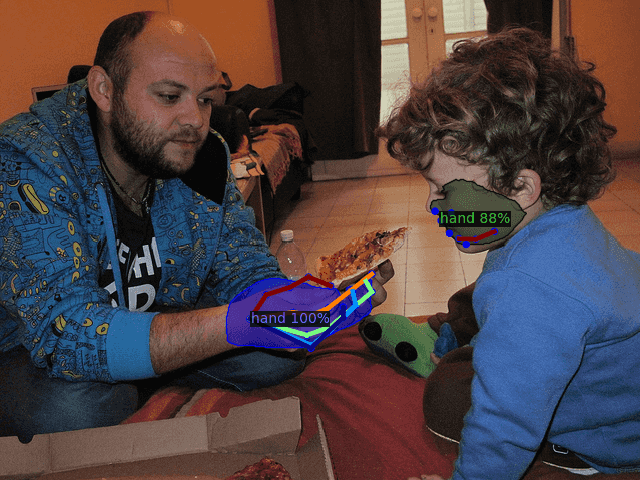}~
 \includegraphics[height=0.244\linewidth]{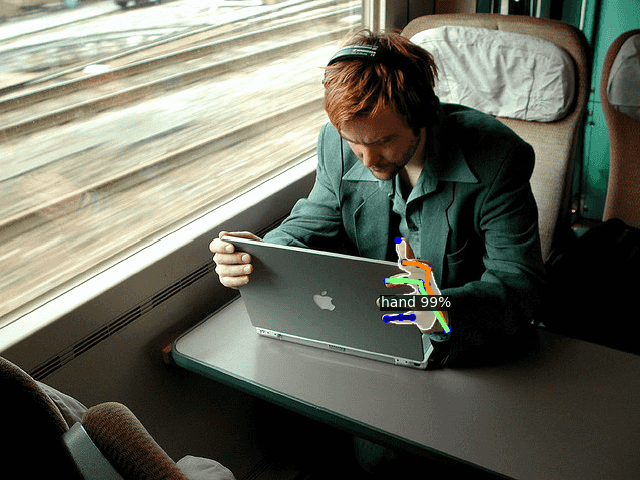}~
 \includegraphics[height=0.244\linewidth]{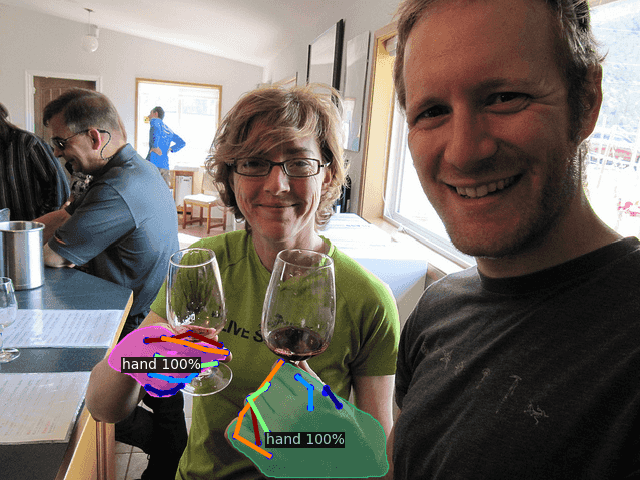}~
 \\ \vspace{1mm}
 \includegraphics[height=0.225\linewidth]{rebuttal-coco-00000042.png}~
 \includegraphics[height=0.225\linewidth]{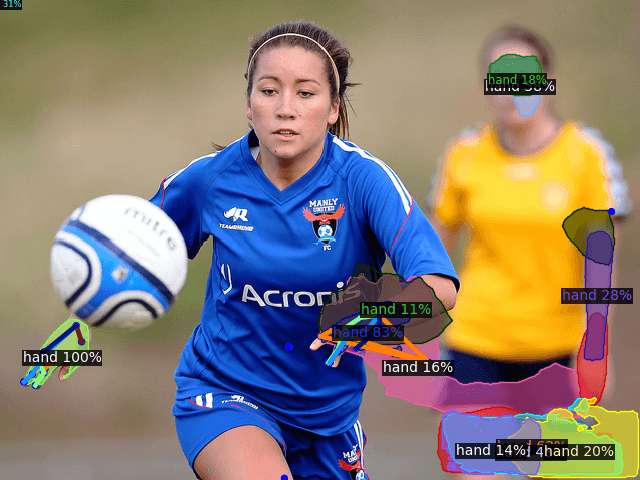}~
 \includegraphics[height=0.225\linewidth]{rebuttal-coco-00000398.png}
 \\ \vspace{1mm}
 \includegraphics[height=0.36\linewidth]{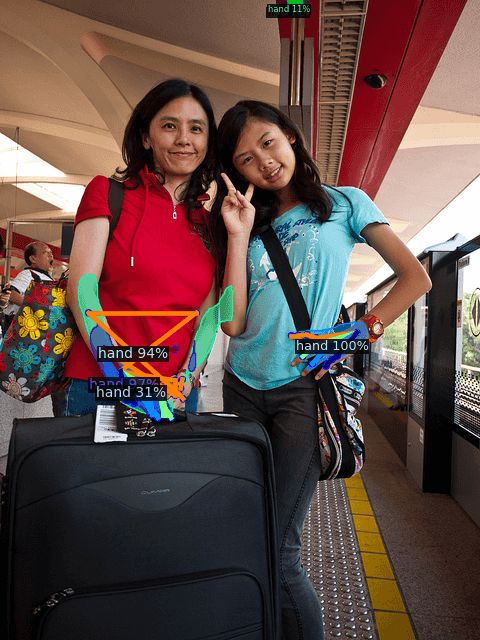}~
 \includegraphics[height=0.36\linewidth]{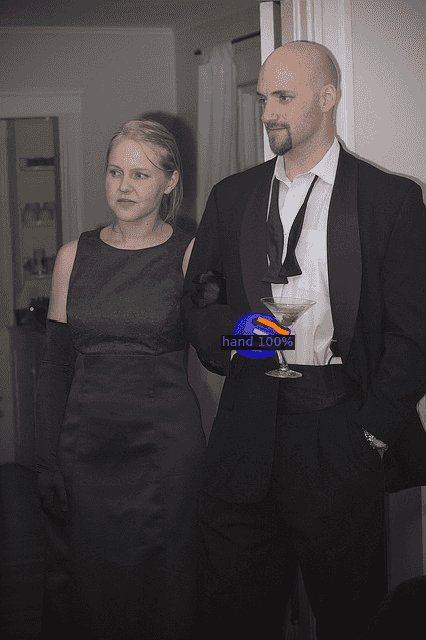}~
 \includegraphics[height=0.36\linewidth]{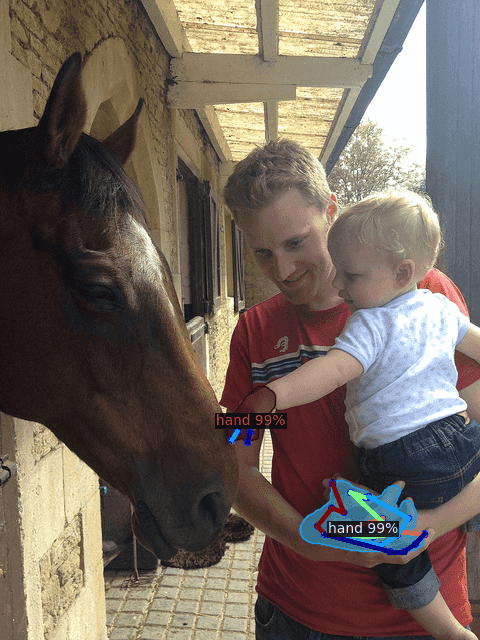}
 \caption{\small Samples of 2D hand and keypoint detection on COCO.}
 \label{fig:coco}
\end{figure*}

Since DexYCB was captured in a controlled lab environment with a constant
background, models trained on the RGB images of DexYCB are not expected to
generalize well to in-the-wild images. We tested a DexYCB-trained model on COCO
images~\cite{lin:eccv2014} and observed an expected drop in performance.
Fig.~\ref{fig:coco} shows qualitative examples of 2D hand and keypoint
detection. We can see that the detector frequently produces false positives
(top left), false negatives (top middle), and inaccurate keypoint detection
(top right). Combining DexYCB with other in-the-wild datasets for training will
be an interesting follow up work.

\begin{figure*}[t]
 \centering
 \includegraphics[width=0.24\linewidth]{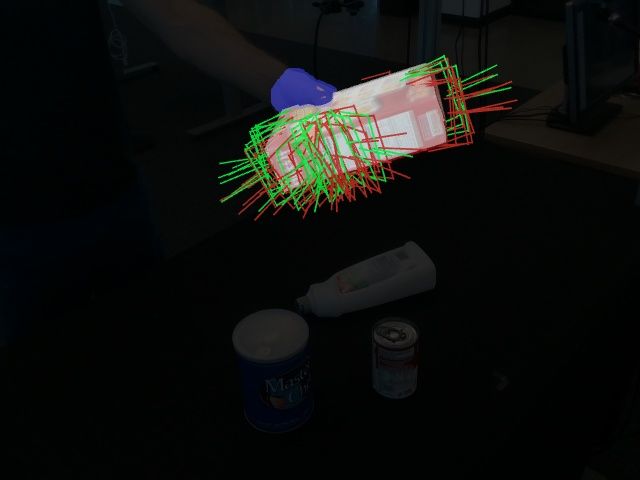}~
 \includegraphics[width=0.24\linewidth]{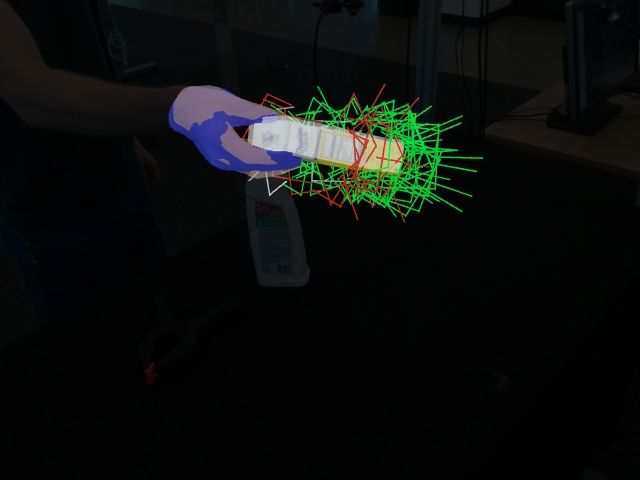}~
 \includegraphics[width=0.24\linewidth]{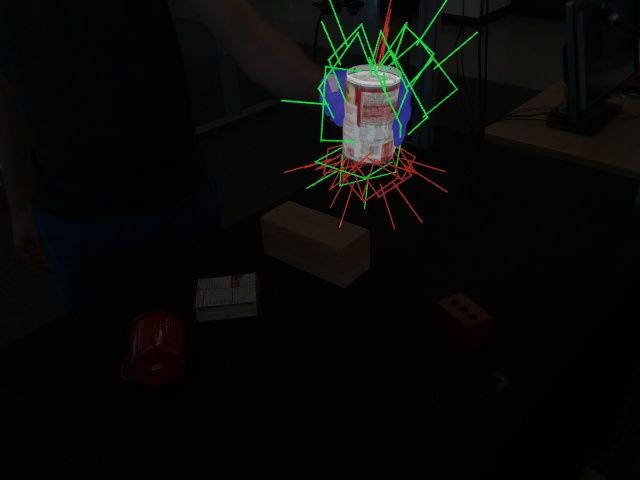}~
 \includegraphics[width=0.24\linewidth]{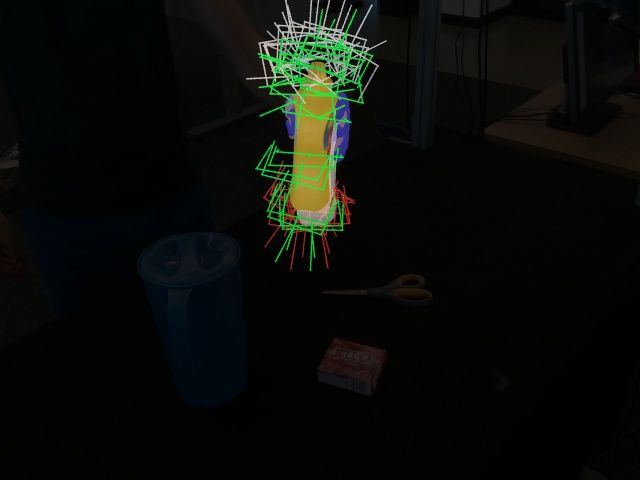}
 \\ \vspace{1mm}
 \includegraphics[width=0.24\linewidth]{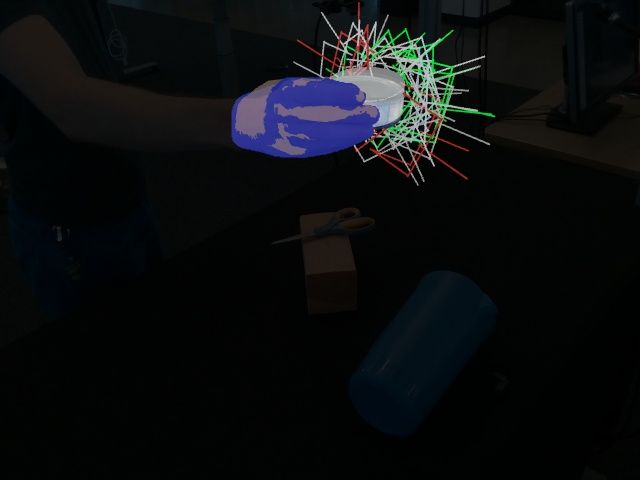}~
 \includegraphics[width=0.24\linewidth]{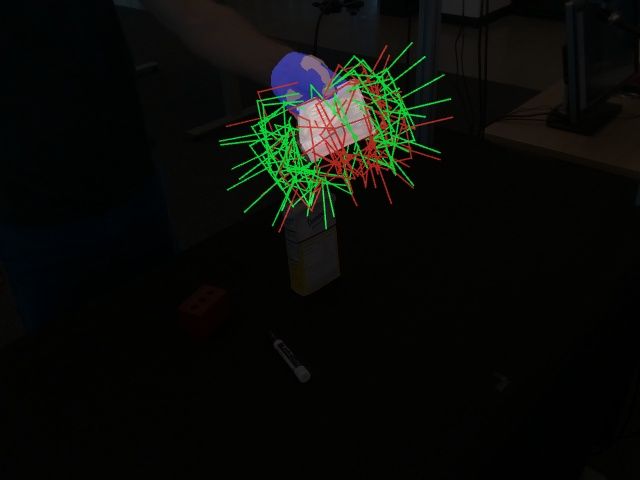}~
 \includegraphics[width=0.24\linewidth]{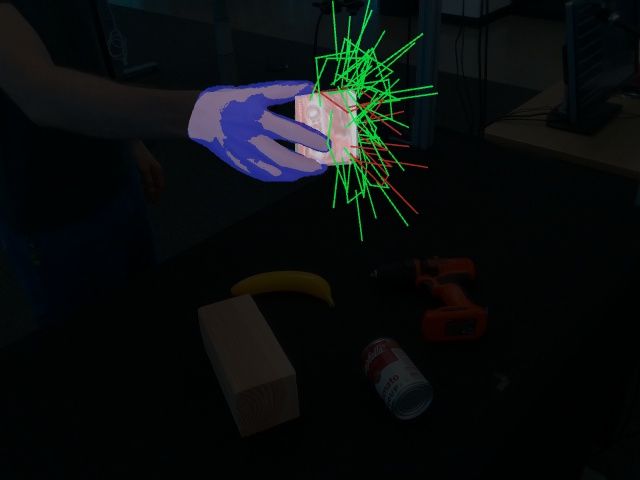}~
 \includegraphics[width=0.24\linewidth]{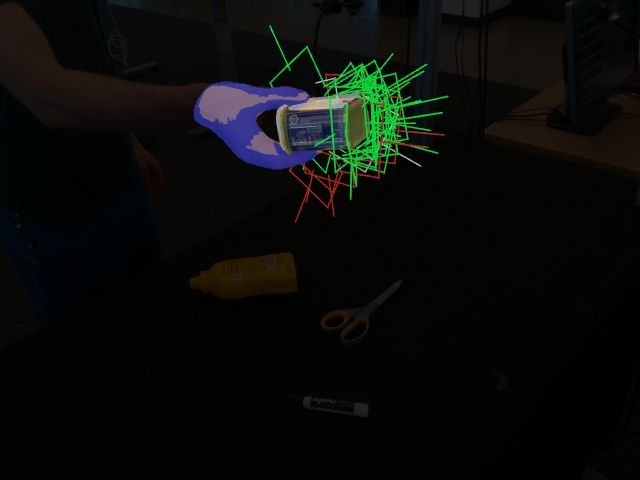}
 \\ \vspace{1mm}
 \includegraphics[width=0.24\linewidth]{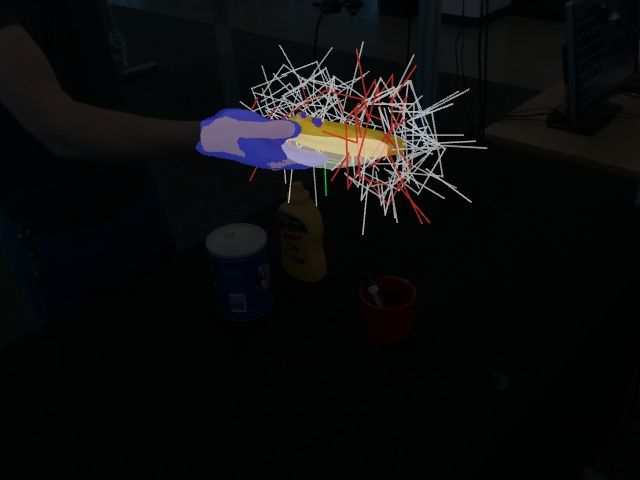}~
 \includegraphics[width=0.24\linewidth]{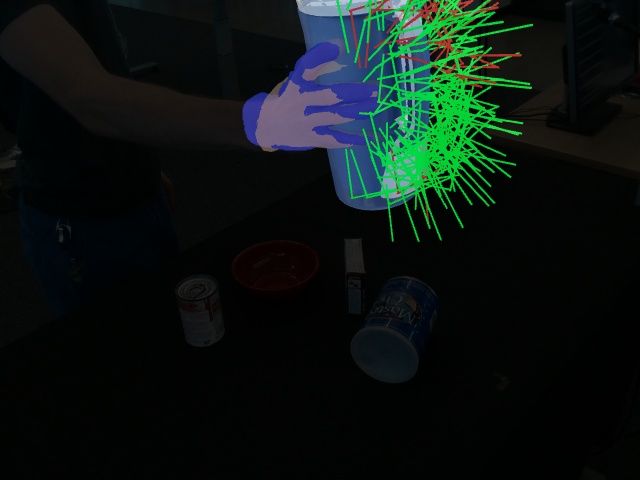}~
 \includegraphics[width=0.24\linewidth]{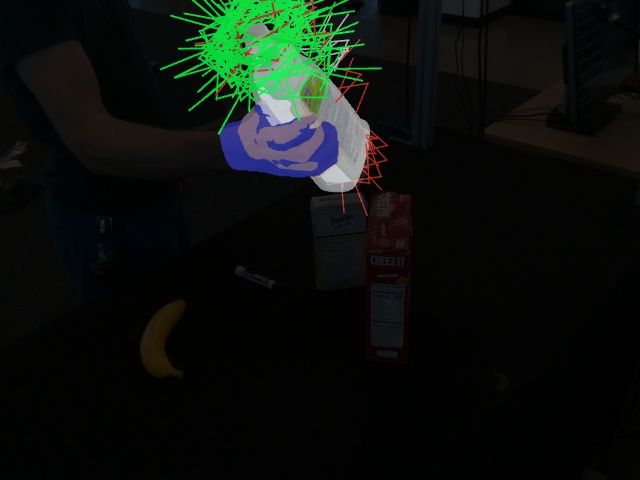}~
 \includegraphics[width=0.24\linewidth]{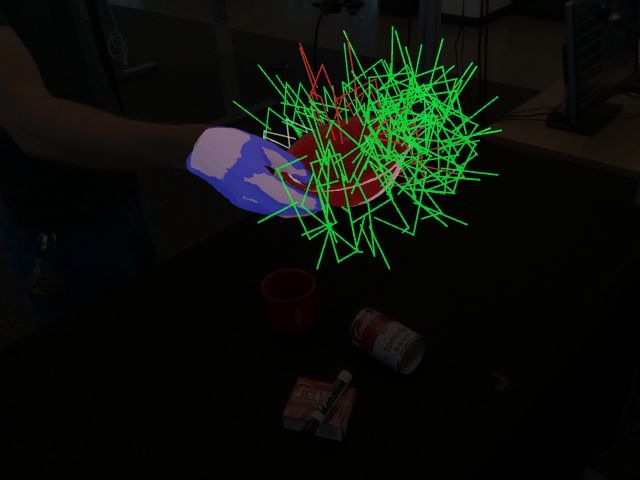}
 \\ \vspace{1mm}
 \includegraphics[width=0.24\linewidth]{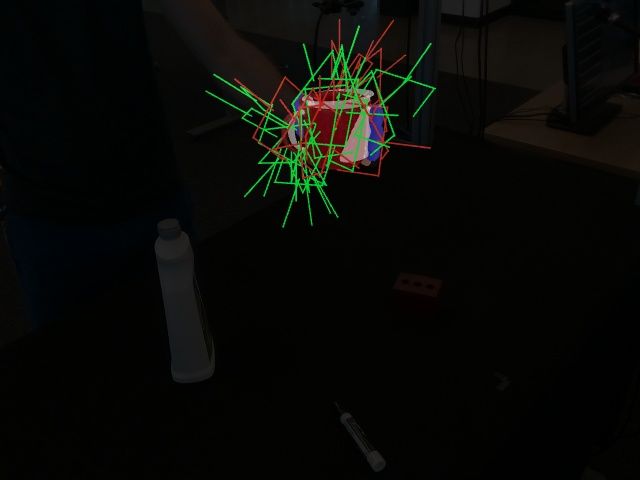}~
 \includegraphics[width=0.24\linewidth]{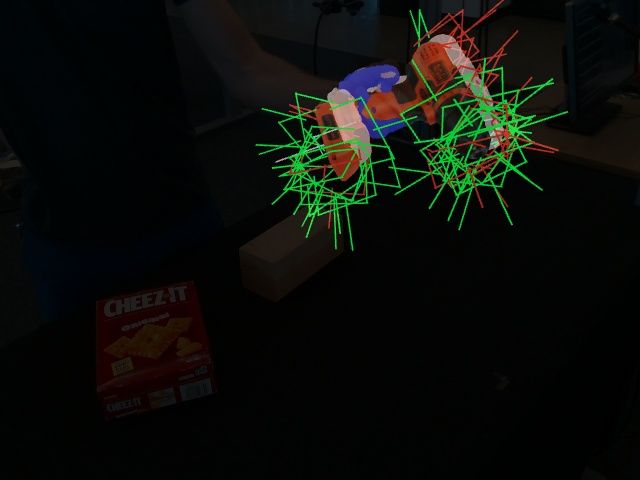}~
 \includegraphics[width=0.24\linewidth]{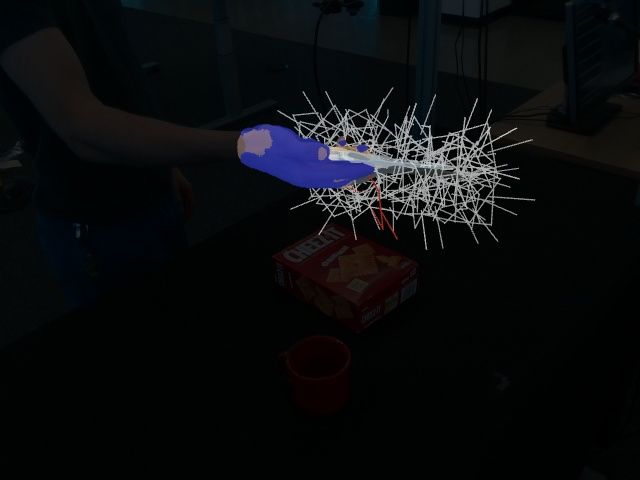}~
 \includegraphics[width=0.24\linewidth]{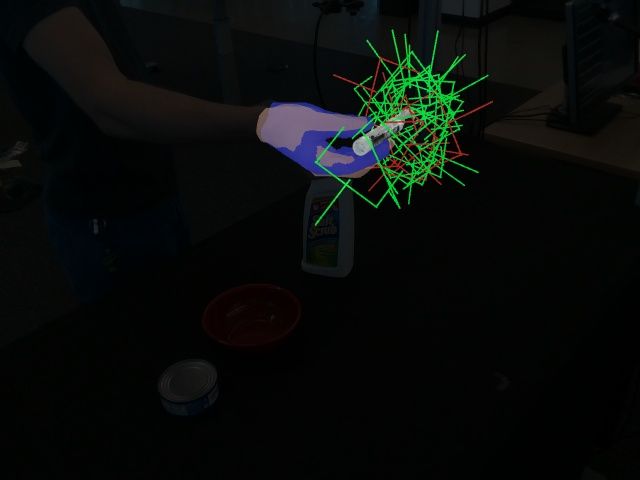}
 \\ \vspace{1mm}
 \includegraphics[width=0.24\linewidth]{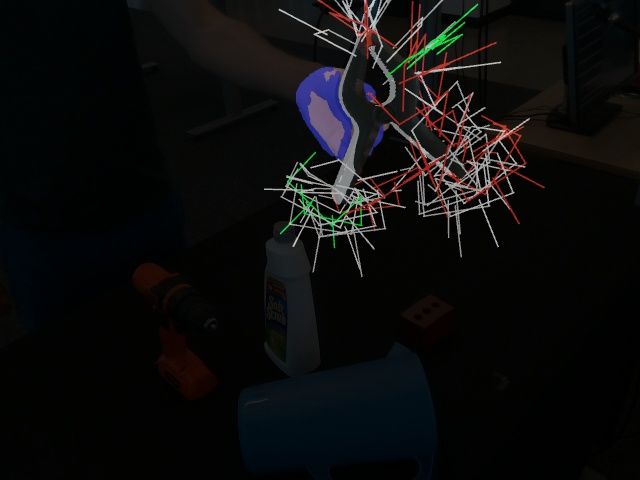}~
 \includegraphics[width=0.24\linewidth]{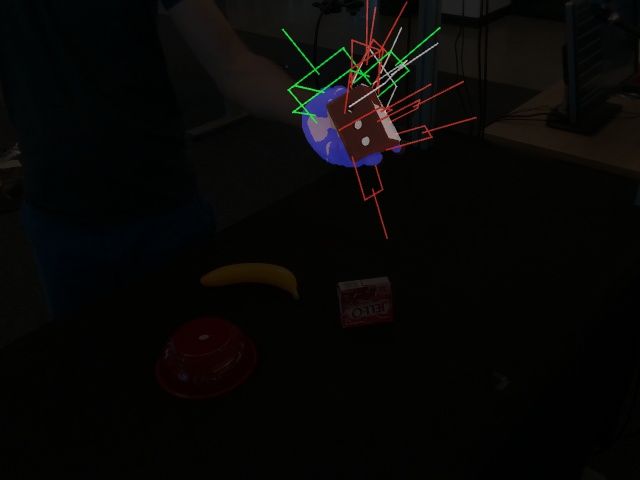}
 \caption{\small Additional qualitative results of the predicted grasps. Green
ones denote those covering successful grasps, red ones denote those collided
with the object or hand, and gray ones are failures not covering any successful
grasps in the reference set. Predicted object poses are visualized by textured
models and hand segmentations are highlighted by blue masks. Ground-truth
objects and hands are shown in translucent white and brown meshes.}
 \label{fig:grasp}
\end{figure*}

\end{document}